\documentclass[10pt,twocolumn,letterpaper]{article}

\usepackage[pagenumbers]{cvpr} 

\usepackage[dvipsnames]{xcolor}
\newcommand{\myparagraph}[1]{\noindent\textbf{#1}}

\newif\ifdraft
\draftfalse

\ifdraft

\newcommand{\kac}[1]{{\color{orange}[\textbf{Kfir:} \textit{#1}]}}
\newcommand{\jwc}[1]{{\color{blue}[\textbf{Jian:} \textit{#1}]}}
\newcommand{\dosc}[1]{{\color{teal}[\textbf{Daniil:} \textit{#1}]}}
\newcommand{\smc}[1]{{\color{violet}[\textbf{Sizhuo:} \textit{#1}]}}

\newcommand{\pc}[1]{{\color{magenta}#1}}

\else
\newcommand{\pcc}[1]{}
\newcommand{\kac}[1]{}
\newcommand{\dosc}[1]{}
\newcommand{\jwc}[1]{}
\newcommand{\smc}[1]{}

\newcommand{\pc}[1]{{\color{black}#1}}

\fi
\usepackage{blindtext}

\definecolor{cvprblue}{rgb}{0.21,0.49,0.74}
\usepackage[pagebackref,breaklinks,colorlinks,citecolor=cvprblue]{hyperref}

\title{Personalized Restoration via Dual-Pivot Tuning}

\author{Pradyumna Chari$^{1,\dagger}$
\and
Sizhuo Ma$^2$
\and 
Daniil Ostashev$^2$
\and
Achuta Kadambi$^1$
\and
Gurunandan Krishnan$^2$
\and
Jian Wang$^{2,\star}$
\and
Kfir Aberman$^2$
\and
$^1$University of California, Los Angeles
\and
$^2$Snap Inc.\\
}

\begin{document}
\twocolumn[{%
\renewcommand\twocolumn[1][]{#1}%
\maketitle
    \captionsetup{type=figure}

 \includegraphics[width=\linewidth]{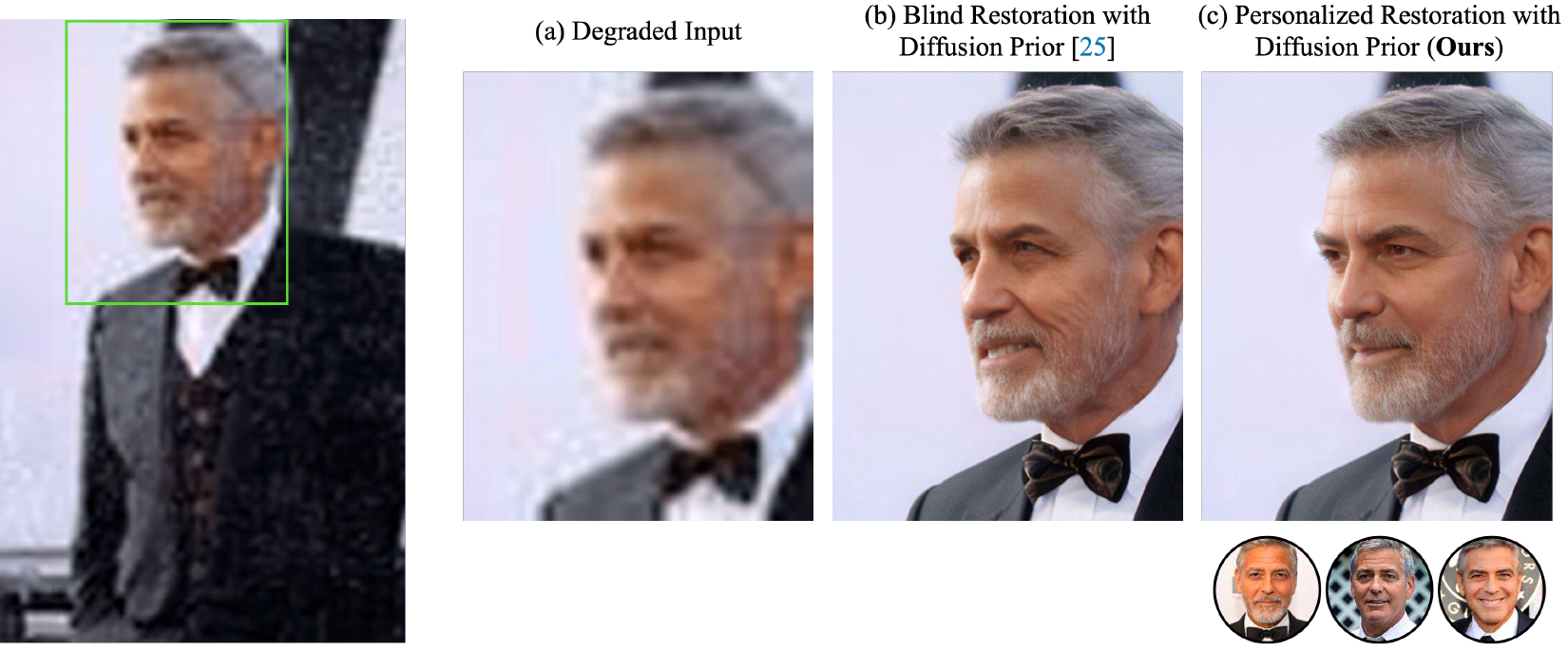}
  \centering

    \caption{
    Given a degraded image of an individual's face (a), diffusion-based blind restoration approaches~\cite{lin2023diffbir} may not retain the individual's identity (b). However, with a few reference images  (bottom right), our \emph{dual-pivot tuning} technique (c) can reconstruct the face while maintaining high identity fidelity to the individual and without any perceivable loss in fidelity to the degraded input.
    }

    \vspace{0.2cm}
    \label{fig:teaser}
}]

\renewcommand*{\thefootnote}{$\dagger$}
\setcounter{footnote}{1}
\footnotetext{Work partially done during an internship at Snap Inc.}
\renewcommand*{\thefootnote}{\arabic{footnote}}
\setcounter{footnote}{0}

\renewcommand*{\thefootnote}{$\star$}
\setcounter{footnote}{1}
\footnotetext{Corresponding author}
\renewcommand*{\thefootnote}{\arabic{footnote}}
\setcounter{footnote}{0}

\begin{abstract}

Generative diffusion models can serve as a prior which ensures that solutions of image restoration systems adhere to the manifold of natural images. However, for restoring facial images, a personalized prior is necessary to accurately represent and reconstruct unique facial features of a given individual. In this paper, we propose a simple, yet effective, method for personalized restoration, called \textit{Dual-Pivot Tuning} —
a two-stage approach that personalize a blind restoration system while maintaining the integrity of the general prior and the distinct role of each component. 
Our key observation is that for optimal personalization, the generative model should be tuned around a fixed text pivot, while the guiding network should be tuned in a generic (non-personalized) manner, using the personalized generative model as a fixed ``pivot".
This approach ensures that personalization does not interfere with the restoration process, resulting in a natural appearance with high fidelity to the person's identity and the attributes of the degraded image. We evaluated our approach both qualitatively and quantitatively through extensive experiments with images of widely recognized individuals, comparing it against relevant baselines. Surprisingly, we found that our personalized prior not only achieves higher fidelity to identity with respect to the person's identity, but also outperforms state-of-the-art generic priors in terms of general image quality. Project webpage: \url{https://personalized-restoration.github.io}

\end{abstract}    
\section{Introduction}
\label{sec:intro}

\begin{figure*}[t]
    \vspace{-8mm}

    \begin{subfigure}[c]{0.16\linewidth}
        \centering    
        \includegraphics[width=\linewidth]{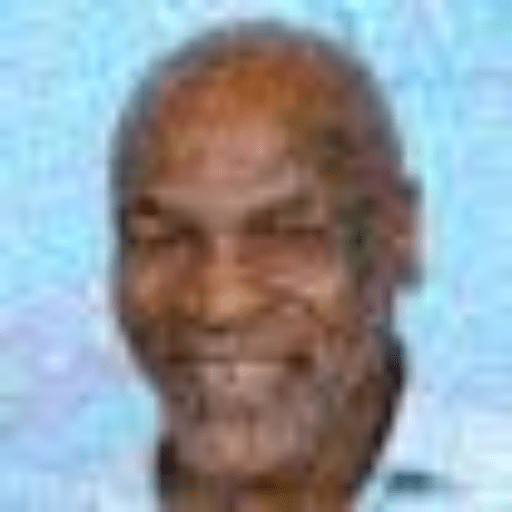}
    \end{subfigure}
    \begin{subfigure}[c]{0.16\linewidth}
        \centering  
        \includegraphics[width=\linewidth]{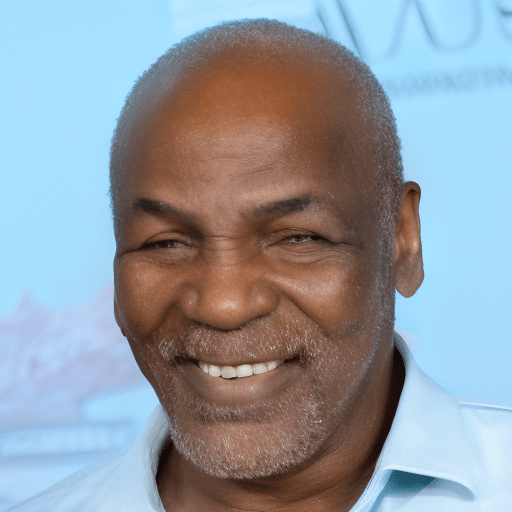}
    
    \end{subfigure}
    \begin{subfigure}[c]{0.16\linewidth}
        \centering    
        \includegraphics[width=\linewidth]{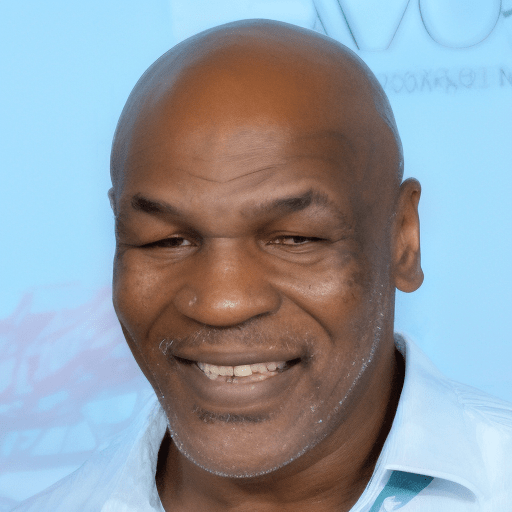}
    
    \end{subfigure}
    \begin{subfigure}[c]{0.16\linewidth}
        \centering    
        \includegraphics[width=\linewidth]{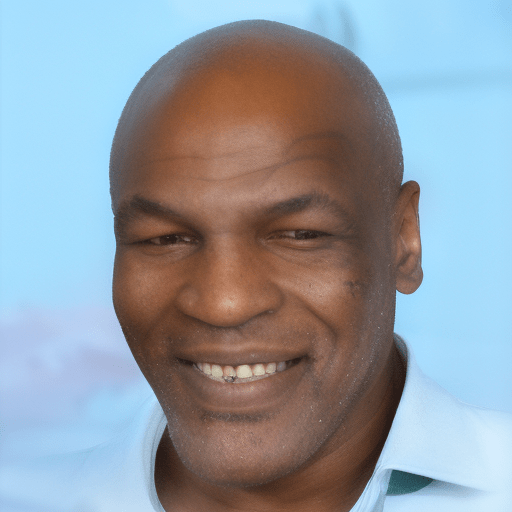}
   
    \end{subfigure}
    \begin{subfigure}[c]{0.16\linewidth}
        \centering    
         \includegraphics[width=\linewidth]{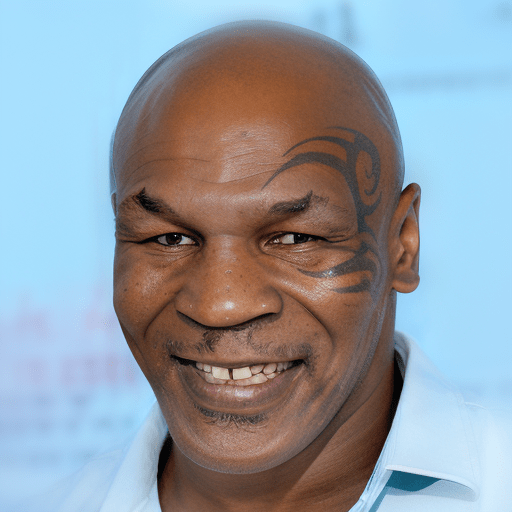}
    
    \end{subfigure}
    \begin{subfigure}[c]{0.16\linewidth}
        \centering    
        \includegraphics[width=\linewidth]{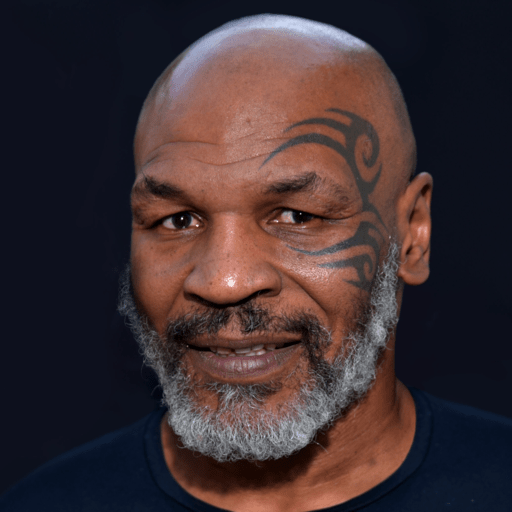}
    \end{subfigure}\\

    \begin{subfigure}[c]{0.16\linewidth}
        \centering    
        \vspace{-0.35cm}
        \includegraphics[width=\linewidth]{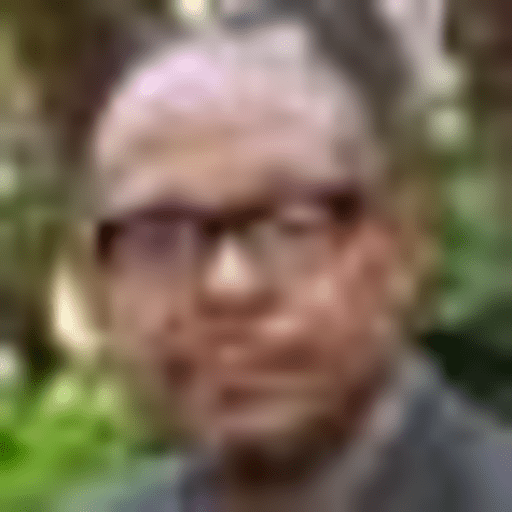}
        \caption{\footnotesize{Input}}
    \end{subfigure}
    \begin{subfigure}[c]{0.16\linewidth}
        \centering  
        \vspace{-0.35cm}
        \includegraphics[width=\linewidth]{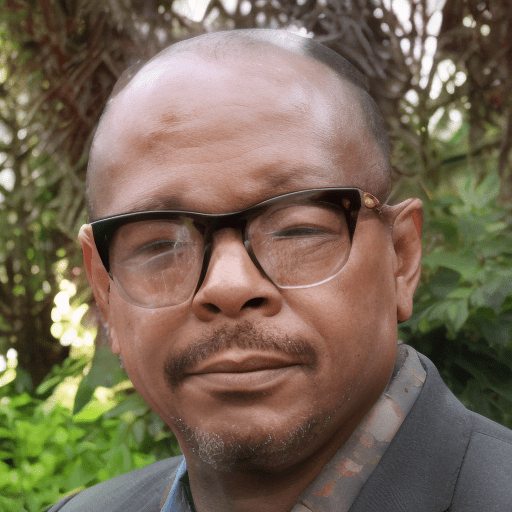}
        \caption{\footnotesize{DiffBIR~\cite{lin2023diffbir}}}
    \end{subfigure}
    \begin{subfigure}[c]{0.16\linewidth}
        \centering    
        \vspace{-0.35cm}
        \vspace{0.32cm}
        \includegraphics[width=\linewidth]{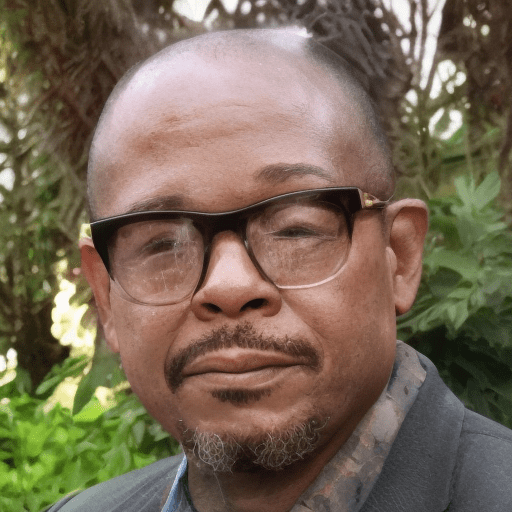}
    
        \caption{Baseline 1 (Pivot 1 only)}
    \end{subfigure}
    \begin{subfigure}[c]{0.16\linewidth}
        \centering    
        \vspace{-0.35cm}
        \vspace{0.32cm}
        \includegraphics[width=\linewidth]{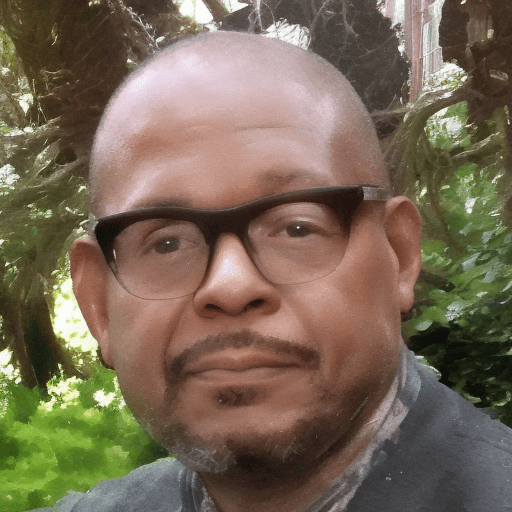}
   
        \caption{Baseline 2 (Pivot 2 only)}
    \end{subfigure}
    \begin{subfigure}[c]{0.16\linewidth}
        \centering    
        \vspace{-0.35cm}
         \includegraphics[width=\linewidth]{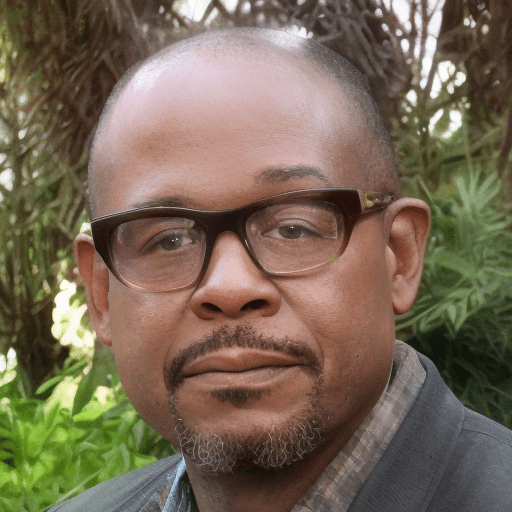}
    
        \caption{\footnotesize{\textbf{Ours}}}
    \end{subfigure}
    \begin{subfigure}[c]{0.16\linewidth}
        \centering  
        \vspace{-0.35cm}
        \includegraphics[width=\linewidth]{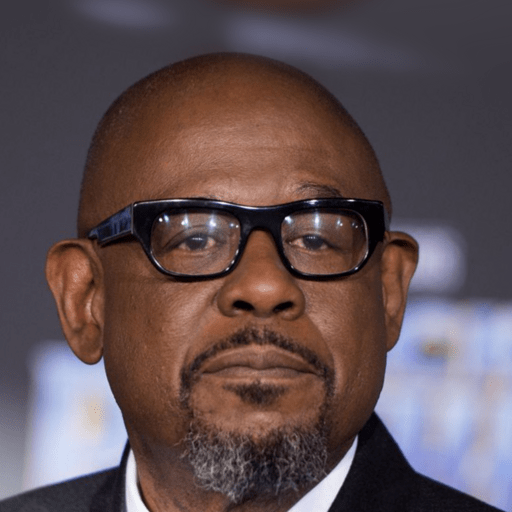}
        \caption{\footnotesize{Reference image}}
    \end{subfigure}
    \caption{\textbf{Restoration Baselines.} Given a real, degraded input image (a) features a person whose identity is referenced in another image (f), a diffusion-based blind restoration method~\cite{lin2023diffbir} with a general (non-personalized) prior, fails to preserve the person's identity (b). When fine-tuning the text-to-image prior with a text pivot only~\cite{ruiz2023dreambooth}, the system ignores the personalized generator when reintegrated into the system (c), while personalizing the feature encoder with the generative model as a pivot only leads to lack of detail in the result (tattoo in the first row, detail such as beard, general high frequencies in the second row), along with absence of generalization across identities (d). In contrast, our method (e) effectively incorporates the individual's identity into the restoration process while retaining a quality comparable to the base model (b). 
    }
    \label{fig:method_motivation}
    \vspace{-3mm}
\end{figure*}

Image restoration is an extensively researched problem, characterized by its inherent ill-posed nature. The goal of the restoration task is to find visually plausible and natural images that maintain perceptual fidelity to the degraded input image~\cite{wang2020deep}.
In blind restoration scenarios where no prior information about the subject or the degradation is available, a prior describing the manifold of natural images is needed. However, for face image restoration, having an identity prior is necessary to ensure that the output image remains within a manifold that accurately represents the distinctive facial features of the individual in the degraded image.
This sets the basis for prior work in reference-based face image restoration~\cite{li2020enhanced,li2022learning,nitzan2022mystyle}.

Text-to-Image Diffusion Models~\cite{rombach2022high,saharia2022photorealistic} have revolutionized image synthesis, enabling the generation of images from textual descriptions. These models act as versatile generative priors for various downstream tasks, and were recently explored in the context of blind image restoration, enhancing the naturalness of restored images~\cite{lin2023diffbir,wang2023exploiting}. Moreover, the ability to personalize these diffusion models with just a few reference images opens new avenues for tailored content creation~\cite{ruiz2023dreambooth, gal2022image}. Despite these advancements, effectively integrating personalization into a diffusion-based blind restoration systems remains an open challenge.

To highlight the challenges associated with personalizing a blind restoration system, \cref{fig:method_motivation} demonstrates how naive personalization approaches fall short.
As shown in \cref{fig:method_motivation} (c), a personalized text-to-image model that simply replaces the general one in a blind restoration system can be disregarded by the unconditional system. Alternatively, naive personalization of the image conditioning encoder (from~\cite{lin2023diffbir}, for example) disrupts the natural image prior, leading to a lack of detail in the restored image as demonstrated in \cref{fig:method_motivation} (d).

In this paper, we propose a simple technique for personalizing a text-to-image prior within the context of a blind image restoration framework. Given a small set ($\sim$10) of high-quality images of a person, our goal is to restore their other degraded images, ensuring that the results exhibit (1) high fidelity to the identity of the person, (2) high fidelity to the degraded image, (3) natural appearance.

Our key idea, coined \textit{dual-pivot tuning}, is to fine-tune the prior within the context of a blind image restoration system, while preserving the integrity of the general prior and the distinct roles of each component within the system.
Specifically, we propose a two-step solution, to personalize a general blind image restoration system which consists of two main components: a diffusion-based generative prior, and a guiding image encoder. 

In the first step, we personalize the diffusion-based generative prior in the context of the blind restoration system, while using a text prompt (for example, ``a photo of a [v] man") as a \emph{textual pivot} - a fixed token that is held constant during the fine-tuning process~\cite{ruiz2023dreambooth}. However, in our case the fine-tuning is performed with the guiding image encoder such that the personalized prior learns to respect the attributes of the guiding image. In the second step, we fine-tune the guiding encoder to align with the personalized, ``shifted", prior of the diffusion model, thereby retargeting its functionality. During this phase, we aim to maintain the identity-agnostic nature of the  encoder, allowing its applicability across various personalized models. Therefore, we fine-tune it with general face images (not of the specific individual). We refer to the fixed personalized network as a pivot, around which we adjust the weights of the guiding encoder.

We find both of these operations, in this sequence, to be essential. While the textual pivot enables identity injection without losing the general face prior of the base model in the restoration system, fine-tuning around the network pivot leads to better utilization of the guiding encoder and higher fidelity to the input image features. Furthermore, we leverage the diffusion process's characteristics, noting that identity formation occurs later in the process, where it primarily shapes finer details rather than the initial, noisier phase that sets coarse image features. Our approach demonstrates that personalization can be effectively applied at specific stages of the diffusion process which can reduce the expansive fine-tuning time by approximately $\times 2$.

We compare our method against few-shot learning baselines and evaluate them using publicly available images of well-known figures, leveraging our pre-existing knowledge of their features. 
Our experimenters shows that our method reconstruct key facial features of the subject in the reference images while maintaining high fidelity to the original degraded image, outperforming others quantitatively and qualitatively. Moreover, user studies, interestingly, confirm that this personalization aspect contributes significantly not only to identity preservation but also to the perceived overall quality improvement of the generated images. \cref{fig:teaser} shows one example that demonstrates how our method is able to surpass existing diffusion-based blind face restoration methods~\cite{lin2023diffbir} in terms of identity retention while maintaining the same quality of face image restoration.

To the best of our knowledge, this is the first approach to leverage a personalized diffusion prior for restoration tasks.

\section{Related Work}
\label{sec:related_work}
\subsection{Blind Face Image Restoration}

Distinct from general scene image restoration, face image restoration typically leverages facial priors to achieve superior results. Depending on how these facial priors are utilized, previous methods can be broadly categorized into three main classes.
(1) Geometric prior: These approaches incorporate geometric cues, such as facial landmarks \cite{chen2018fsrnet, bulat2018super, kim2019progressive}, parsing maps \cite{chen2021progressive,shen2018deep,yang2020hifacegan}, and component heatmaps \cite{yu2018face}, into the network's design.
(2) Dictionary prior: In this method, a dictionary is first learned from a collection of face images, either in the image space \cite{li2020blind} or feature space \cite{wang2022restoreformer,zhao2022rethinking, gu2022vqfr, zhou2022towards}. Subsequently, a degraded image is reconstructed using high-quality words from this dictionary.
(3) Generative prior: This category encompasses techniques like GAN (Generative Adversarial Network) prior \cite{chan2021glean,yang2021gan,wang2021gfpgan,luo2021time} or diffusion prior \cite{yue2022difface,lin2023diffbir,wang2023dr2}. Among these methods, approaches falling under categories (2) and (3) have demonstrated the most promising results. Notable algorithms in this context include GFP-GAN \cite{wang2021gfpgan}, CodeFormer \cite{zhou2022towards}, and DiffBIR \cite{lin2023diffbir}.

However, blind face image restoration faces a significant challenge known as the quality-fidelity tradeoff \cite{zhou2022towards,lin2023diffbir}. This arises from the inherent limitations in the information available in the original image. Striking the right balance between generating high-quality results while staying faithful to the original image can be a delicate task. Generating results with too little modification may not yield a significant improvement in quality, while excessive generation can lead to a departure from the identity of the original image. Our proposed method differs from these methods: using personalized diffusion-based methods, we are able to push the tradeoff frontiers by achieving restoration quality while retaining fidelity with respect to identity.

\subsection{Reference-Based Face Image Restoration}

A high-quality reference image of the same person can greatly benefit face image restoration and help avoid the need of the tradeoff. Depending on the number of references used, such methods can be divided into two categories.
(1) Single-reference methods: Examples include GFRNet ~\cite{Li_2018_ECCV} and GWAINet \cite{Dogan_2019_CVPR_Workshops}.
(2) Multi-reference methods: Examples include ASFFNet \cite{li2020enhanced}, Wang \etal \cite{wang2020multiple}, and DMDNet \cite{li2022learning} ($\leq$ 10 reference images), and MyStyle~\cite{nitzan2022mystyle} ($\sim$ 100 images). 
It is evident that multi-reference approaches yield superior results as they leverage more information. Notably, ASFFNet \cite{li2020enhanced} selects an optimal guidance, Wang \etal \cite{wang2020multiple} employs pixel-wise weights for multiple references, and DMDNet \cite{li2022learning} constructs a dictionary from multiple references; MyStyle \cite{nitzan2022mystyle}, on the other hand, fine-tunes StyleGAN based on personal images.

Our proposed method also utilizes multiple reference images to aid personalized restoration. However, there are several distinctions. We use a diffusion-based personalized generative prior, while~\cite{Li_2018_ECCV,Dogan_2019_CVPR_Workshops,li2020enhanced,wang2020multiple} use feedforward architectures. On the other hand,~\cite{nitzan2022mystyle} uses a GAN-based prior, however it requires on the order of 100 images for effective personalization and strict spatial alignment of the face landmarks within each image, while our method is able to operate with 10 reference images and has no restriction on the alignment, operating directly on images in the wild. These distinctions lead to higher quality restoration with less restrictions for our proposed method.

\subsection{Personalized Diffusion Models}

Diffusion models~\cite{ho2020denoising, song2021scorebased, dhariwal2021diffusion} have notably excelled in the area of generating images from text (T2I)~\cite{ramesh2022hierarchical, saharia2022photorealistic, rombach2022high} and many other visual computing tasks~\cite{po2023state}. Recent advancements in this field involve customization of these established model through fine-tuning, aiming to enhance features like controllability, customization, or to cater to specific applications. One approach to customization involves modifying the T2I model itself~\cite{ruiz2023dreambooth, chen2023subject,tewel2023key,tang2023realfill} or the text embedding process~\cite{gal2022image,alaluf2023neural,voynov2023p+}, using selected images. This allows to personalize the generation of images based on a particular subject or style, driven by textual input. More recently, neurons in a model specific to a concept can be identified and manipulated specifically to enable sparse personalization~\cite{liu2023cones}, methods for fast personalization of the model have been proposed~\cite{gal2023encoder,arar2023domain,ruiz2023hyperdreambooth}, and approaches for multisubject personalization~\cite{po2023orthogonal,avrahami2023break}. Alternatively, other methods involve adapting the T2I model to introduce new conditioning factors. These modifications are either for purposes of image modification~\cite{kawar2023imagic, brooks2023instructpix2pix, wang2023imagen}  or for generating more controlled images~\cite{zhang2023adding, mou2023t2i}.

In this work, we tackle the task of contextual customization, where the goal is to fine-tune a prior within the context of a system, preserving the distinct roles of each component within the system, while customizing the image prior to a specific subject.

\section{Method}
\label{sec:method}
\begin{figure*}[t]
    \centering    
    \includegraphics[width=\linewidth]{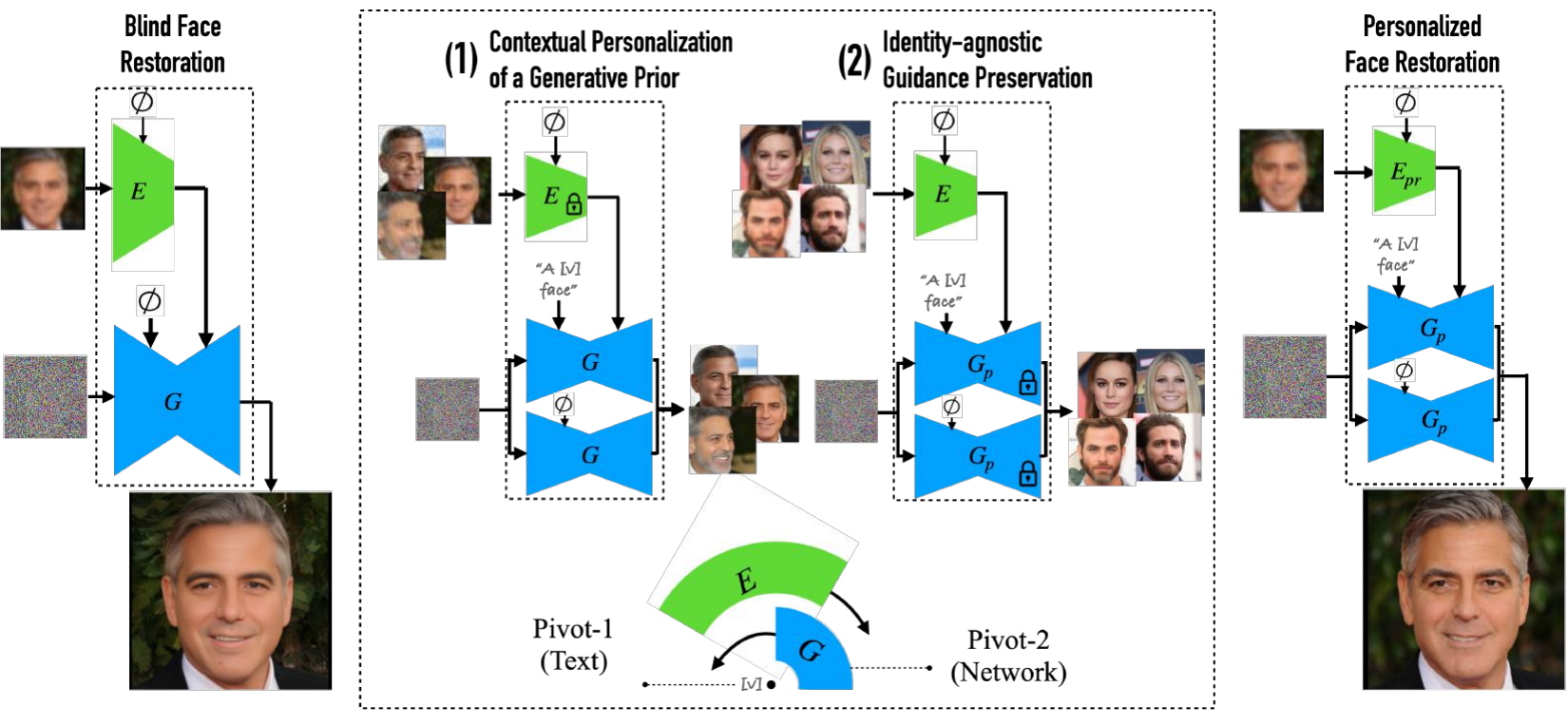}
    \caption{\textbf{High-level overview.} Our approach aims to personalize a blind face restoration system (left) and consists of two main steps: (1) text-anchored personalization of the generative prior $\bf{G}$ within the context of the system, and (2) adjusting $\mathbf{E}$ in retarget the encoder $\mathbf{E}$ in the presence of the strong personalized prior in $\mathbf{G}_p$, while guiding the network. Then, at inference time (3), our system embeds the personalized prior and can generate output images with high fidelity to the individual appearing in the reference images. 
    }
    \label{fig:scheme}
\end{figure*}

In this section, we outline our method for personalizing guided diffusion models. We start by recapitulating the personalization process for text-to-image models and the application of diffusion priors in blind image restoration. Subsequently, we present our dual-pivot tuning: firstly, employing text-based fine-tuning to embed identity-specific information within diffusion priors, and secondly, the necessity of model-centric pivoting to harmonize the guiding image encoder with the integrated personalized priors. This is followed by an exposition on how our method's inherent characteristics facilitate accelerated per-identity personalization. A high-level summary of the main idea behind our approach is illustrated in~\cref{fig:scheme}.

\subsection{Personalizing Text-Guided Diffusion Models}
\label{subsec:DB}

Given a text prompt $\mathbf{p}$, text-guided denoising diffusion models learn to sequentially denoise samples of random noise $\epsilon ~\sim~\mathcal{N}(\mathbf{0},\mathbf{I})$ into samples of images $\bf{x}$. 
During training, a neural network $\epsilon_\theta(\mathbf{x}_t,\mathbf{p})$
is trained to predict $\epsilon$ from a noisy version of the image $\mathbf{x}_t=\alpha_t \mathbf{x}+\sigma_t \epsilon$, where $\alpha_t$ and $\sigma_t$ are noise scheduling parameters, and $t$
refers to the time step in the diffusion process. 

A common way to sample the network at inference time is by using the classifier-free guidance technique~\cite{ho2022classifier}, that sums up a conditional instance of the model together with an unconditional one:
\begin{equation}
    \mathbf{G}_\theta(\mathbf{x},\mathbf{p}) = (1+w)\epsilon_\theta(\mathbf{x}_t, \mathbf{p}) - w\epsilon_\theta(\mathbf{x}_t, \varnothing),
    \label{eq:cfg}
\end{equation}
where \(w \) is a guidance scale, and $\varnothing$ represents a null-text prompt. These conditional and unconditional branches aim to strike a balance between prompt-fidelity and diversity within the generated images.

\begin{figure}
    \centering
    \begin{subfigure}[c]{0.32\linewidth}
         \centering   
         \includegraphics[width=\linewidth]{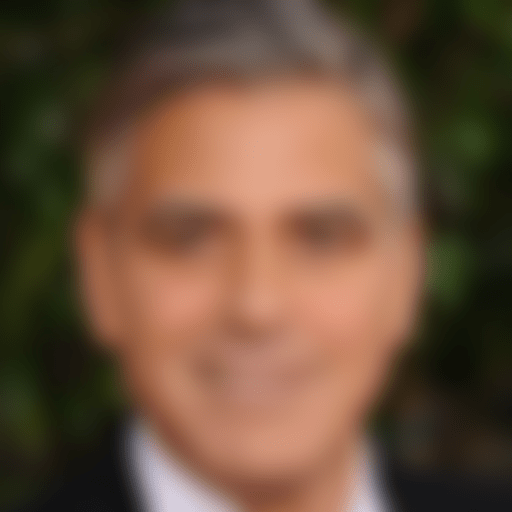}
         \caption{Input}
    \end{subfigure}  
    \begin{subfigure}[c]{0.32\linewidth}
         \centering   
         \includegraphics[width=\linewidth]{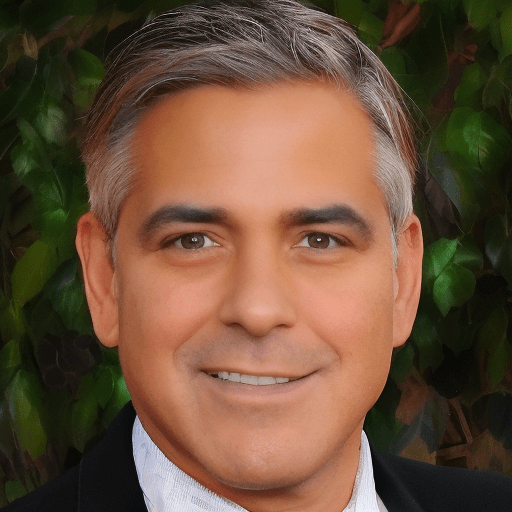}
         \caption{Seed 1}
    \end{subfigure} 
    \begin{subfigure}[c]{0.32\linewidth}
         \centering   
         \includegraphics[width=\linewidth]{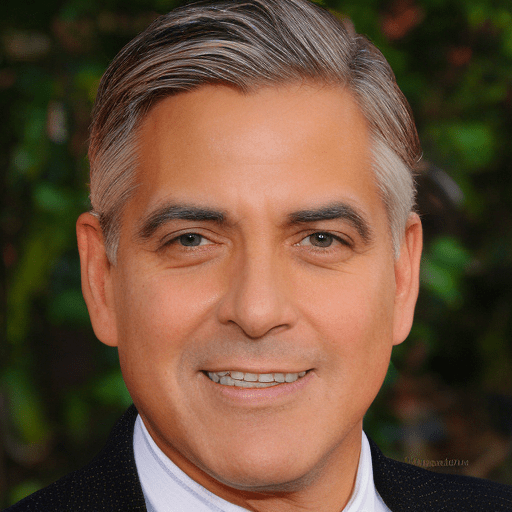}
         \caption{Seed 2}
    \end{subfigure} \\

    \begin{subfigure}[c]{0.32\linewidth}
         \centering   
         \includegraphics[width=\linewidth]{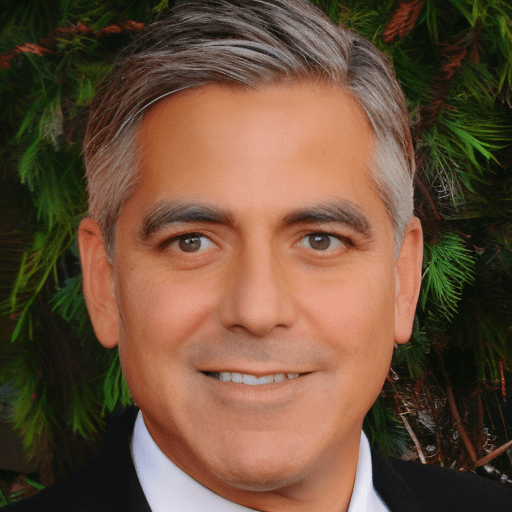}
         \caption{Seed 3}
    \end{subfigure}  
    \begin{subfigure}[c]{0.32\linewidth}
         \centering   
         \includegraphics[width=\linewidth]{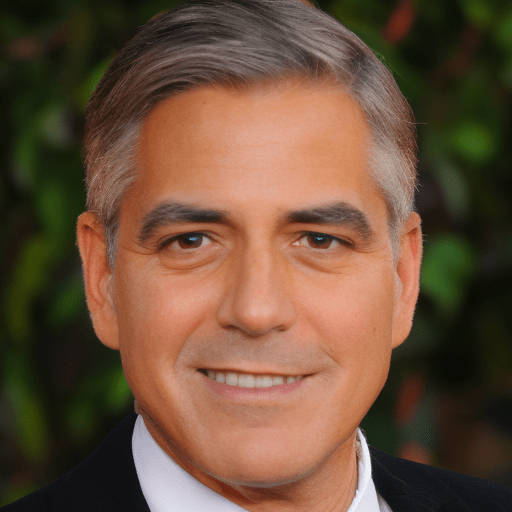}
         \caption{Seed 4}
    \end{subfigure} 
    \begin{subfigure}[c]{0.32\linewidth}
         \centering   
         \includegraphics[width=\linewidth]{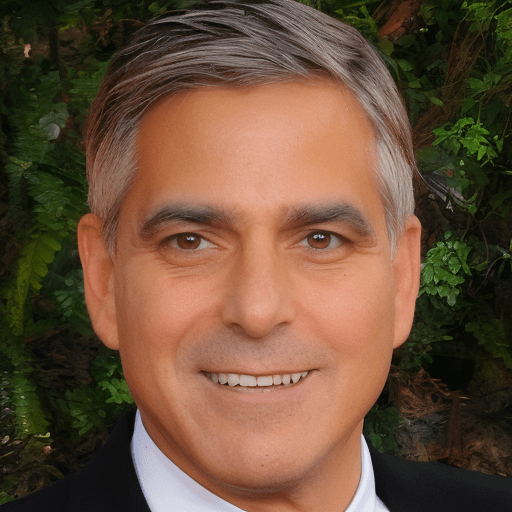}
         \caption{Seed 5}
    \end{subfigure} \\

    \caption{\textbf{Identity variation.} Given a fixed degraded input image, it can be seen that different seeds applied to ~\cite{lin2023diffbir} result in images with different identities. Please zoom in closer to observe differences, especially in terms of eyes, teeth and so on.}
    \label{fig:id_change}
    \vspace{-4mm}
    
\end{figure}

Recent work by~\citet{ruiz2023dreambooth} introduced the ability to personalize text-guided diffusion models using a few reference images of a subject, where the personalization is performed by fine-tuning the model around a text anchor $p$ (typically designed as a ``rare token"+``class noun", such as \textit{'a [v] dog'}), so the identity of the subject can be embedded into the diffusion prior, and utilized once $\mathbf{p}$ appears in the condition during inference. 
In practice, the weights, $\theta$, are optimized by minimizing
\begin{multline}
    \mathcal{L}_\text{DB}=\mathbb{E}_{\mathbf{x},\mathbf{p},\epsilon,\epsilon',t}\left[\lvert\lvert  \mathbf{G}_{\theta}(\mathbf{x}_t,\mathbf{p})-\epsilon\lvert\lvert_2^2 \right. \\
    \left. +\lvert\lvert  \mathbf{G}_{\theta}(\mathbf{x}^\text{pr}_{t},\mathbf{p}^{\text{pr}})-\epsilon'\lvert\lvert_2^2\right],
\end{multline}
where $\mathbf{x}^\text{pr}_{t}$ is drawn from a separate, prior preservation dataset, and $\mathbf{p}^{\text{pr}}$ is the prior prompt (``class noun", such as \textit{'a dog'}). This structure allows the optimization to revolve around a text pivot $\mathbf{p}$, without ruining the general prior of the model~\cite{roich2022pivotal}.

\pc{
We define the personalization operator of text-to-image diffusion models by
\begin{equation}
    \mathbf{M}_\mathcal{W} = \mathcal{P}\{\mathbf{M}, \mathcal{W} \},
\end{equation}
where $\mathcal{P}\{\cdot\}$ is the operator, and $\mathbf{M}_\mathcal{W}$ is the personalized version of a model $\mathbf{M}$ which is fine-tuned with $\mathcal{W}$ as a pivot. In our case, this pivot can be both a text prompt or a network, as discussed subsequently.
}

\subsection{Diffusion-Guided Image Restoration}
Recent advances in blind face image restoration~\cite{lin2023diffbir,wang2023exploiting} have attempted to utilize a pretrained diffusion model (such as Stable Diffusion~\cite{rombach2022high}) as generative prior to guide the restoration process. Commonly, these pipelines contain a diffusion-guided image restoration step that consist of two components: (1) A general image prior in the form of text-to-image diffusion model $\mathbf{G}(\cdot)$, that encodes the manifold of natural images and (2) an encoder $\mathbf{E}(\cdot)$, that captures context information from the degraded image and guides the generation process such that it maintains high fidelity to the visual attributes of the degraded image. 
Guiding the diffusion process with spatial features extracted from an image encoder $\textbf{E}$ is a common way to control the diffusion processes~\cite{zhang2023adding}. It should be noted that in case of blind restoration, the general prior $\textbf{G}$ plays a key role in dictating the identity of the person. As demonstrated in ~\cref{fig:id_change}, for a given degraded input image (namely, fixed set of guiding features that are extracted from $\textbf{E}$), different seeds in the input of $\mathbf{G}(\cdot)$, lead to different identities in the output.

Our goal in this work is to personalize the general prior $\textbf{G}$ such that the restored face maintains high fidelity to the identity of the individual without degrading the integrity of the general image prior and the distinct roles of each component in the system.

We consider a general setting, where the encoder $\mathbf{E}(\cdot)$ takes a low-quality image as an input and provides guidance for the diffusion model. 
\pc{
That is, the diffusion model receives conditioning from the encoder as well, such that
\begin{equation}
    \hat{I} = \mathbf{G}(\epsilon,\mathbf{p},\mathbf{E}(I_{LQ})),
\end{equation}
where $I_\text{LQ}$ is a low quality image as input and $\mathbf{p}$ is the text conditioning. Note that $I_\text{LQ}$ could be a degraded image~\cite{wang2023exploiting}, or the output of a preliminary restoration model, like in~\cite{lin2023diffbir}. 
In the context of the blind image restoration setting, we will use the following notation:
\begin{equation}
    \hat{I} = \mathbf{B}(\mathbf{G}(\epsilon,\varnothing),\mathbf{E}(I_\text{LQ})),
\end{equation}
where $\mathbf{B}$ represents the blind restoration system. In this blind restoration setting, the text conditioning is null (as in~\cite{lin2023diffbir}).
}

\subsection{Dual-Pivot Tuning}

\paragraph{Step-1 - Textual Pivoting.}
Theoretically, personalization of a generative prior can be decoupled from the context of the system. In our case, we can personalize $\textbf{G}$ with a pivot text prompt $p$ independently of the blind restoration system and plug it back to the framework, namely,
\begin{equation}
    \mathbf{B}_p = \{\mathbf{E},\mathbf{G}_p\},
\end{equation}
where $\mathbf{G}_p =\mathcal{P}\{\mathbf{G},p\}$.
However, as demonstrated in \cref{fig:method_motivation}(c), if we do so, the restored images are unable to imbibe identity information and the outputs resemble the images generated with $\textbf{G}$, the non-personalized version of the model (\cref{fig:method_motivation}(b)). 

The occurrence of this issue can be attributed to the initial training of $\mathbf{E}$, where it was coupled with the unconditional branch of $\mathbf{G}$. As a result, $\mathbf{E}$ cannot exert influence on the conditional branch of $\mathbf{G}$, which is responsible to contribute the personalized features around the anchor prompt $p$.

We therefore find that for effective personalization of face image restoration, text-based pivoting alone is not enough, and that $\mathbf{E}$ must be fine-tuned in conjunction with $\mathbf{G}_p$.

\paragraph{Step-2 - Model-based Pivoting.}
We are introducing a second fine-tuning step for $\mathbf{E}$ within the context of $\mathbf{G}_p$. Essentially, our objective is to fine-tune $\mathbf{E}$ in a way that it relinquishes identity cues to the personalized prior in $\mathbf{G}_p$ while focusing on other detail cues. In this scenario, our pivot is not a text but a network ($\mathbf{G}_p$), which we keep fixed during the optimization process to align the encoder with the personalized prior.

To preserve the identity-agnostic role of the guiding encoder, we intentionally aim to avoid personalizing it. One approach to achieve this is by updating $\mathbf{E}$ across different individuals, allowing it to adapt to the conditional branch of $\mathbf{G}_p$ while remaining agnostic to fine-grained identity features, which should be determined by $\textbf{G}_p$, as demonstrated in ~\cref{fig:id_change}. This identity-agnostic retargeting concept bears similarities to the Prior Preservation loss~\cite{ruiz2023dreambooth} in the context of pivotal-tuning around a model.

Mathematically, with $\mathbf{G}_p$ as our starting point, we perform fine-tuning on $\mathbf{E}$ as follows:
\pc{
\begin{equation}
\mathbf{B}_p = \mathcal{P}
     \{\mathbf{B}_{\textbf{E}}=\{\mathbf{G}_p,\bf{E}\},\mathbf{G}_p\},
\end{equation}
}
where $\mathbf{B}_{\textbf{E}}$ represents the restoration system (with the embedded personalized prior), and we are now optimizing the weights of $\textbf{E}$.

\subsection{Speeding-Up Dual-Pivot Tuning}
In general, personalization of generative models is a time-consuming endeavor. Given our dual pivot tuning approach, personalization involves two finetuning steps per identity: textual pivoting, followed by model-based pivoting. Hence we next suggest steps to significantly reduce (by about 2x) fine-tuning time of each pivot tuning. 

\paragraph{Speeding up textual pivoting.}
The presence of both conditional and unconditional branches within our inference pipeline affords an additional unique opportunity. We noticed that the initial steps of the restoration do not rely on identity as much, as these steps largely focus on coarse geometry and semantic detail~\cite{voynov2023p+}. Therefore, during inference, we observe that high-quality, identity-preserving restoration can be achieved even when the initial denoising steps are only unconditional, using a guidance scale of 0, followed by text-guided denoising for later steps with a non-zero guidance scale. This is mathematically represented as follows:
\begin{equation}
    \begin{split}
        \epsilon_{t+1}=&\mathbf{B}_p(\mathbf{x}_t,\varnothing)\,\, | \,\,t<\gamma,\\
        \epsilon_{t+1}=w\cdot\mathbf{B}_p(\mathbf{x}_t,p)+&(1-w)\mathbf{B}_p(\mathbf{x}_t,\varnothing)\,\,|\,\,\gamma\leq t<1.0,
    \end{split}
\end{equation}
where $\gamma$ denotes the fraction of denoising steps for which unconditional inference is carried out. Note that in the above expression, we denote the initial noise to be at $t=0$ and the restored image to be at $t=1$.

\begin{figure}
    \centering
    \begin{subfigure}[c]{0.24\linewidth}
         \centering   
         \includegraphics[width=\linewidth]{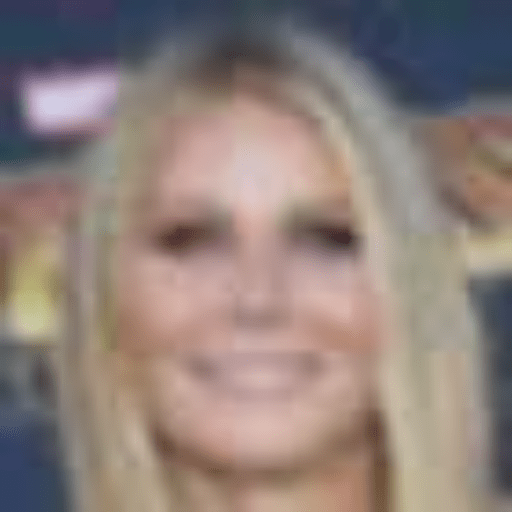}
         \caption{}
    \end{subfigure}  
    \begin{subfigure}[c]{0.24\linewidth}
         \centering   
         \includegraphics[width=\linewidth]{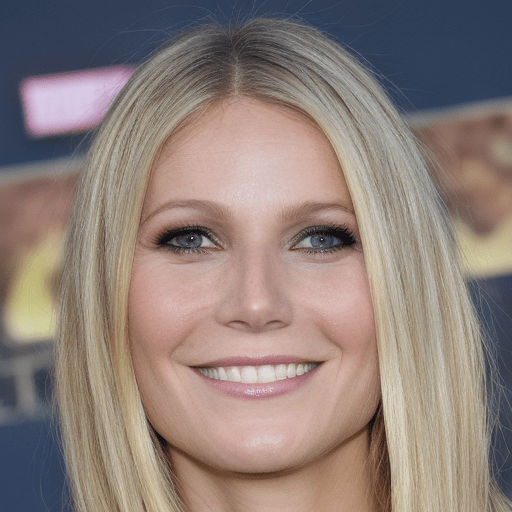}
         \caption{}
    \end{subfigure} 
    \begin{subfigure}[c]{0.24\linewidth}
         \centering   
         \includegraphics[width=\linewidth]{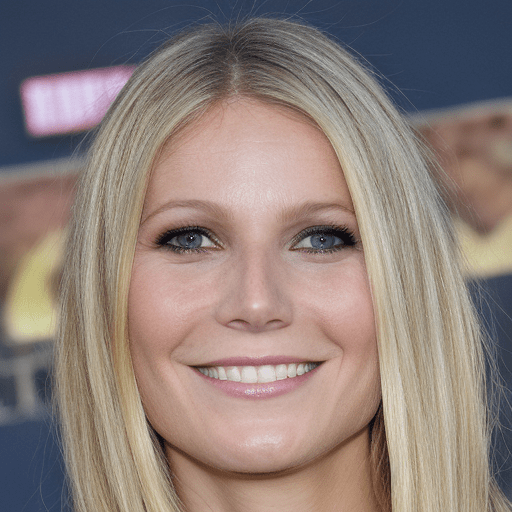}
         \caption{}
    \end{subfigure} 
    \begin{subfigure}[c]{0.24\linewidth}
         \centering   
         \includegraphics[width=\linewidth]{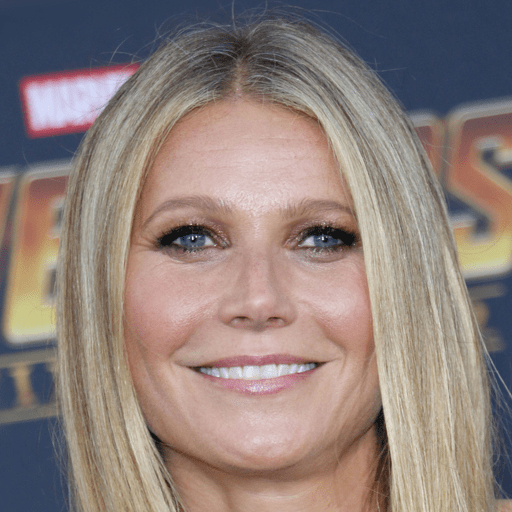}
         \caption{}
    \end{subfigure} 

    \caption{\textbf{Unconditional Sampling.} We found that during inference, identity-preserved restoration is possible even with unconditional denoising for initial steps, followed by conditional denoising for the remaining steps. For a degraded input (a), we use unconditional sampling for the first 50\% of steps, followed by conditional sampling for the remaining steps (c). We find this to still be consistent with the reference image (d), both in terms of identity as well as faithfulness to the original image, as well as with the restored image using conditional restoration for all the steps (b).}
    \label{fig:uncond_cond_infer}
\end{figure}

We demonstrate this in \cref{fig:uncond_cond_infer}, where we restore the degraded image (a) unconditionally for the first $\gamma = 50\%$ of steps, and conditionally for the remaining steps (c). Despite this, we find the restored image being faithful to the reference image (d), as well as to the restored image obtained through conditional restoration for all the steps (b). $\gamma = 50\%$ is identified experimentally. This observation enables us to reduce personalization time for the textual pivoting by about 2x since the denoiser no longer needs to be personalized for higher noise levels (since the unconditional model suffices for those), and can, therefore, be trained with a focus on relatively lower noise levels.

\paragraph{Speeding up model-based pivoting.}
Coupling the generative prior $\mathbf{G}$ and the guiding encoder $\mathbf{E}$ during the textual pivoting step can provide enocder context to the personalized generative prior. To achieve this, we propose to fine-tune $\mathbf{G}$ within the context of $\mathbf{B}$, while pivoting around the prompt $p$, such that it leverages the conditioning cues from $\mathbf{E}$, namely,
\begin{equation}
    \mathbf{B}_p = \mathcal{P}\{\mathbf{B_G}, p\},
\end{equation}
where $\mathbf{B_G}$ denotes that we modify the weights of $\mathbf{G}$ within the context of $\mathbf{B}$. Specifically, this personalization pivots around the text prompt $p$ in the context of image restoration. The performance of this so-called in-context textual pivoting, and comparison with the previously discussed out-of-context textual pivoting is shown in \cref{fig:variants_supp} and discussed in detail later: while in-context textual pivoting improves identity information in the restored images, model-based pivoting is still required to retarget the encoder.

With the personalization of the generative prior being in-context in our proposed pipeline, we find that the model-based pivoting is feasible with half the number of finetuning steps, when compared with an out-of-context textual pivoting pipeline (\cref{fig:variants_supp}). On an NVIDIA Tesla V100 GPU, this leads to a model-based pivoting time of about 10 minutes, as compared to 20 minutes in the case of out-of-context textual pivoting.

\begin{figure}
    \centering
    \begin{subfigure}[c]{0.24\linewidth}
         \centering   
         \includegraphics[width=\linewidth]{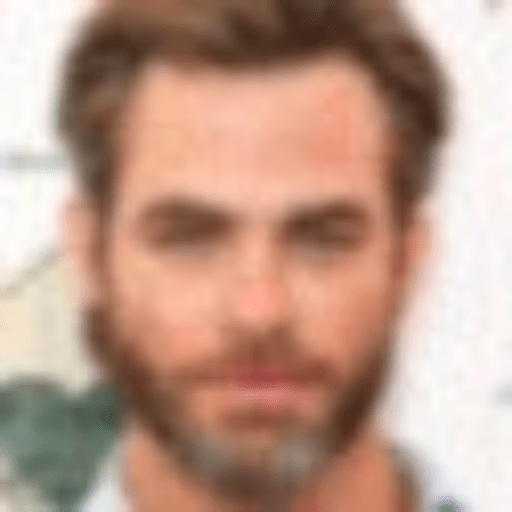}
         \caption{Input}
    \end{subfigure}  
    \begin{subfigure}[c]{0.24\linewidth}
         \centering   
         \includegraphics[width=\linewidth]{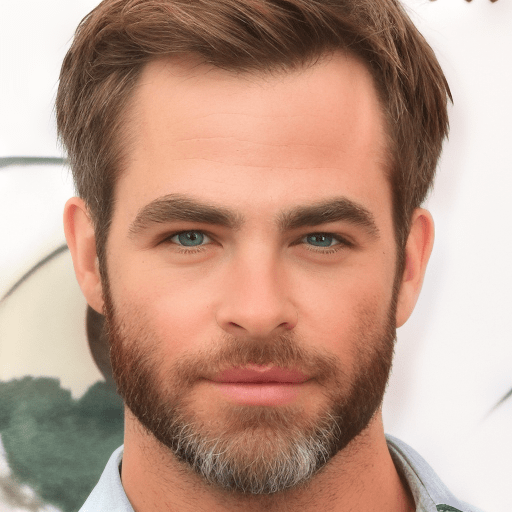}
         \caption{$\mathbf{E}$, same id.}
    \end{subfigure} 
    \begin{subfigure}[c]{0.24\linewidth}
         \centering   
         \includegraphics[width=\linewidth]{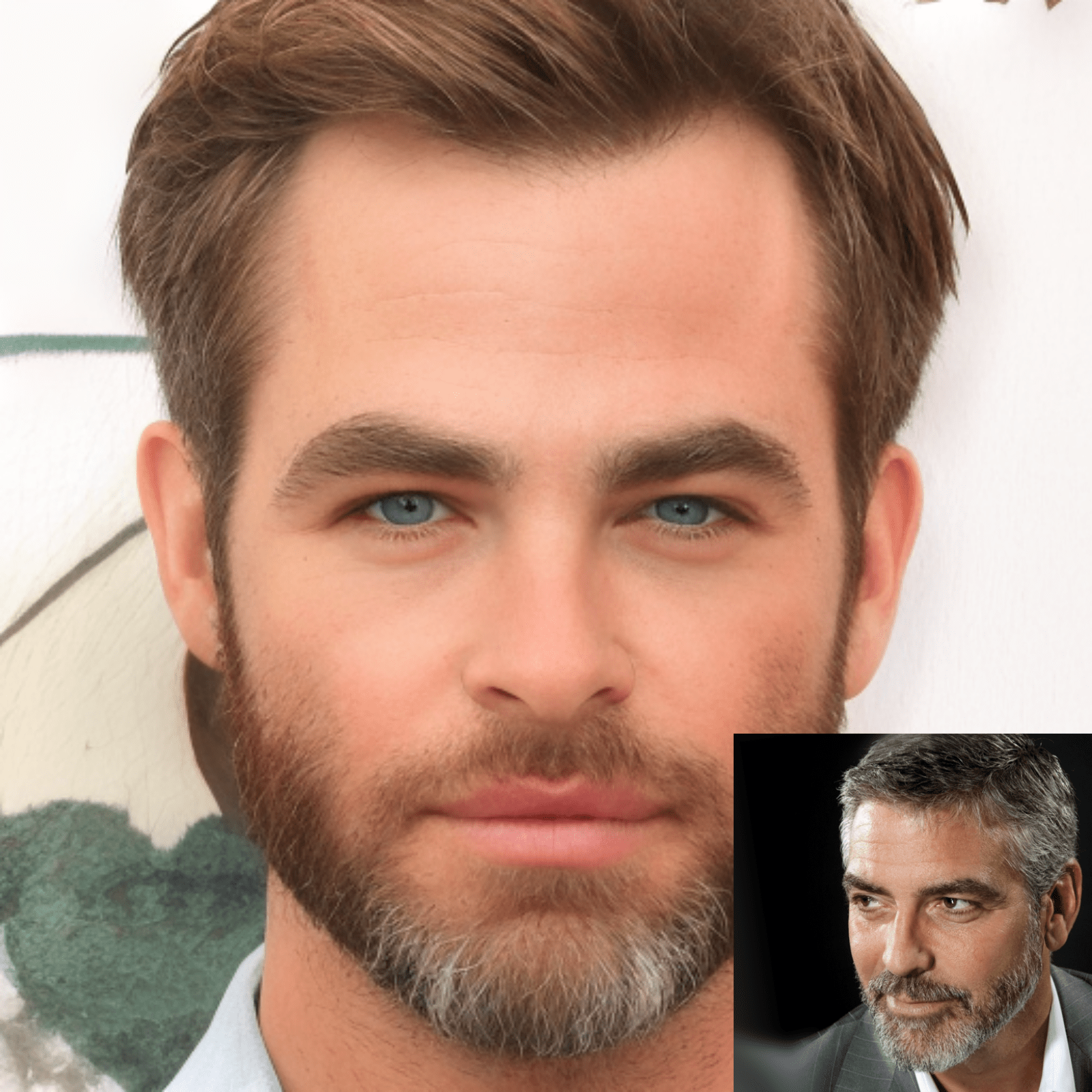}
         \caption{$\mathbf{E}$, diff. id.}
    \end{subfigure} 
    \begin{subfigure}[c]{0.24\linewidth}
         \centering   
         \includegraphics[width=\linewidth]{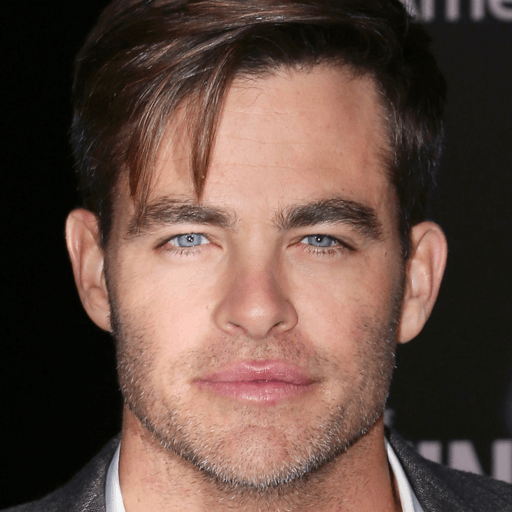}
         \caption{Id. Reference}
    \end{subfigure} 

    \caption{\textbf{Finetuning $\mathbf{E}$ on different identities.} We show that for the model-based pivot tuning $\mathbf{E}$ can be finetuned on the same identity (b), as well as on different identities as in the inset (c), while providing similar plausible restorations with respect to the identity in the reference image (d). } 
    \label{fig:encoder_swap}
\end{figure}

In practice, we have observed that the personalized prior in $\mathbf{G}_p$ is sufficiently strong, such that even when performing the model-based pivoting of $\mathbf{E}$ on a single individual, $\mathbf{E}$ is retargeted to capture general cues and can be shared across various identities, as demonstrated in~\cref{fig:encoder_swap}. This is critical from a training time perspective: the model-based pivoting, once performed on one identity, can be shared across models for different identities. Therefore, the only personalization step needed for each identity is the textual pivoting.

\section{Experiments}
\label{sec:experiments}

\begin{figure*}[t]
  \centering
   \includegraphics[width=\linewidth]{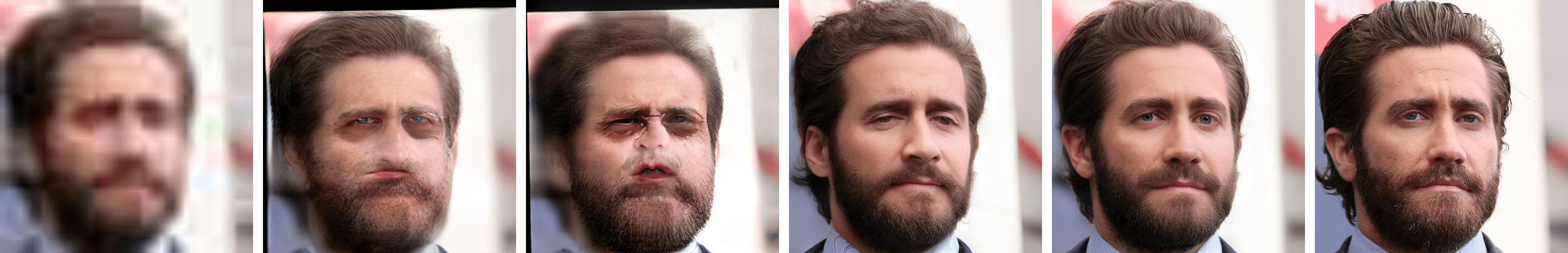}\\
   \includegraphics[width=\linewidth]{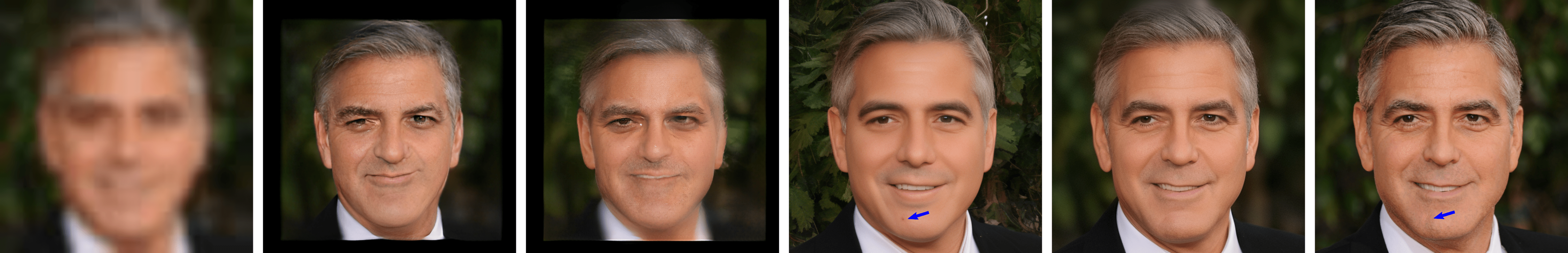}\\
   \includegraphics[width=\linewidth]{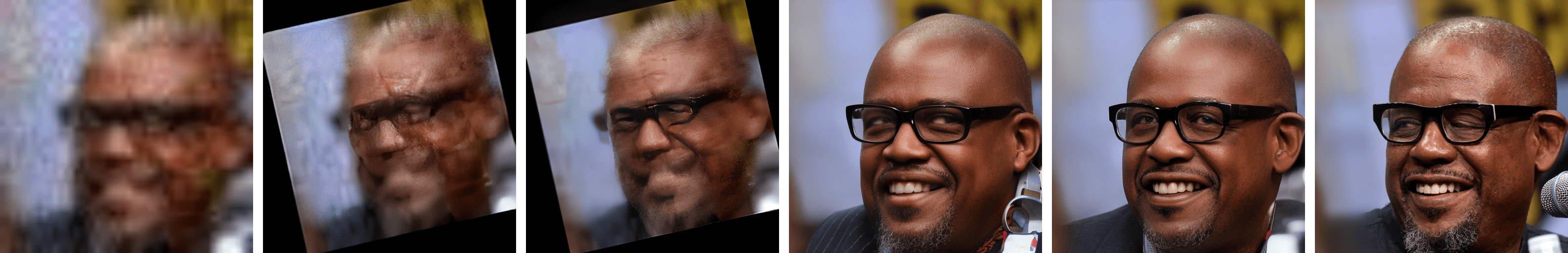}\\
   \includegraphics[width=\linewidth]{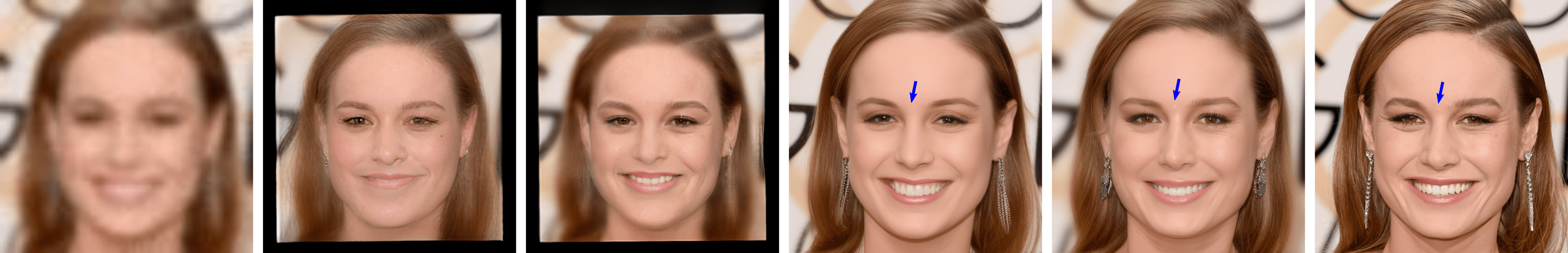}\\
   \includegraphics[width=\linewidth]{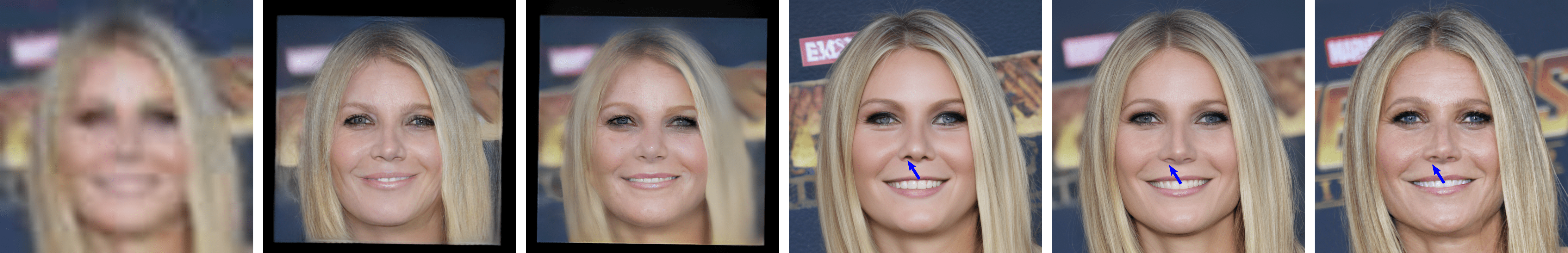}\\
   \includegraphics[width=\linewidth]{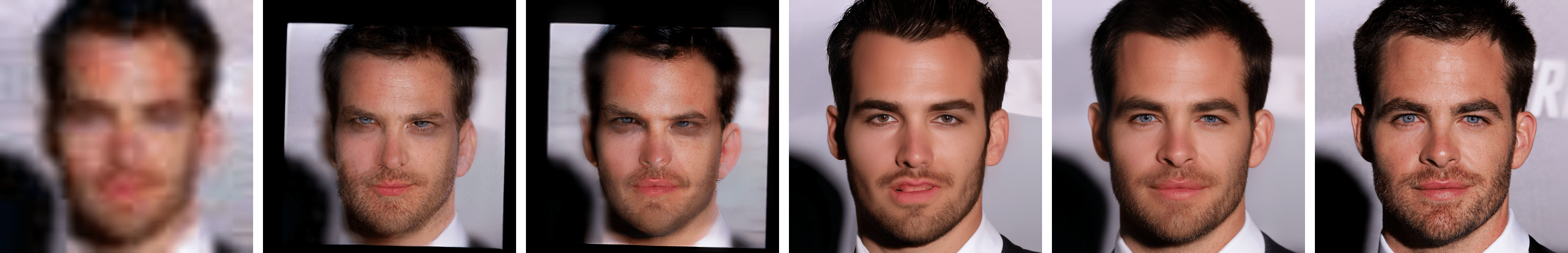}\\

   \begin{subfigure}[c]{0.16\textwidth}
        \centering    \caption{\footnotesize{\textsc{Degraded Image}}}
    \end{subfigure}
    \begin{subfigure}[c]{0.16\textwidth}
        \centering    \caption{\footnotesize{\textsc{ASFFNet~\cite{Li_2020_CVPR}}}}
    \end{subfigure}
    \begin{subfigure}[c]{0.16\textwidth}
        \centering    \caption{\footnotesize{\textsc{DMDNet~\cite{li2022learning}}}}
    \end{subfigure}
    \begin{subfigure}[c]{0.16\textwidth}
         \centering    \caption{\footnotesize{\textsc{DiffBIR~\cite{lin2023diffbir}}}}
    \end{subfigure}
    \begin{subfigure}[c]{0.16\textwidth}
         \centering    \caption{\footnotesize{\textsc{Ours}}}
    \end{subfigure}
    \begin{subfigure}[c]{0.16\textwidth}
         \centering    \caption{\footnotesize{\textsc{Ground Truth}}}
    \end{subfigure}

   \caption{\textbf{Results on synthetically degraded images.} A considerable shift in identity can be observed between our proposed method and alternative baselines. Identity preservation can be observed in terms of overall geometric features, as well as attributes: eye shape (row 1), unnatural mole on the chin in the DiffBIR result (row 2), teeth structure and shape (row 3), eye color and feature between eyes (row 4), nose and nostrill shape (row 5), eye color (row 6).}
   \label{fig:synth_deg1}
\end{figure*}

\begin{figure*}[t]
  \centering

   \includegraphics[width=0.16\linewidth]{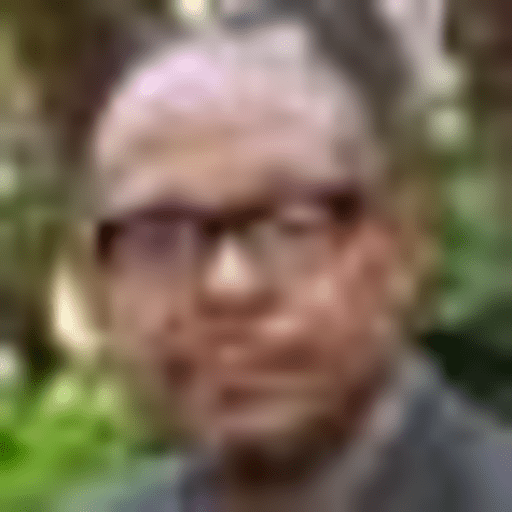}
   \includegraphics[width=0.16\linewidth]{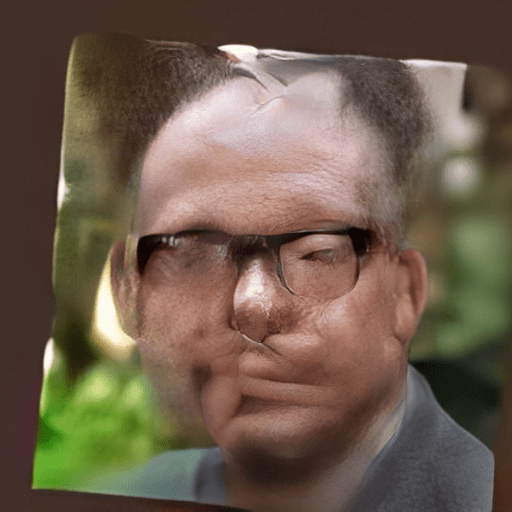}
   \includegraphics[width=0.16\linewidth]{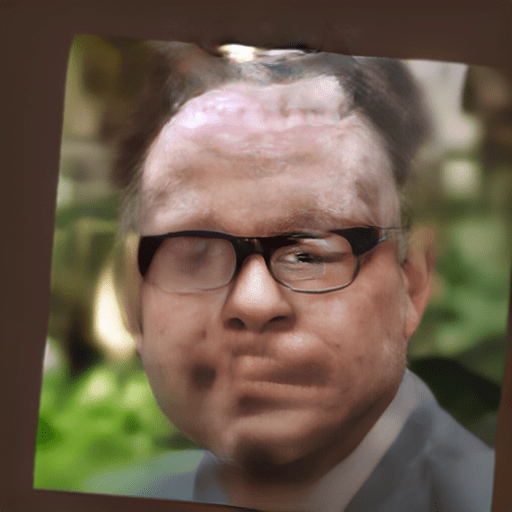}
   \includegraphics[width=0.16\linewidth]{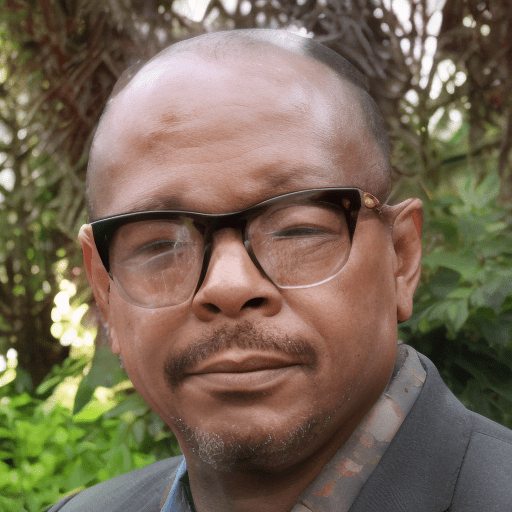}
   \includegraphics[width=0.16\linewidth]{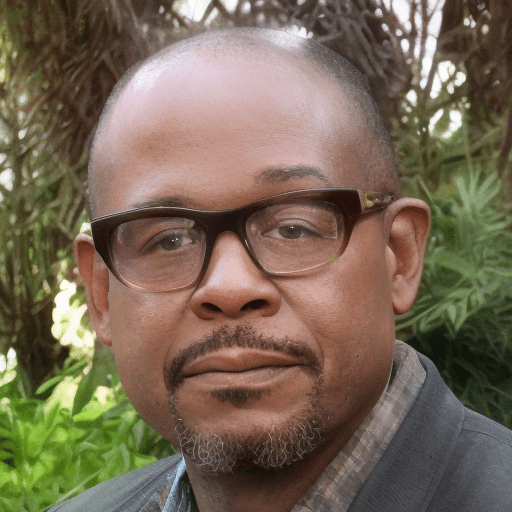}
   \includegraphics[width=0.16\linewidth]{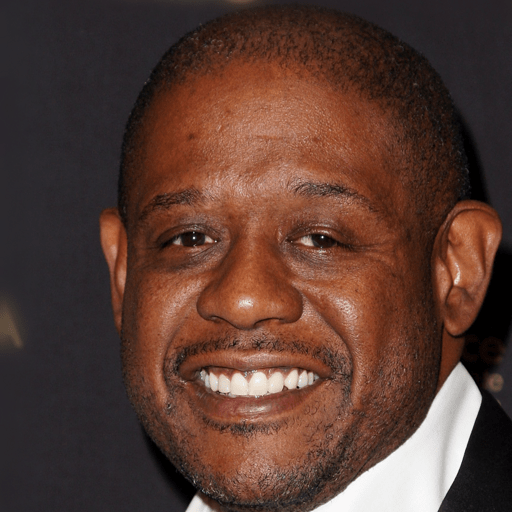}\\

    \includegraphics[width=0.16\linewidth]{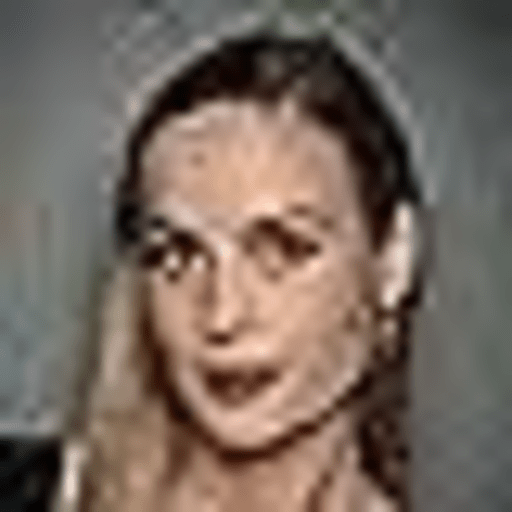}
    \includegraphics[width=0.16\linewidth]{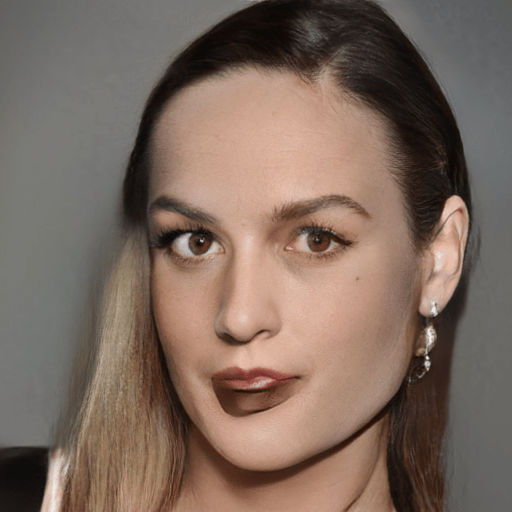}
   \includegraphics[width=0.16\linewidth]{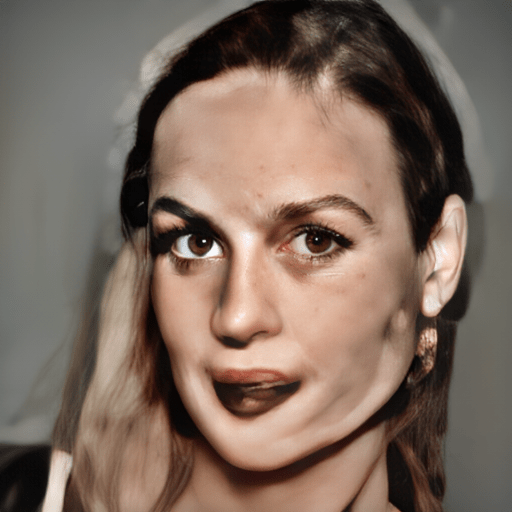}
   \includegraphics[width=0.16\linewidth]{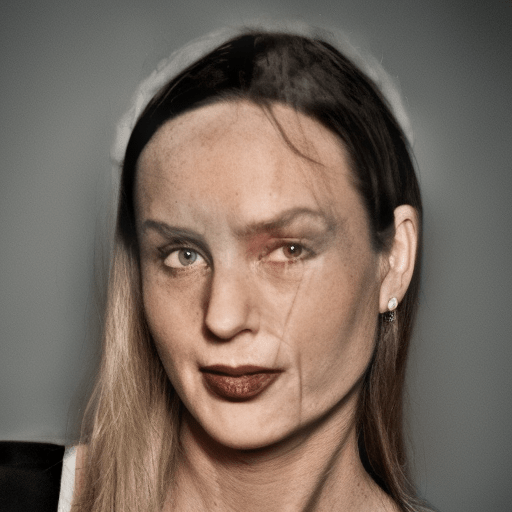}
   \includegraphics[width=0.16\linewidth]{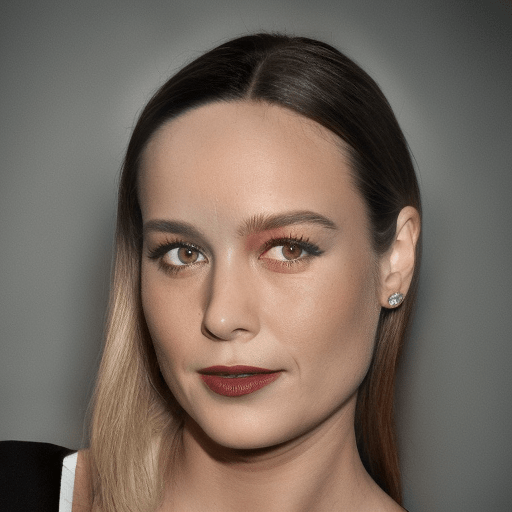}
   \includegraphics[width=0.16\linewidth]{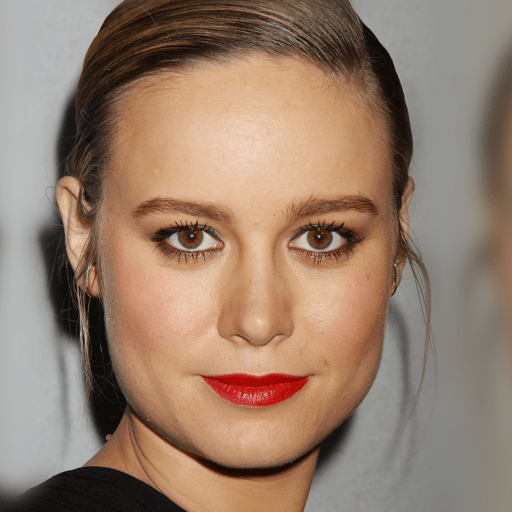}\\

   \includegraphics[width=0.16\linewidth]{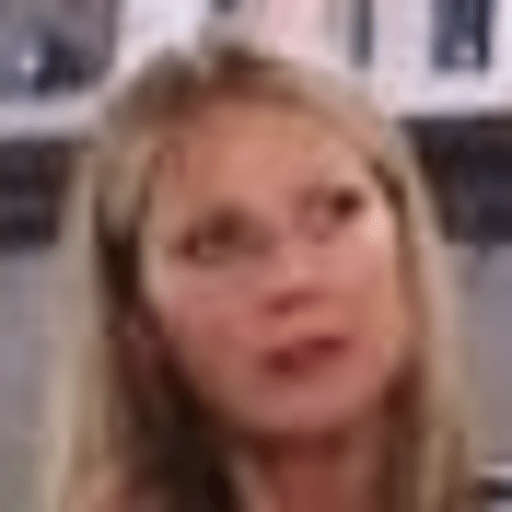}
    \includegraphics[width=0.16\linewidth]{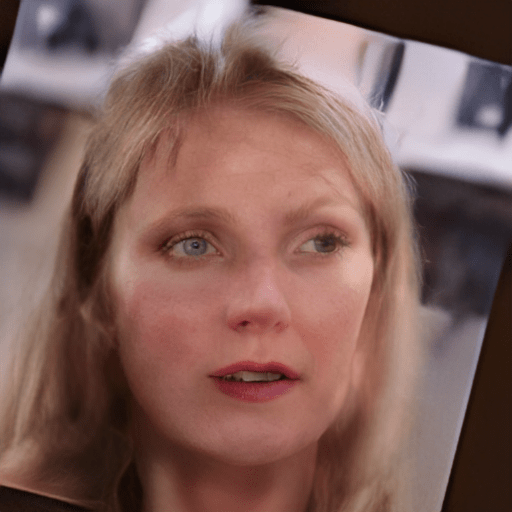}
   \includegraphics[width=0.16\linewidth]{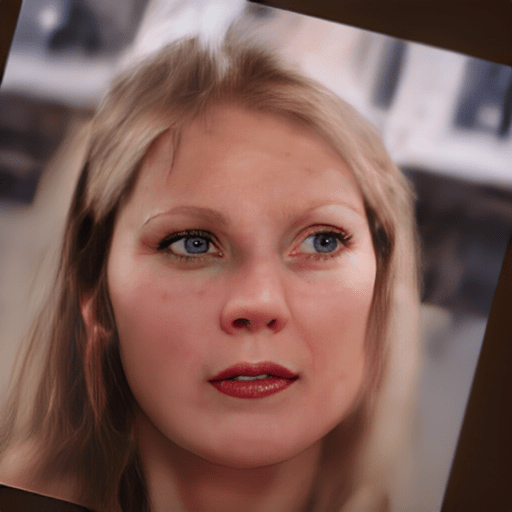}
   \includegraphics[width=0.16\linewidth]{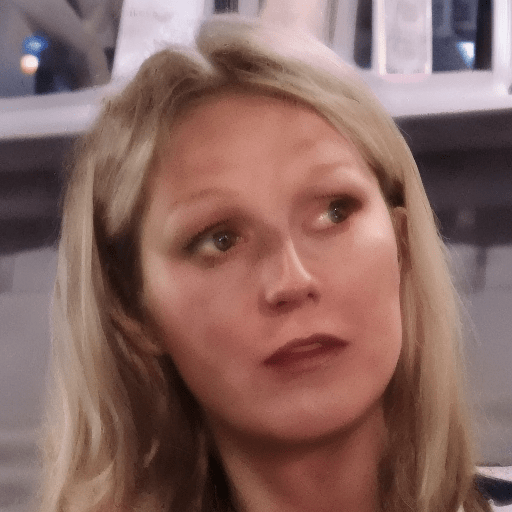}
   \includegraphics[width=0.16\linewidth]{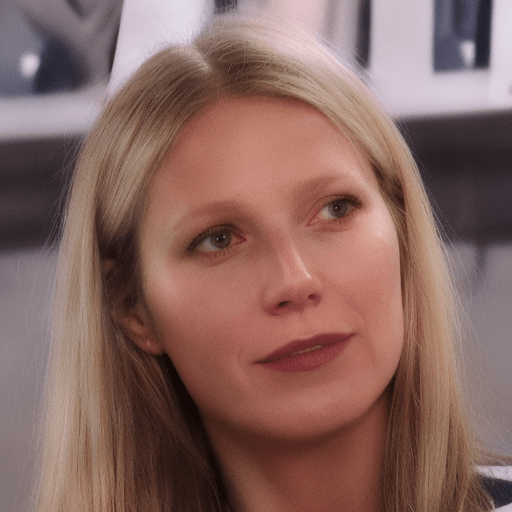}
   \includegraphics[width=0.16\linewidth]{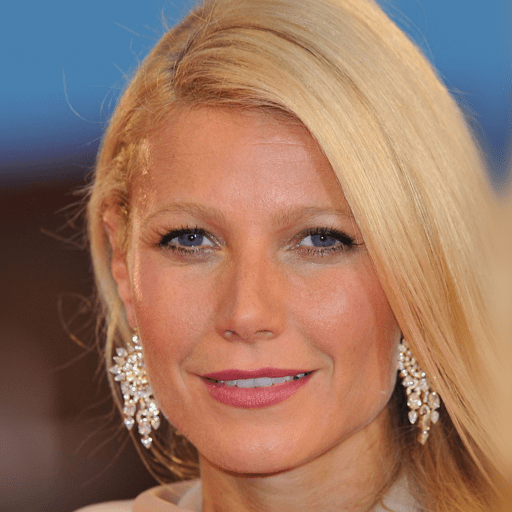}\\

   \begin{subfigure}[c]{0.16\textwidth}
        \centering    \caption{\footnotesize{\textsc{Degraded Image}}}
    \end{subfigure}
    \begin{subfigure}[c]{0.16\textwidth}
        \centering    \caption{\footnotesize{\textsc{ASFFNet~\cite{Li_2020_CVPR}}}}
    \end{subfigure}
    \begin{subfigure}[c]{0.16\textwidth}
        \centering    \caption{\footnotesize{\textsc{DMDNet~\cite{li2022learning}}}}
    \end{subfigure}
    \begin{subfigure}[c]{0.16\textwidth}
         \centering    \caption{\footnotesize{\textsc{DiffBIR~\cite{lin2023diffbir}}}}
    \end{subfigure}
    \begin{subfigure}[c]{0.16\textwidth}
         \centering    \caption{\footnotesize{\textsc{Ours}}}
    \end{subfigure}
    \begin{subfigure}[c]{0.16\textwidth}
         \centering    \caption{\footnotesize{\textsc{Id. Reference}}}
    \end{subfigure}

   \caption{\textbf{Results on real degraded images.} It can be seen that even in images in the wild with real, unknown degradation kernels, our proposed method is superior to the baselines in terms of identity retention while maintaining high fidelity to the degraded input.
   }
   \label{fig:real_deg1}
\end{figure*}

In this section, we show through qualitative and quantitative analysis that the proposed method is superior to prior reference-based and blind image restoration methods.

\subsection{Training Process and Datasets}
We use images from CelebRef-HQ dataset \cite{li2022learning} to personalize the model; more specifically, this dataset has $>10$ $512\times 512$ images of the same person with $500$ identities, and each time we use one identity's images.
Our method is trained on synthetic data that covers a wide range of degradation similar to the real world. 

We follow the same second-order degradation strategy as \cite{lin2023diffbir}. At each stage we first convolve the image with a blur kernel $k_\sigma$, and downsample with a scale factor $r$. Following that, additive noise $n_\delta$ is added, and finally the image is JPEG-compressed with quality $q$. Formally, a single stage can be described as
\begin{equation}
    \mathbf{x}_d = [(\mathbf{x}\circledast k_{\sigma})\downarrow_{r} + {n}_{\delta}]_{\mathtt{JPEG}_{q}},
    \label{eq:degrade}
\end{equation}
where $\circledast$ is the convolution operator. The final image is obtained after applying \cref{eq:degrade} twice. We refer you to \cite{lin2023diffbir} for more details on the degradation process.

\myparagraph{Test data}
We did test on three datasets: (1) CelebRef-HQ test set, (2) Google searched images including low-quality and high-quality images from the same person, (3) our collected real data from acquaintance. 

\subsection{Comparisons with other strategies}
We show the comparisons of our personalization strategy to the alternative strategies in Fig.~\ref{fig:method_motivation}. As can be seen, the proposed strategy maximizes identity preservation through the restoration process, while comparison strategies, namely personalizing when pivoting only on the text condition (c) or personalizing when pivoting only on the generative model both are unable to effectively incorporate identity while retaining image fidelity.

\begin{table}[t]
    \caption{\textbf{Fidelity and idenitity metrics.} The proposed method generates images with high fidelity, while showing superior identity retention, when personalized on 10 images. We compare along PSNR and SSIM as fidelity metrics, while the ArcFace similarity serves as an identity metric.}
    \footnotesize
   \vspace{-4mm}
  \begin{center}
    \setlength{\tabcolsep}{0.01\columnwidth}
    
\resizebox{0.95\linewidth}{!}{
  \begin{tabular}{lccccc} 
    \toprule
    \textbf{Method} & \textbf{PSNR (dB)}  &
    \textbf{SSIM} &
    \textbf{ArcFace (Identity)}  \\
    \midrule
    GFPGAN~\cite{wang2021gfpgan} & 23.67 & {0.58} &  0.75\\
    CodeFormer~\cite{zhou2022towards} & {23.98} & 0.57 & 0.76\\
    MyStyle~\cite{nitzan2022mystyle} & 17.24 & 0.53 &   0.68\\
    DR2~\cite{wang2023dr2} & 21.56 & 0.54 &   0.40\\
    ASFFNet~\cite{Li_2020_CVPR} & 11.21 & 0.40 &  0.55\\
    DMDNet~\cite{li2022learning} & 11.33 & 0.40 & 0.61\\
    DiffBIR~\cite{lin2023diffbir} & \textbf{24.27} & \textbf{0.58} &   0.76\\
    \midrule
    \textbf{Ours} & 23.72 & 0.55 &  \textbf{0.88}\\
    \bottomrule
  \end{tabular} } 
 \end{center}
  \label{tab:emprical_analysis}
  \vspace{-8mm}
\end{table}

\subsection{Comparisons with SOTA methods}
We perform qualitative and quantitative comparisons between our method and several state-of-the-art reference-based and blind face image restoration methods. For reference-based methods, we chose ASFFNet \cite{li2020enhanced}, DMDNet \cite{li2022learning}, and MyStyle \cite{nitzan2022mystyle}; for a fair comparison, all methods use 10 reference images for guidance. For blind face image restoration, we chose DiffBIR \cite{lin2023diffbir}, DR2 \cite{wang2023dr2}, GFP-GAN \cite{wang2021gfpgan}, and CodeFormer \cite{zhou2022towards}. 
For ease of reading, we put qualitative results from MyStyle, DR2, GFP-GAN, and CodeFormer in the appendix.

\begin{figure*}[t]
  \centering
   \includegraphics[width=0.16\linewidth]{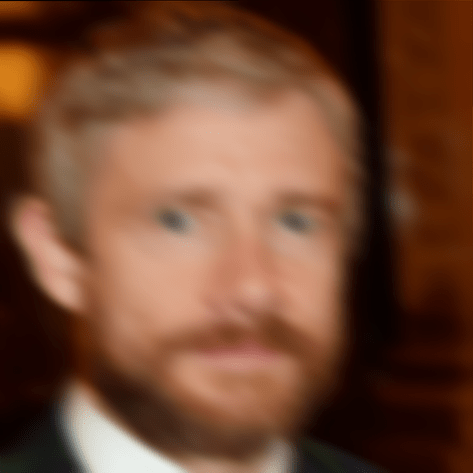}
   \includegraphics[width=0.16\linewidth]{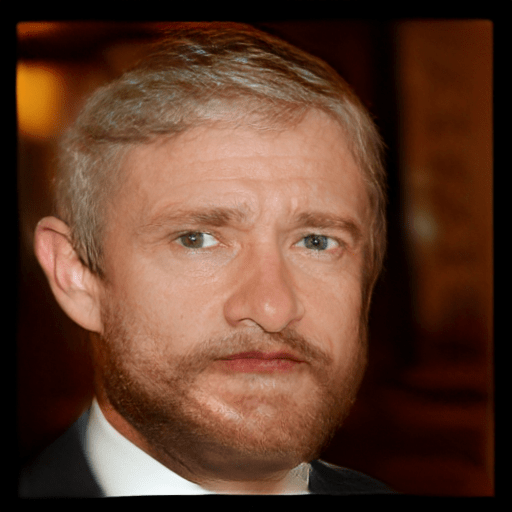}
   \includegraphics[width=0.16\linewidth]{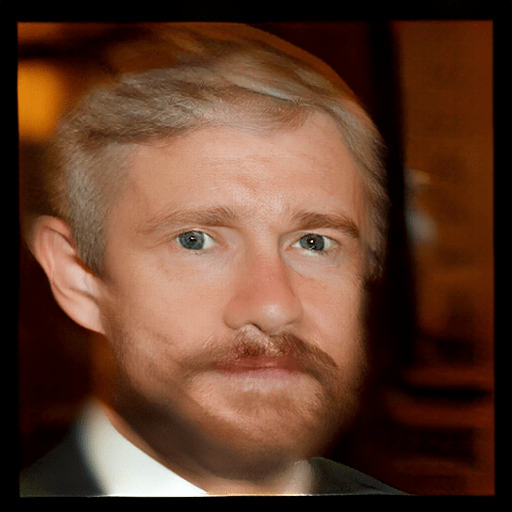}
   \includegraphics[width=0.16\linewidth]{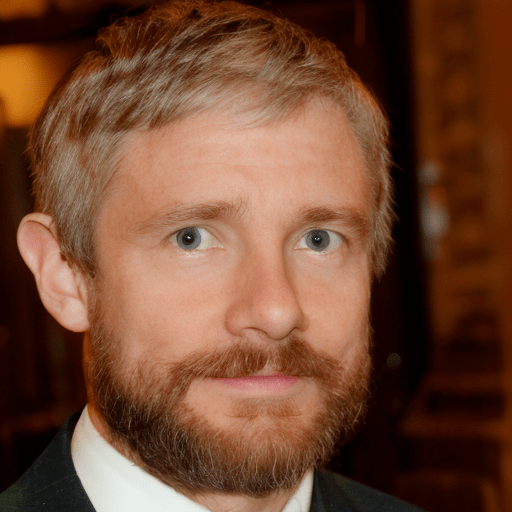}
   \includegraphics[width=0.16\linewidth]{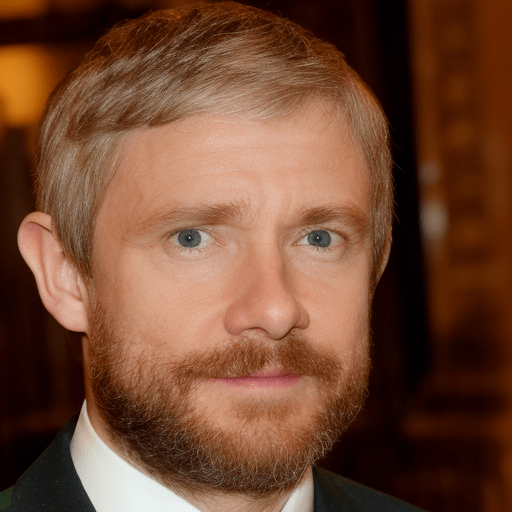}
   \includegraphics[width=0.16\linewidth]{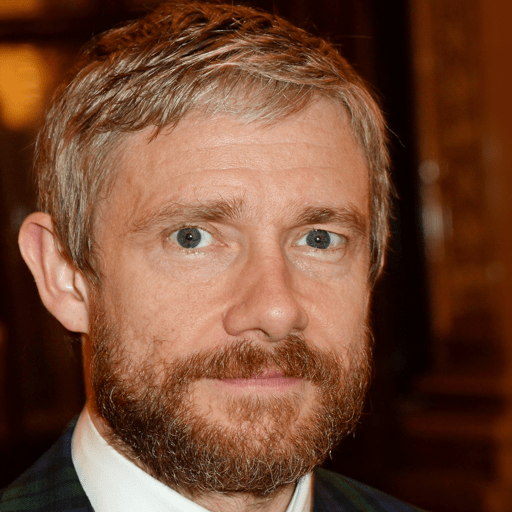}\\

    \includegraphics[width=0.16\linewidth]{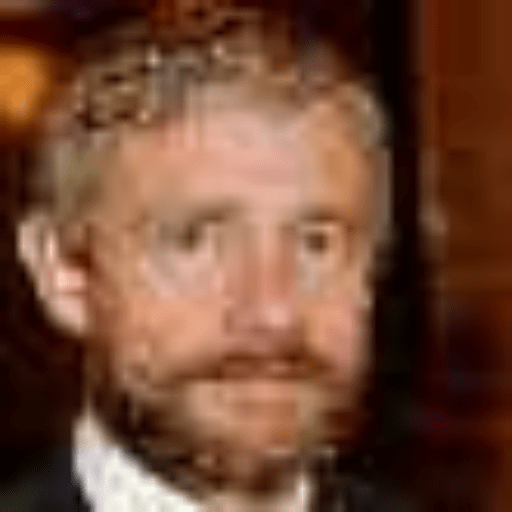}
    \includegraphics[width=0.16\linewidth]{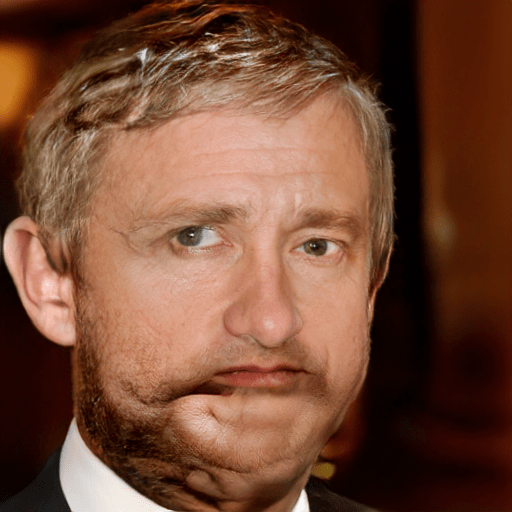}
   \includegraphics[width=0.16\linewidth]{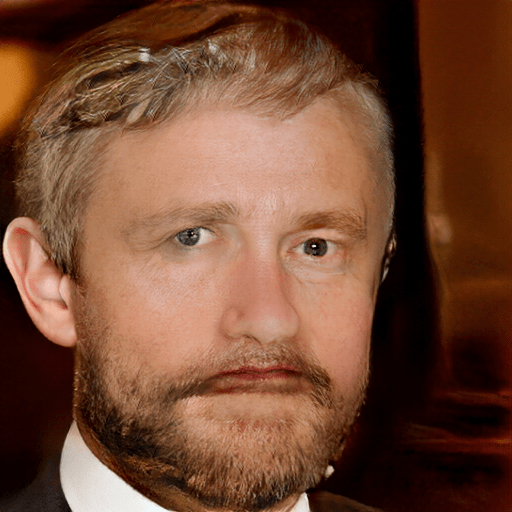}
   \includegraphics[width=0.16\linewidth]{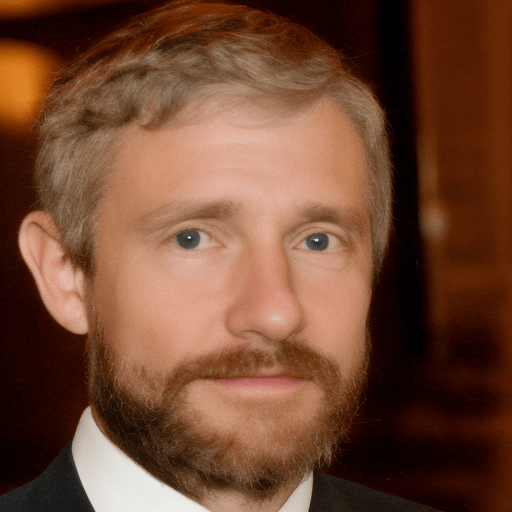}
   \includegraphics[width=0.16\linewidth]{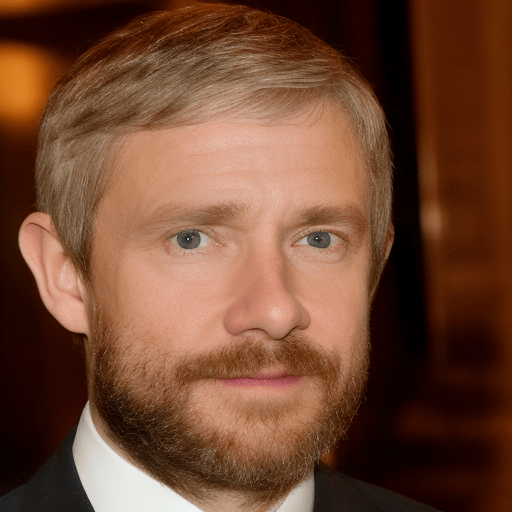}
   \includegraphics[width=0.16\linewidth]{parent/figures/f_diff_synth_degs/16_gt.png}\\

   \begin{subfigure}[c]{0.16\textwidth}
        \centering    \caption{\footnotesize{\textsc{Degraded Image}}}
    \end{subfigure}
    \begin{subfigure}[c]{0.16\textwidth}
        \centering    \caption{\footnotesize{\textsc{ASFFNet~\cite{Li_2020_CVPR}}}}
    \end{subfigure}
    \begin{subfigure}[c]{0.16\textwidth}
        \centering    \caption{\footnotesize{\textsc{DMDNet~\cite{li2022learning}}}}
    \end{subfigure}
    \begin{subfigure}[c]{0.16\textwidth}
         \centering    \caption{\footnotesize{\textsc{DiffBIR~\cite{lin2023diffbir}}}}
    \end{subfigure}
    \begin{subfigure}[c]{0.16\textwidth}
         \centering    \caption{\footnotesize{\textsc{Ours}}}
    \end{subfigure}
    \begin{subfigure}[c]{0.16\textwidth}
         \centering    \caption{\footnotesize{\textsc{Ground Truth}}}
    \end{subfigure}

   \caption{\textbf{Our approach is agnostic to different types of degradation, such as blur (top row), compression (middle row).} Prior face image restoration methods (b, c) have their estimates considerably affected by the nature of the degradation (note artifacts near eyes and mouth). Prior unconditional diffusion methods (d) have more consistent performance, but with lost identity information (note \textbf{eye shape}, \textbf{nose shape}). Our proposed method provides consistent restoration across a range of degradations, while retaining identity.
   }
   \label{fig:high_deg1}
\end{figure*}

\smallskip
\myparagraph{Comparison on synthetic degradations.} \cref{fig:synth_deg1} and \cref{fig:synth_deg_supp} show qualitative comparisons of the methods on the CelebRef-HQ test set subject to the synthetic degradations described above. It is noteworthy that ASFFNet and DMDNet are trained on perfectly aligned face images. During test time, highly accurate face alignment is required to eliminate domain gap and get high-quality restoration results, which is hard to achieve on severely degraded images. As a result, the results of both methods are underwhelming. On the other hand, diffusion-based methods (DiffBIR and ours) are possess a generative prior trained on huge amount of unaligned faces and can generalize well on test images without specific alignment, which leads to significantly higher restoration quality.
MyStyle, which relies on a personalized generative prior, shows very good identity retention, however the restored images lack fidelity to the input degraded image. Blind restoration techniques, such as CodeFormer, GFP-GAN and DiffBIR lead to significant identity drifts, while retaining fidelity to the input image.  
Compared to DiffBIR, our method achieves better fidelity to the ground truth face details, especially in terms of eye shape, eye color, nose shapes and eyebrows, demonstrating that the proposed method can indeed better preserve identities.

In addition to qualitative comparisons, we also report the quantitative results on the entire CelebRef-HQ test set in \cref{tab:emprical_analysis}. We report PSNR and SSIM as full-reference metrics. 
To quantify the identity disparity from the ground truth image, we use ArcFace~\cite{deng2019arcface} similarity. As observed in the qualitative comparisons, ASFFNet and DMDNet cannot restore the images effectively, resulting in low scores on all the metrics. \pc{On the other hand, MyStyle results in poor quantitative performance, even through the qualitative results show restored images with high fidelity. This can be attributed to the lack of faithfulness of the restored image to the input image. The blind restoration methods (GFP-GAN, CodeFormer, DR2 and DiffBIR) show varying degrees of performance, however DiffBIR is found to perform the best among these.} Our method achieves higher identity preservation, at the cost of slightly lower PSNR and SSIM to DiffBIR. One explanation is that while DiffBIR tends to generate smooth restorations to which PSNR and SSIM are not sensitive, our method injects realistic high-frequency details that may not be perfectly aligned with the ground truth image, thus decreasing the scores. \pc{In summary, across all compared methods, the proposed method is still able to provide high image fidelity, while retaining identity. Some additional notes on the baseline qualitative metrics may be found in~\cref{sec:quantitative_analysis_supp}.}  

\pc{
We also conduct a user study to identify the perceptual benefit of our contextual personalized prior (Appendix~\cref{sec:quantitative_analysis_supp}). We make two important observations: first, comparing unpersonalized diffusion-based face restoration~\cite{lin2023diffbir} with our personalized method, a large majority of partipants in our study note that personalization in fact improves identity-independent image quality as well. Second, we analyze the benefit of our personalized prior towards perceived identity retention while being faithful to the input image. Specifically, we note that the proposed method is predominatly found to be superior, when compared with unconditional diffusion-based image restoration~\cite{lin2023diffbir}, reference-based image restoration~\cite{zhou2022towards}, and GAN-based personalized priors~\cite{nitzan2022mystyle}.
}

\smallskip
\myparagraph{Comparison on real degradations.} \cref{fig:real_deg1} and \cref{fig:real_deg_supp}  shows the performance of each method on real-world images. For each test image, the model is personalized with 10 reference images of the same identity from CelebRef-HQ. We also provide a reference image in column (f) for better evaluation of the identity preservation. This set of experiments show that our personalized model generalizes well to real-world degraded images. In all cases, our method achieves significantly better image quality than ASFFNet DMDNet, GFP-GAN and CodeFormer, and preserves identity better than DiFFBIR. When compared with MyStyle, in this operating regime the proposed method is considerably more faithful to the input degraded image, while being able to retain identity.

\smallskip
\myparagraph{Consistent restoration through personalized prior.} To show the strength of our personalized prior, we conduct an experiment by applying different synthetic degradations on the same image and observe how each method restores them, as shown in \cref{fig:high_deg1}. The upper row shows a blurring degradation, while the lower row shows a compression degradation. For ASFFNet and DMDNet, the quality of the restored images varies with the degradations being applied. DiffBIR generates higher quality images, but the identity remains sensitive to the input degraded image. For example, in \cref{fig:high_deg1}(d), note the inconsistent eye and nose shape, when compared with \cref{fig:high_deg1}(f). Our personalized model effectively restores the images with consistent perceptual quality and identity fidelity. This indicates the stability of the contextual prior and the reliability of the encoder conditioning, across degradation types.

\begin{figure}
    \centering
    \includegraphics[width=\linewidth]{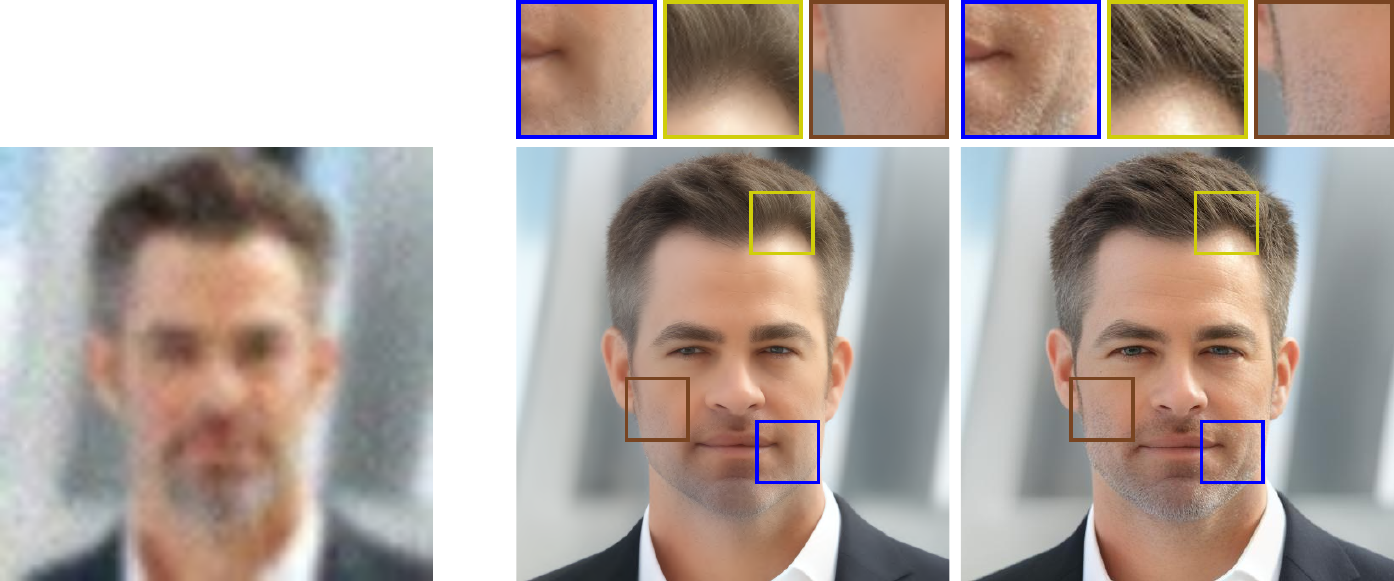}\\
    
    \begin{subfigure}[c]{0.35\linewidth}
        \centering    \caption*{\footnotesize{{Deg. Image~~~~~~~~~~~}}}
    \end{subfigure}
    \begin{subfigure}[c]{0.30\linewidth}
        \centering    \caption{\footnotesize{{In-Context Textual Pivoting}}}
    \end{subfigure}
    \begin{subfigure}[c]{0.29\linewidth}
        \centering    \caption{\footnotesize{{+ Model-based Pivoting}}}
    \end{subfigure}\\

    \includegraphics[width=\linewidth]{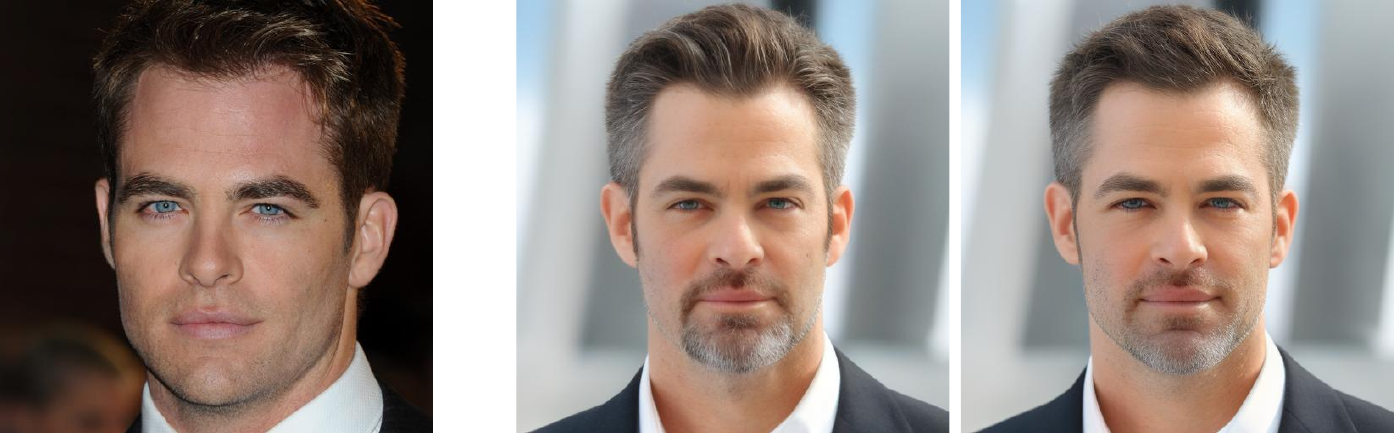}\\
    
    \begin{subfigure}[c]{0.35\linewidth}
        \centering    \caption*{\footnotesize{{Id. Reference~~~~~~~~~~~}}}
    \end{subfigure}
    \begin{subfigure}[c]{0.30\linewidth}
        \centering    \caption{\footnotesize{{Out-of-Context Textual Pivoting}}}
    \end{subfigure}
    \begin{subfigure}[c]{0.29\linewidth}
        \centering    \caption{\footnotesize{{+ Model-based Pivoting}}}
    \end{subfigure}\\
    
    \caption{\pc{\textbf{Ablation: understanding dual-pivot tuning.} (a) In-context textual pivoting enables identity injection, however there still remains a gap in both identity and quality. This is bridged through the model-based pivoting (b). Note the difference in texture information within the insets, for features such as hair (yellow) and beard (blue, brown). Please zoom in for a clearer distinction. This can be compared with the alternative proposed approach. By beginning with out-of-context textual pivoting (c), significant drifts in identity are seen (note the eye shape) as the restoration pipeline is unable to leverage the personalized prior. However in this setting, the effect of the model-based pivoting is much more pronounced (d): alignment of $\mathbf{E}$ with the personalized prior leads to significant improvement in identity preservation, and comparable performance with the upper row.}}
    \label{fig:variants_supp}
\end{figure}

\begin{figure}
    \centering

    \begin{subfigure}[c]{0.24\linewidth}
         \centering   
         \includegraphics[width=\linewidth]{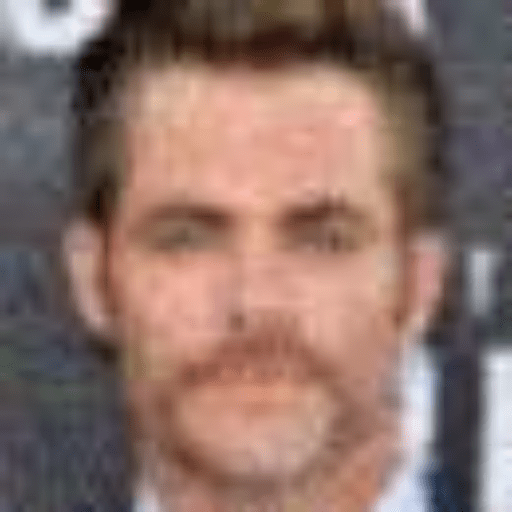}
         \caption{Degraded inp.}
    \end{subfigure}  
    \begin{subfigure}[c]{0.24\linewidth}
         \centering   
         \includegraphics[width=\linewidth]{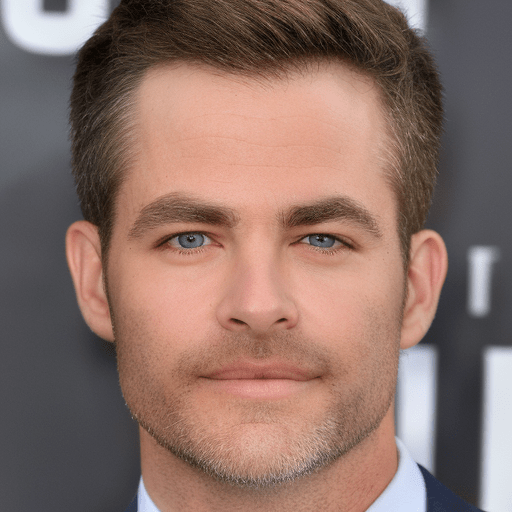}
         \caption{CFG=2.0}
    \end{subfigure} 
    \begin{subfigure}[c]{0.24\linewidth}
         \centering   
         \includegraphics[width=\linewidth]{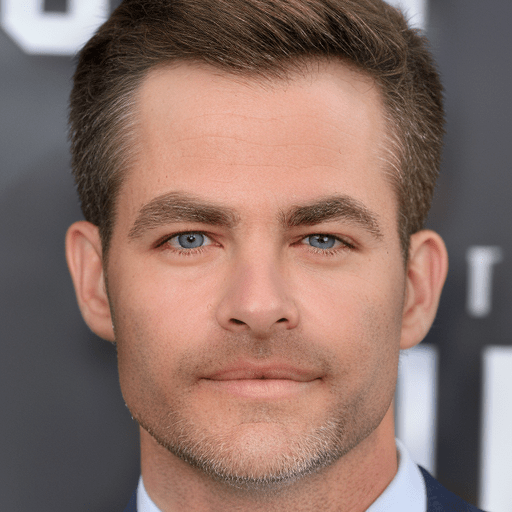}
         \caption{CFG=4.0}
    \end{subfigure} 
    \begin{subfigure}[c]{0.24\linewidth}
         \centering   
         \includegraphics[width=\linewidth]{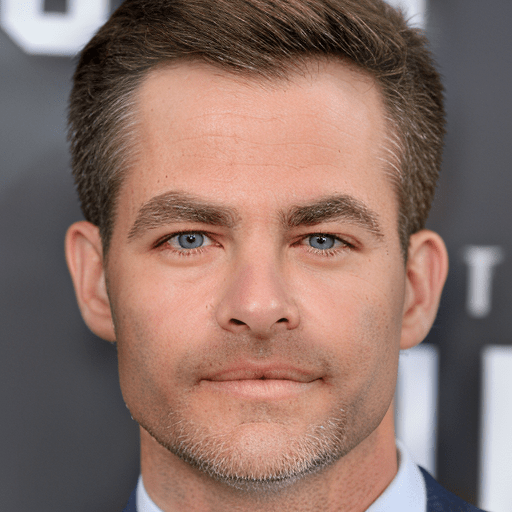}
         \caption{CFG=6.0}
    \end{subfigure}

    \caption{\textbf{Effect of classifier free guidance (CFG).} Sweeping across CFG enables a fidelity-diversity tradeoff, also manifesting as a tradeoff between image sharpness with realism (higher the CFG, higher the sharpness). This provides a degree of user control. }
    \label{fig:cfg}

    \vspace{-4mm}
    
\end{figure}

\smallskip
\pc{
\paragraph{Ablation: Understanding Dual-Pivot Tuning.}
 We find the proposed dual-pivot tuning approach to be a general technique for personalization of guided diffusion models. \cref{fig:variants_supp} qualitatively explores the benefits of this method. Specifically in our case, we find that alternate strategies for personalization exist. The first is our chosen strategy: in-context textual pivoting (\cref{fig:variants_supp}(a)), which injects identity information, followed by model-based pivoting (\cref{fig:variants_supp}(b)), which enables better utilization of the general restoration prior to get high-fidelity restored images. The alternate approach involves beginning with out-of-context textual pivoting (where $\mathbf{G}$ is personalized outside the context of $\mathbf{B}$). As shown in \cref{fig:variants_supp}(c), this step leads to significant gap in identity as the restoration pipeline is unable to utilize the newly injected identity information. However, post the model-based pivoting (\cref{fig:variants_supp}(d)), this is resolved, leading to high fidelity, identity preserving image restoration. The dual-pivot tuning approach successfully personalized the diffusion model in both these settings. As discussed previously, we find the in-context textual pivoting to enable faster personalization in terms of the the model-based pivoting step.

}

\smallskip
\myparagraph{Classifier-free guidance scale.} \cref{fig:cfg} shows an ablation study on the effect of classifier-free guidance scale. As can be seen, the parameter allows for trading off image sharpness with realism, allowing for pereference-based tuning. Specifically, we find that increasing the CFG scale makes the restored images sharper, at the cost of realism (additional quantitative analysis in Appendix~\cref{sec:additional_observations}). \pc{We also note in \cref{fig:uncond_cond_infer} that the proposed method does not require conditional inference across all denoising steps. That is, to enable personalized restoration, we find that even though initial denoising steps are carried out unconditionally, effective personalized restoration is possible.}

\begin{figure}
    \centering
    \begin{subfigure}[c]{0.24\linewidth}
        \centering 
        \caption*{\textbf{Face Swapping}}
    \end{subfigure}\\
    \begin{subfigure}[c]{0.24\linewidth}
         \centering   
         \includegraphics[width=\linewidth]{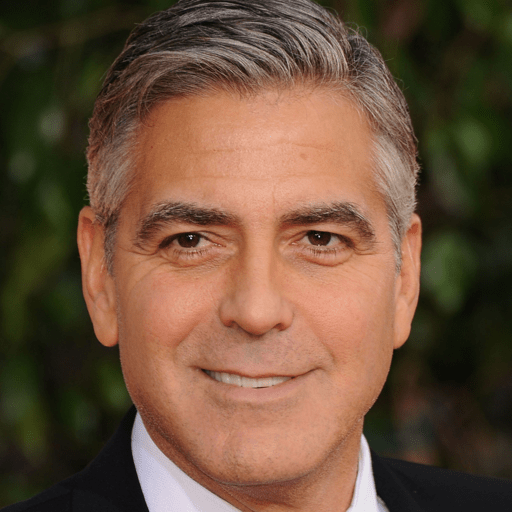}
         \caption*{Input}
    \end{subfigure}  
    \begin{subfigure}[c]{0.24\linewidth}
         \centering   
         \includegraphics[width=\linewidth]{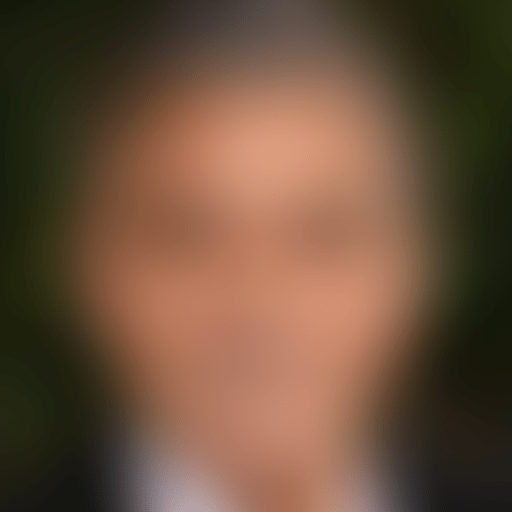}
         \caption*{Induced blur}
    \end{subfigure} 
    \begin{subfigure}[c]{0.24\linewidth}
         \centering   
         \includegraphics[width=\linewidth]{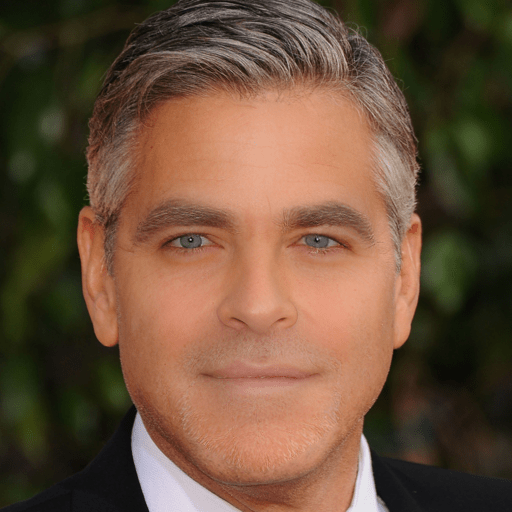}
         \caption*{Face swap}
    \end{subfigure} 
    \begin{subfigure}[c]{0.24\linewidth}
         \centering   
         \includegraphics[width=\linewidth]{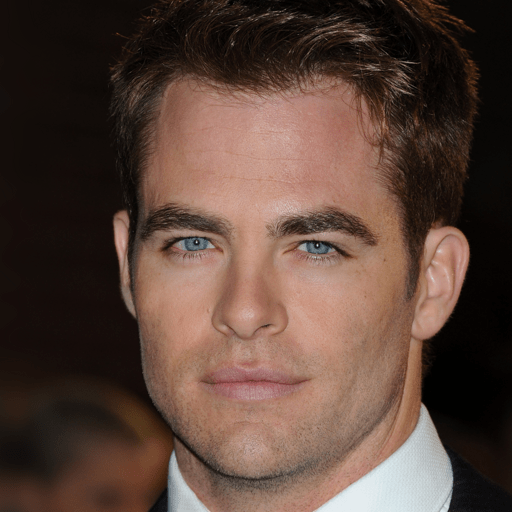}
         \caption*{Reference}
    \end{subfigure} \\
    \begin{subfigure}[c]{0.35
    \linewidth}
        \centering 
        \caption*{\textbf{Text-guided Editing}}
    \end{subfigure}\\

    \begin{subfigure}[c]{0.32\linewidth}
         \centering   
         \includegraphics[width=\linewidth]{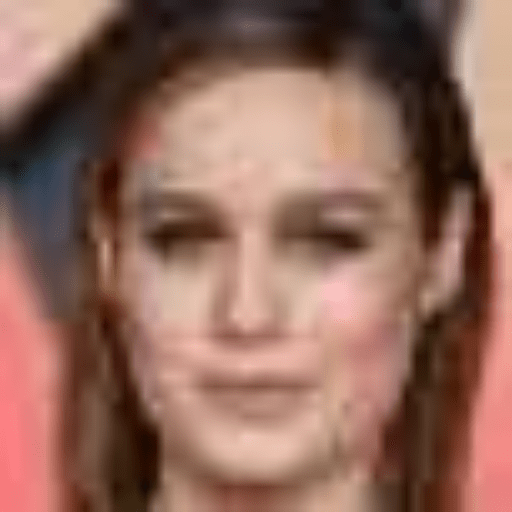}
         \caption*{Deg. Inp.}
    \end{subfigure}  
    \begin{subfigure}[c]{0.32\linewidth}
         \centering   
         \includegraphics[width=\linewidth]{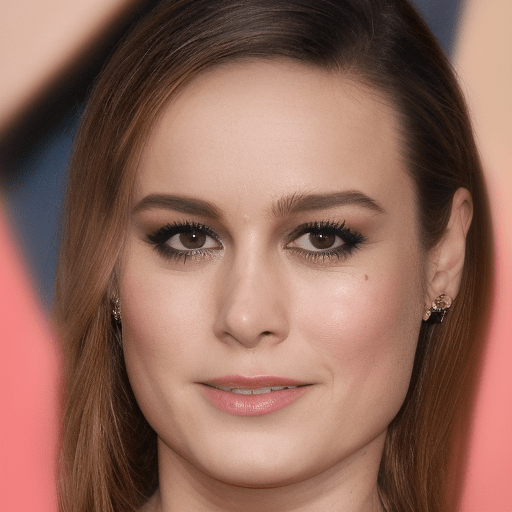}
         \caption*{No edit}
    \end{subfigure} 
    \begin{subfigure}[c]{0.32\linewidth}
         \centering   
         \includegraphics[width=\linewidth]{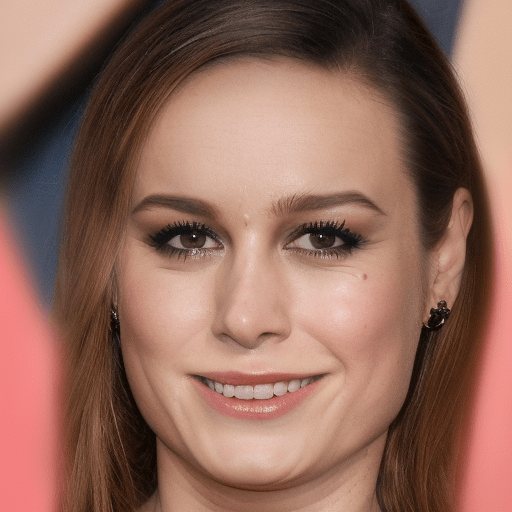}
         \caption*{``smile''}
    \end{subfigure} \\
    
    \begin{subfigure}[c]{0.32\linewidth}
         \centering   
         \includegraphics[width=\linewidth]{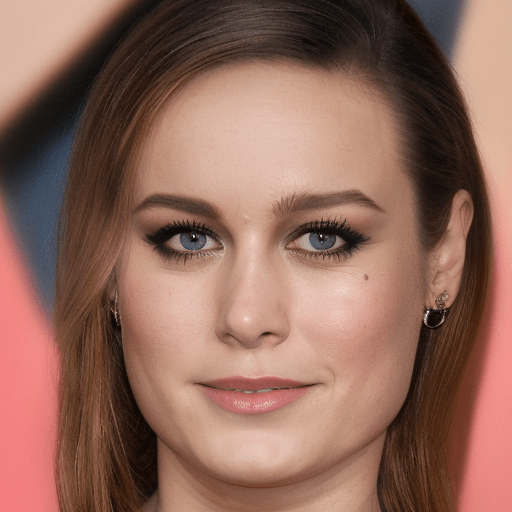}
         \caption*{``blue eyes''}
    \end{subfigure} 
    \begin{subfigure}[c]{0.32\linewidth}
         \centering   
         \includegraphics[width=\linewidth]{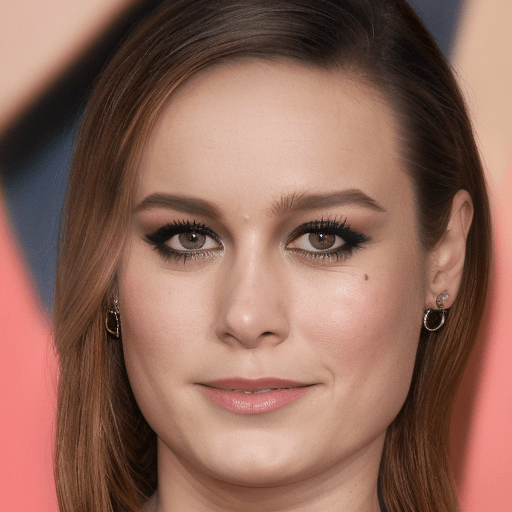}
         \caption*{``green eyes''}
    \end{subfigure} 
    \begin{subfigure}[c]{0.32\linewidth}
         \centering   
         \includegraphics[width=\linewidth]{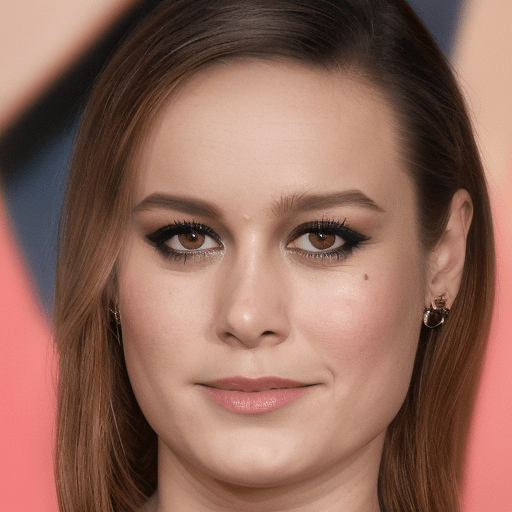}
         \caption*{``yellow eyes''}
    \end{subfigure}

    \caption{\textbf{Editing applications: face swapping text-guided editing.} The upper row shows the application of face swapping, while the lower row shows performance on text-guided editing.}
    \label{fig:scene-edit}

    \vspace{-4mm}
    
\end{figure}

\smallskip
\myparagraph{Additional Applications:} \cref{fig:scene-edit} shows the ability of our personalized generative prior to enable applications such as face swapping (top row) and text-guided editing (bottom row). Specifically, we blur the source image to obscure identity. Then, for face swapping we follow by restoration using the personalized prior for a new identity, then stitching the restored face back to the unblurred, original image. In the text-guided editing application we utilize the semantic prior of the diffusion model for text-conditional editing of the restored image. For example, with the term ``smiling" as part of the prompt, we see a smile in the expression of the individual in the output image. In addition, with the phrase ``blue", ``green" or ``yellow eyes", we see a clear variation in the eye color. Please zoom in for clearer visualization.

\vspace{-2mm}
\section{Conclusion}
\label{sec:conclusion}

We propose a technique for personalizing a diffusion prior for face image restoration, leveraging the capabilities of few-shot fine-tuning based on a set of example images of the person. Our method achieves high fidelity to both the input image and the identity of the person. We conduct extensive experiments to demonstrate the superiority of our method in comparison to various state-of-the-art alternatives for both blind and few-shot personalized face image restoration methods. 

\vspace{-4mm}
\paragraph{Limitations and Future Work.}

Although we make the very first step in using contextually personalized diffusion prior for face image restoration, it requires a computationally consuming fine-tuning process for each identity, which limits its deployability in large-scale systems. An important research direction is to inject few-shot identity in a feed-forward way, which we leave for future work. Furthermore, we would like to point out that, while the personalized model has improved fidelity to the identity, its general fidelity and quality are fundamentally limited by the underlying restoration method. Even though improving general quality is not the focus of this work, we believe the idea of contextual personalization holds promise for application in forthcoming restoration methods, achieving improved performance in both image quality and identity fidelity. \pc{Additionally, potential limitations native to personalization~\cite{ruiz2023dreambooth} on a small set of images can find their way to our method as well: namely overfitting towards features such as open eyes and smiling mouth if most of the training images show this, and the input degraded image is ambiguous in this aspect.}

\pc{
\vspace{-4mm}
\paragraph{Acknowledgements.}

We thank Jackson Wang and Fangzhou Mu for helpful discussions and feedback on this manuscript.
}

{
    \small
    \bibliographystyle{ieeenat_fullname}
    \bibliography{main}
}

\clearpage
\twocolumn[
        \centering
        \Large
        \vspace{0.5em}\textbf{Appendix} \\
        \vspace{1.0em}
       ]


\setcounter{section}{0}
\setcounter{figure}{0}
\setcounter{table}{0}

\renewcommand\thesection{\Alph{section}} 
\renewcommand\thesubsection{\thesection.\alph{subsection}} 
\renewcommand\thefigure{\Alph{figure}} 
\renewcommand\thetable{\Alph{table}} 

\section{Appendix Contents}
\label{sec:contents}
This appendix contains the following contents:
\begin{enumerate}
    \item Additional quantitative analysis (\cref{sec:quantitative_analysis_supp})
    \item Additional qualitative analysis (\cref{sec:qualitative_analysis_supp})
    \item Additional observations (\cref{sec:additional_observations})
    \item Ethical considerations (\cref{sec:ethical_considerations})
\end{enumerate}

\section{Additional Quantitative Analysis}
\label{sec:quantitative_analysis_supp}

\paragraph{Additional notes on  \cref{tab:emprical_analysis}.} 

An interesting result to note is the performance of~\cite{wang2023dr2}. We note that DR2 allows for tunable parameters $(N,\tau)$ selection to tradeoff restoration with fidelity to the original image. We choose $(N,\tau)=(16,35)$ as we find this combination able to restore the degradations in our synthetic images. However, this does come at a cost of fidelity to the input - this is reflected by low fidelity as well as identity metrics for DR2.

Another interesting observation is the performance of MyStyle~\cite{nitzan2022mystyle} in terms of identity retention. We see that the ArcFace metric for MyStyle is similar to other blind restoration techniques. This is a somewhat surprising observation, since MyStyle uses a personalized prior for restoration. This may be explained as follows: while MyStyle indeed shows strong indentity retention (as evidenced by qualitative results in \cref{fig:synth_deg_supp,fig:real_deg_supp,fig:synth_deg_supp_add_comb,fig:real_deg_supp_add_comb}), the method is unable to retain fidelity (in terms of color, lighting, texture, makeup and even pose). All these factors can affect the perceived identity of the image from the perspective of the ArcFace metric, leading the this anomaly. On the other hand, the ArcFace metric is a valid comparison of identity retention with all other comparison methods, since they are able to retain fidelity with the input degraded image.

\begin{figure}
    \centering
    \includegraphics[width=0.47\linewidth]{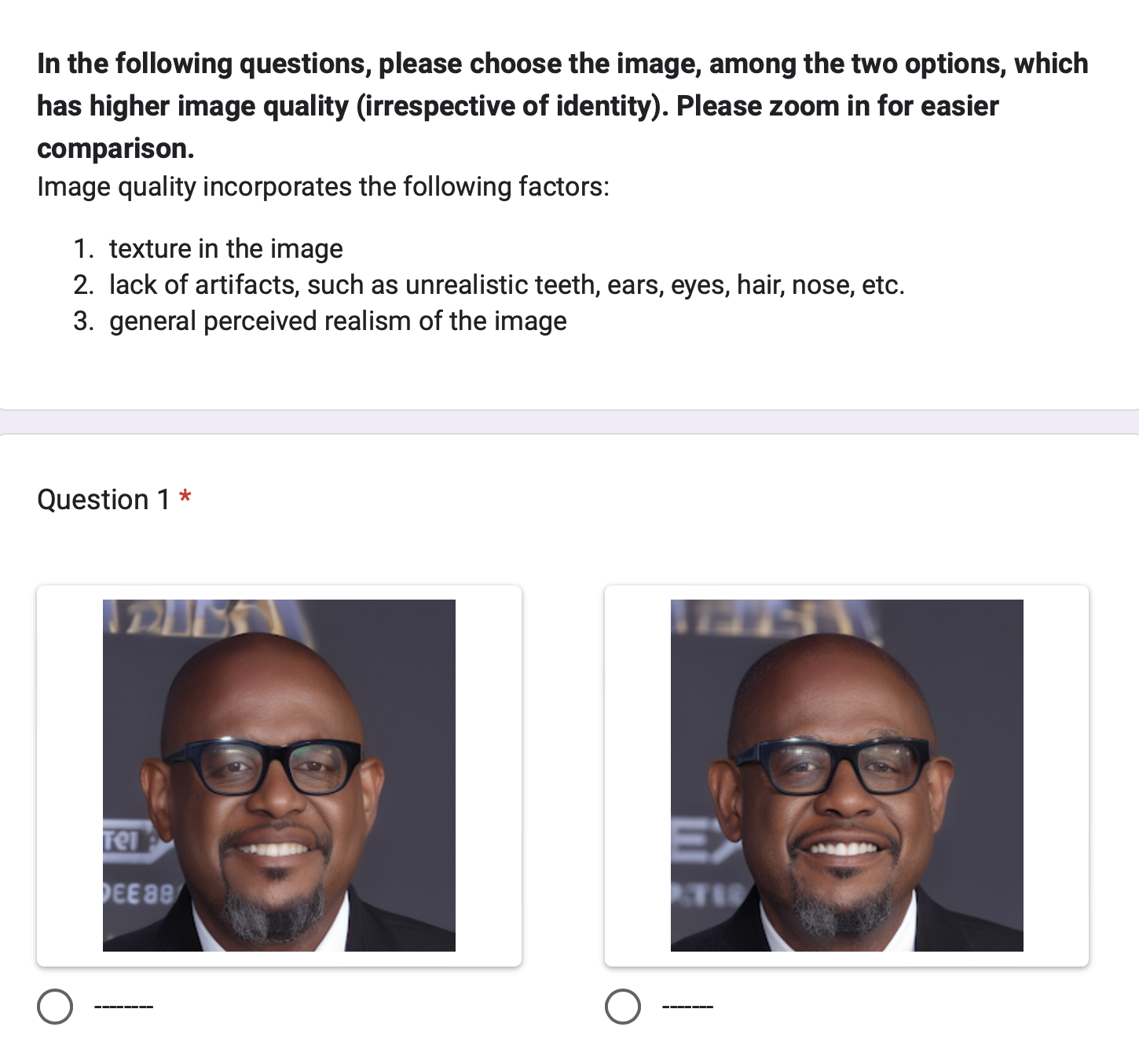}
    \includegraphics[width=0.51\linewidth]{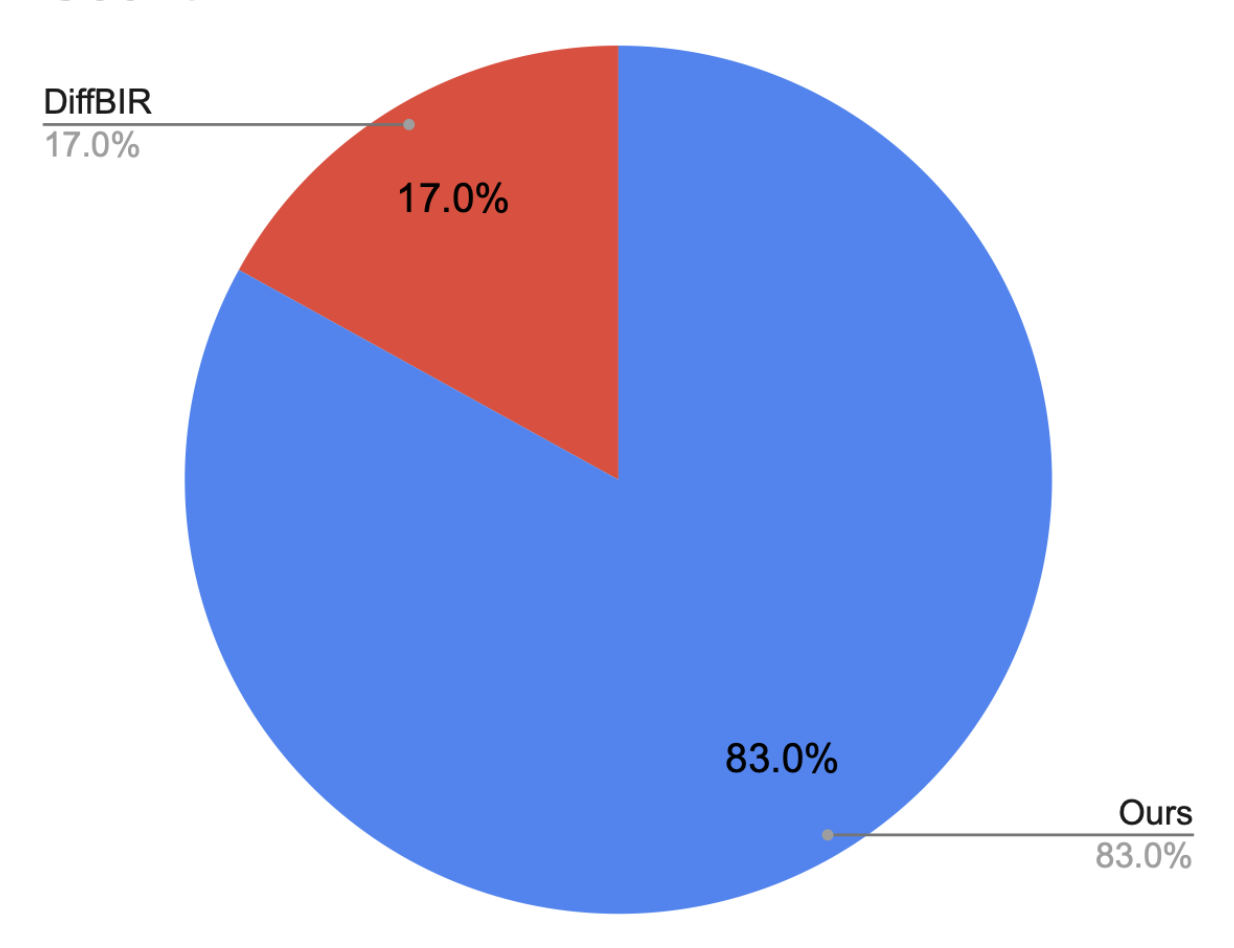}

    \begin{subfigure}[c]{0.47\linewidth}
        \centering    \caption{\footnotesize{{Example survey question}}}
    \end{subfigure}
    \begin{subfigure}[c]{0.51\linewidth}
        \centering    \caption{\footnotesize{{Survey Result}}}
    \end{subfigure}
    
    \caption{\textbf{User study: effect of personalization on perceived image quality.} When asked to choose the image with better perceived quality, we find users predominantly choosing the images with our personalized restoration. Our method is indicated in blue, while DiffBIR~\cite{lin2023diffbir} is shown in red, in the pie chart.}
    \label{fig:study_quality}
\end{figure}

\begin{figure}
    \centering
    \includegraphics[width=0.36\linewidth]{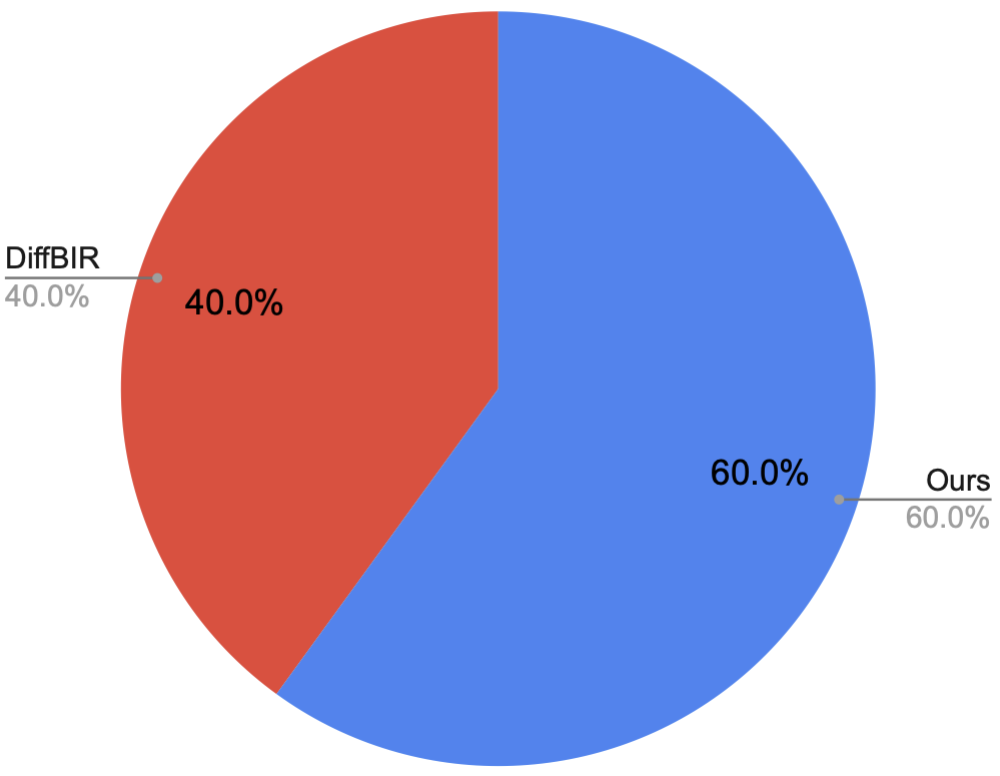}
    \includegraphics[width=0.30\linewidth]{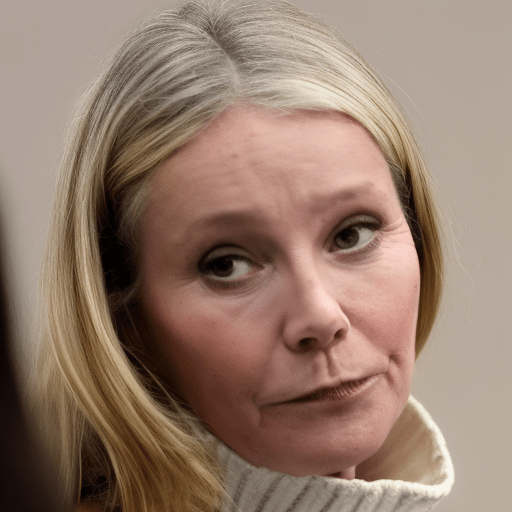}
    \includegraphics[width=0.30\linewidth]{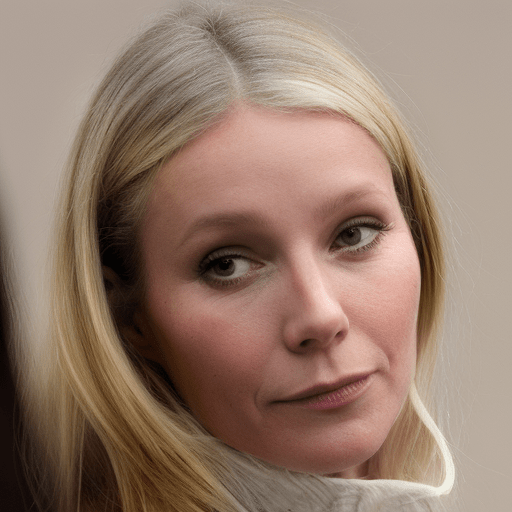}\\

    \includegraphics[width=0.36\linewidth]{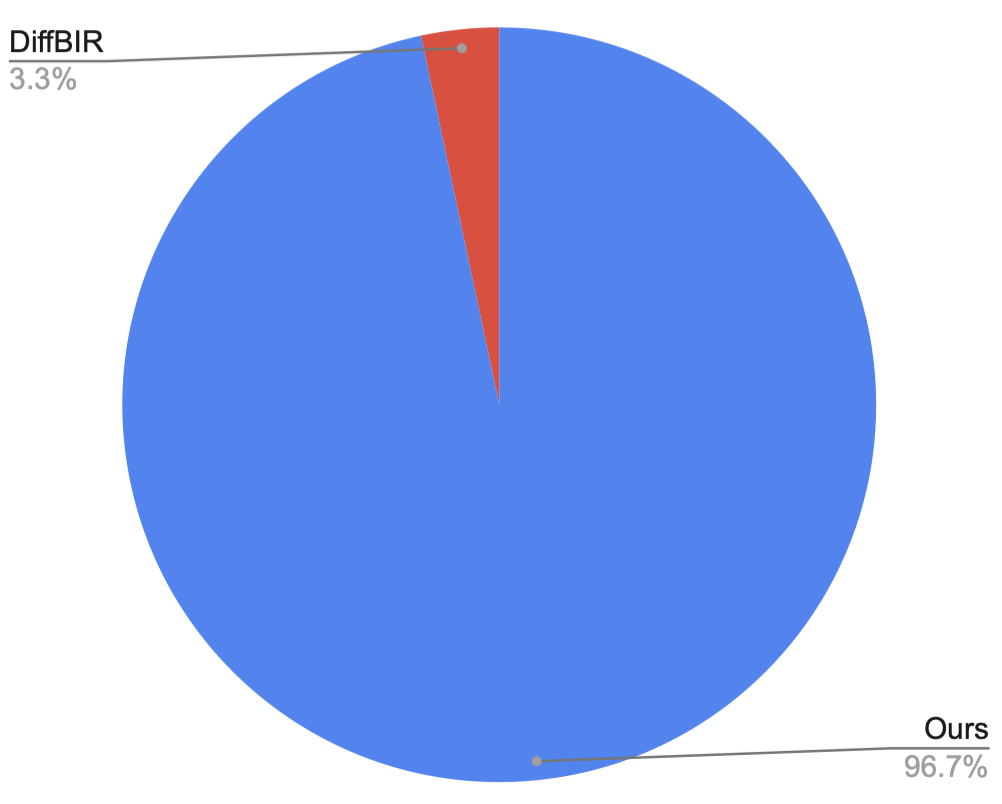}
    \includegraphics[width=0.30\linewidth]{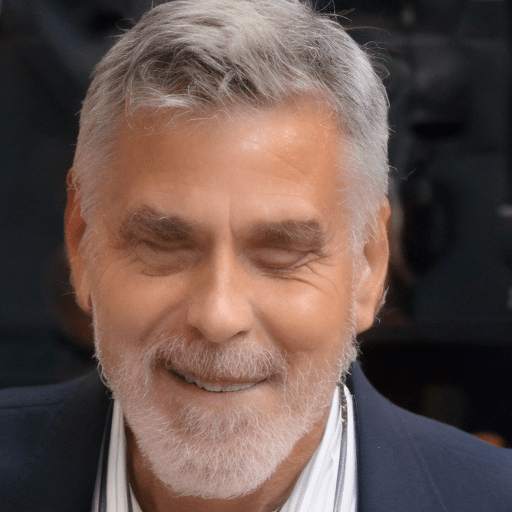}
    \includegraphics[width=0.30\linewidth]{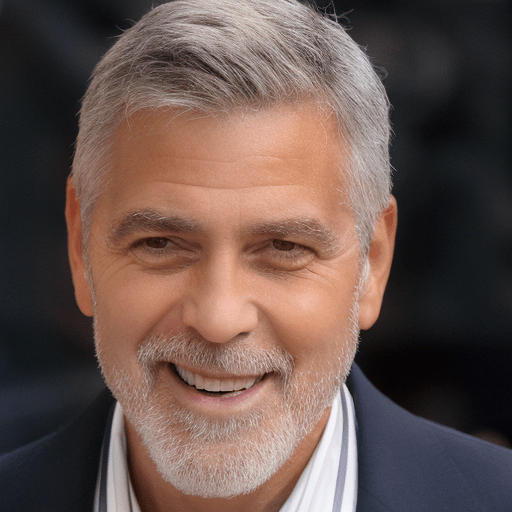}\\

    \begin{subfigure}[c]{0.36\linewidth}
        \centering    \caption{\footnotesize{{Survey Result (quality)}}}
    \end{subfigure}
    \begin{subfigure}[c]{0.30\linewidth}
        \centering    \caption{\footnotesize{{DiffBIR~\cite{lin2023diffbir}}}}
    \end{subfigure}
    \begin{subfigure}[c]{0.30\linewidth}
        \centering    \caption{\footnotesize{{Ours}}}
    \end{subfigure}
    
    \caption{\textbf{Two specific real degradation restorations, from the lens of image quality.} The upper row shows a case where respondent opinion is split - DiffBIR provides specific detail like wrinkles, while our approach provides structure and identity. The lower row shows a case with unanimous favor towards our method. This arises from specific artifacts in the DiffBIR output, which is avoided by having a strong personalized prior. Our method is indicated in blue, while DiffBIR~\cite{lin2023diffbir} is shown in red, in the pie charts.}
    \label{fig:study_quality_instance}
\end{figure}

\begin{figure}
    \centering
    \includegraphics[width=0.40\linewidth]{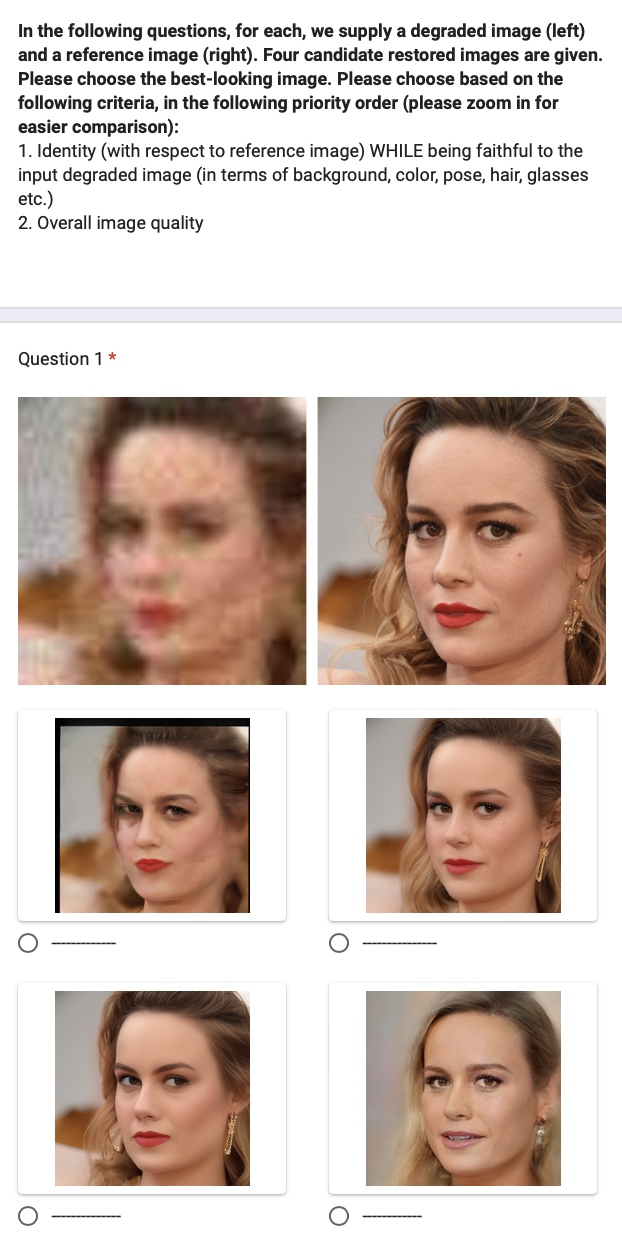}
    \includegraphics[width=0.57\linewidth]{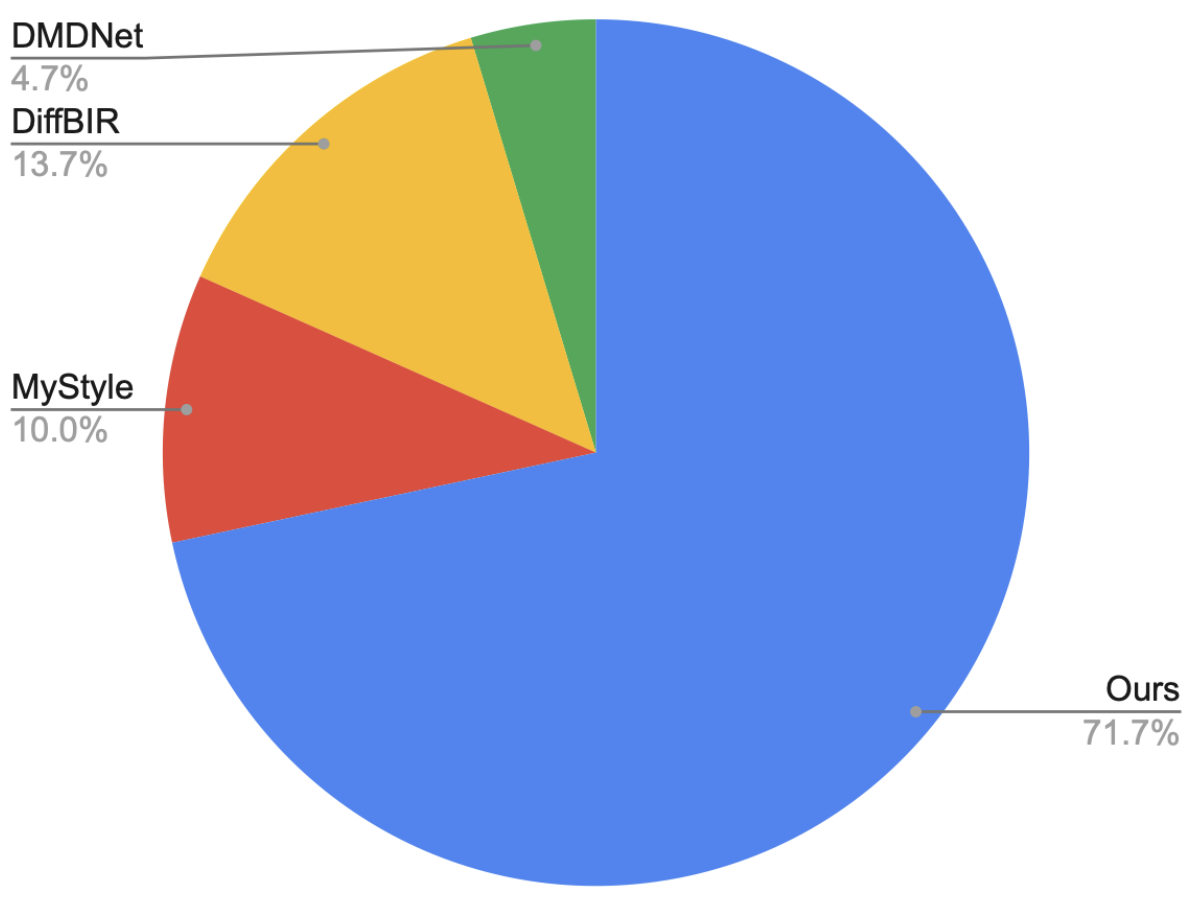}

    \begin{subfigure}[c]{0.40\linewidth}
        \centering    \caption{\footnotesize{{Example survey question}}}
    \end{subfigure}
    \begin{subfigure}[c]{0.57\linewidth}
        \centering    \caption{\footnotesize{{Survey Result}}}
    \end{subfigure}
    
    \caption{\textbf{User study: the effect of personalization on perceived image identity retention.} When asked to choose the image with the better-looking image, while prioritizing identity and faithfulness to input degraded image, we find users predominantly choosing the images with our personalized restoration. Our method is indicated in blue,  DiffBIR~\cite{lin2023diffbir} is shown in yellow, MyStyle~\cite{nitzan2022mystyle} is shown in red, and DMDNet~\cite{li2022learning} is shown in green in the pie chart.}
    \label{fig:study_identity}
\end{figure}

\begin{figure*}[t]
  \centering

   \includegraphics[width=0.16\linewidth]{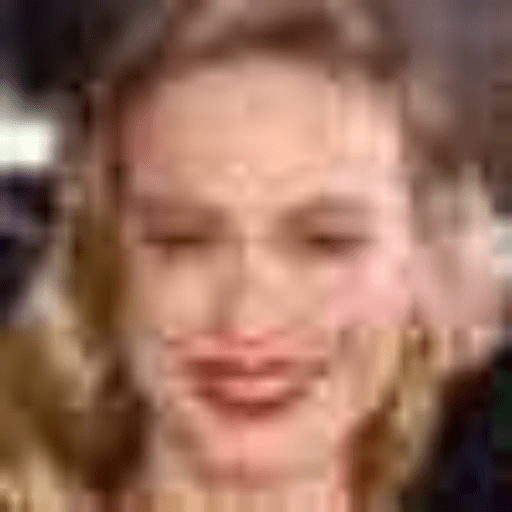}
   \includegraphics[width=0.16\linewidth]{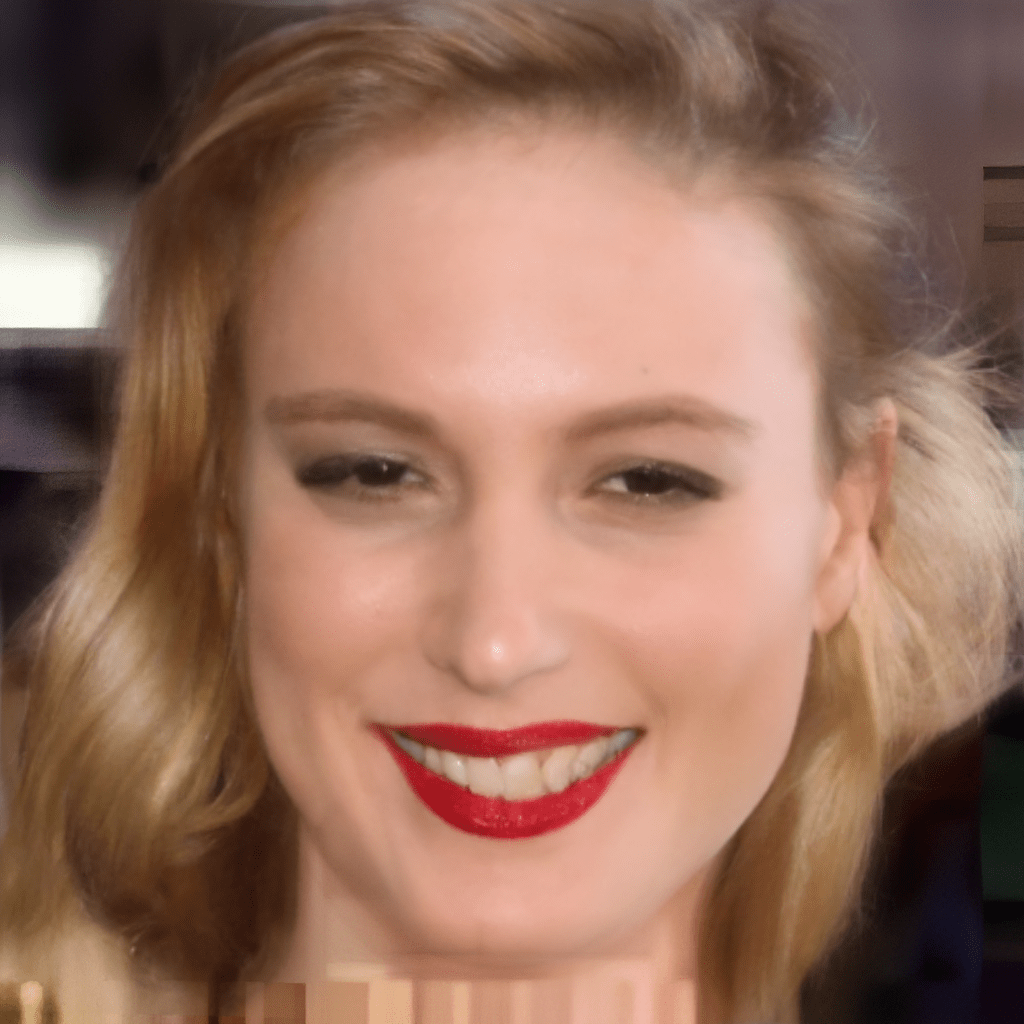}
   \includegraphics[width=0.16\linewidth]{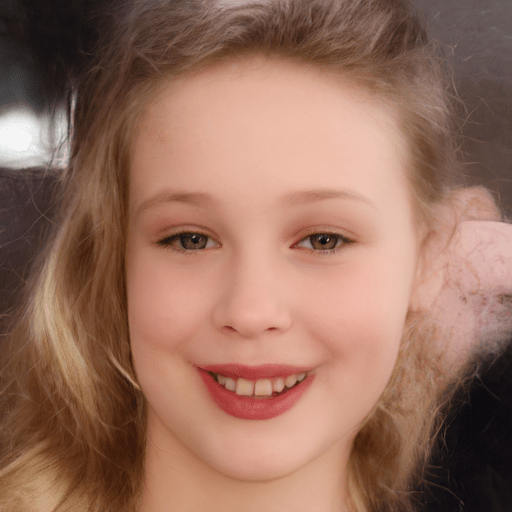}
   \includegraphics[width=0.16\linewidth]{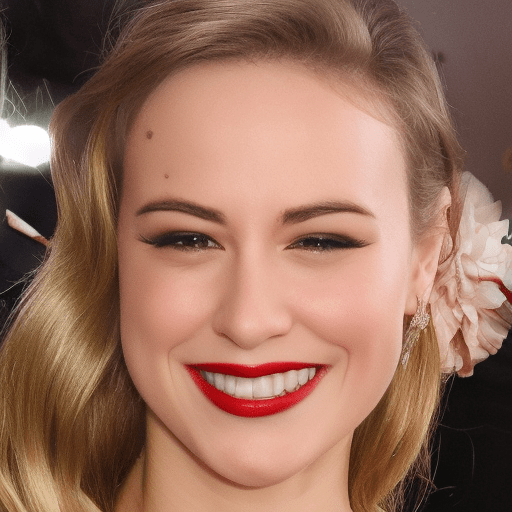}
   \includegraphics[width=0.16\linewidth]{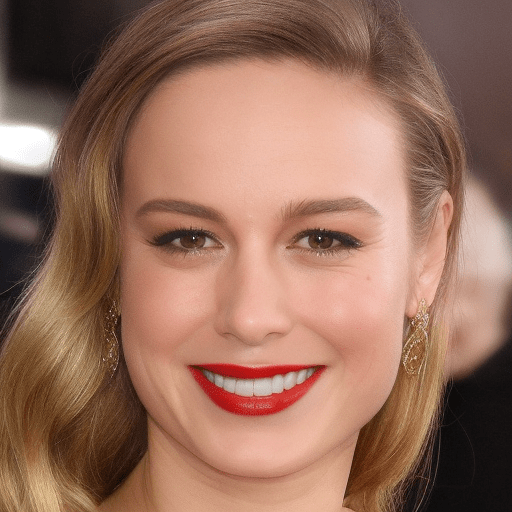}
   \includegraphics[width=0.16\linewidth]{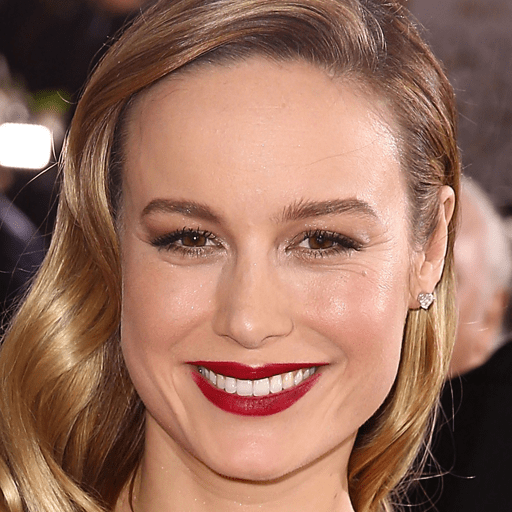}\\

    \includegraphics[width=0.16\linewidth]{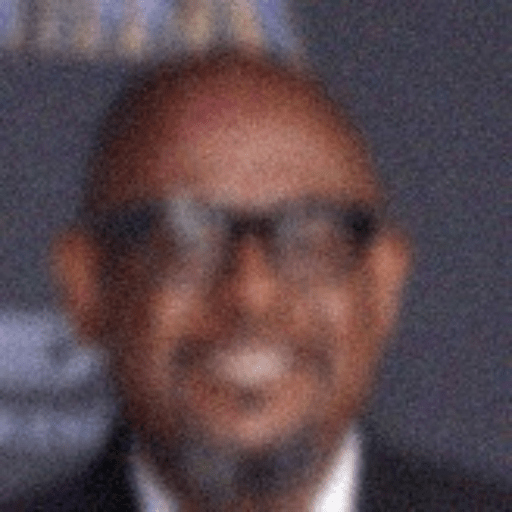}
    \includegraphics[width=0.16\linewidth]{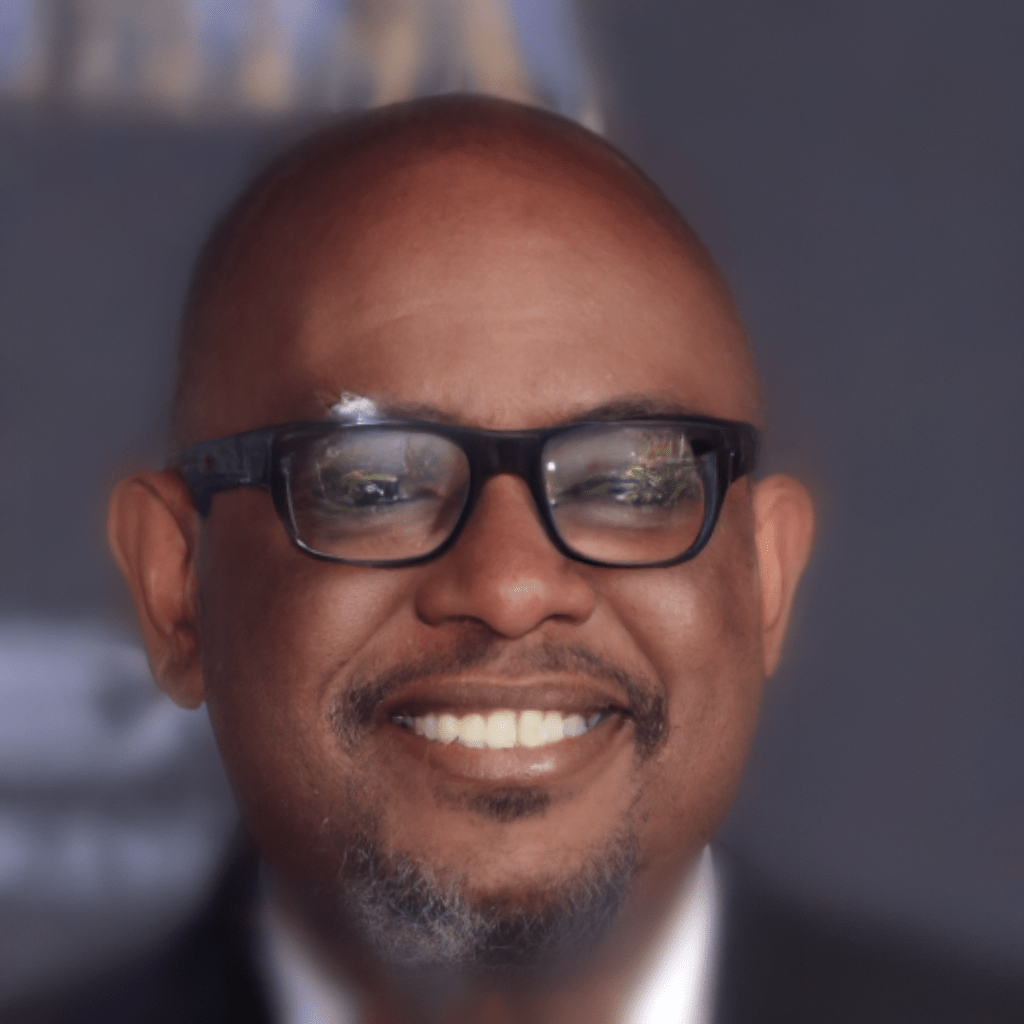}
   \includegraphics[width=0.16\linewidth]{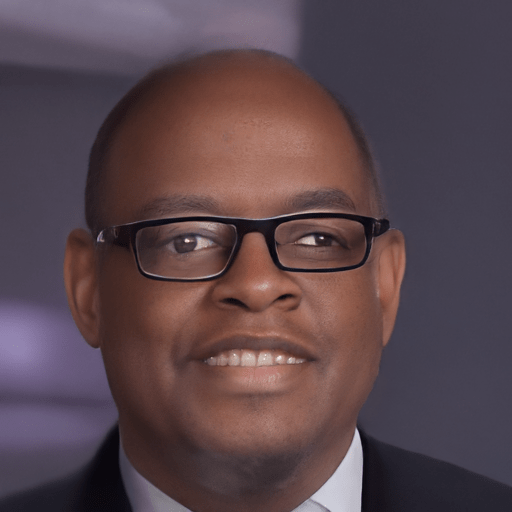}
   \includegraphics[width=0.16\linewidth]{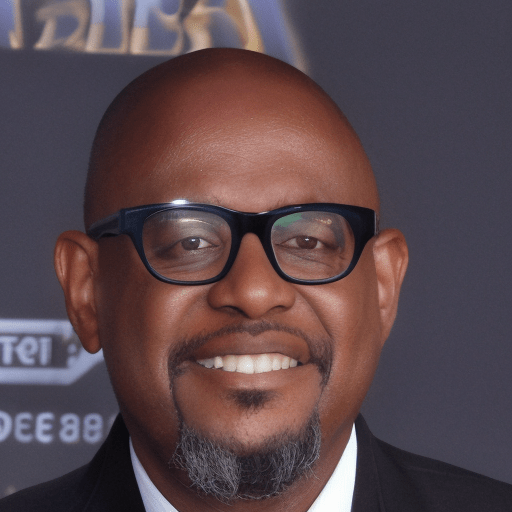}
   \includegraphics[width=0.16\linewidth]{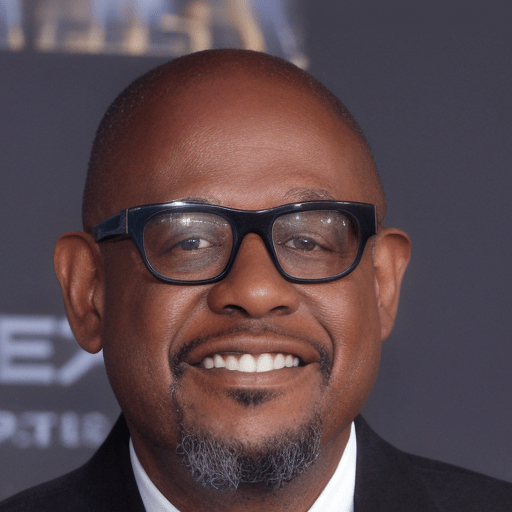}
   \includegraphics[width=0.16\linewidth]{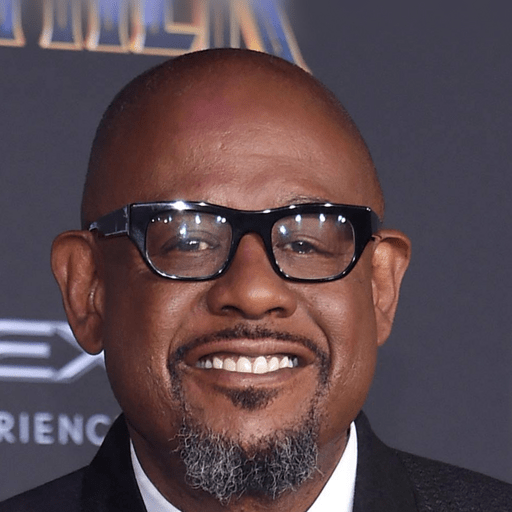}\\

   \includegraphics[width=0.16\linewidth]{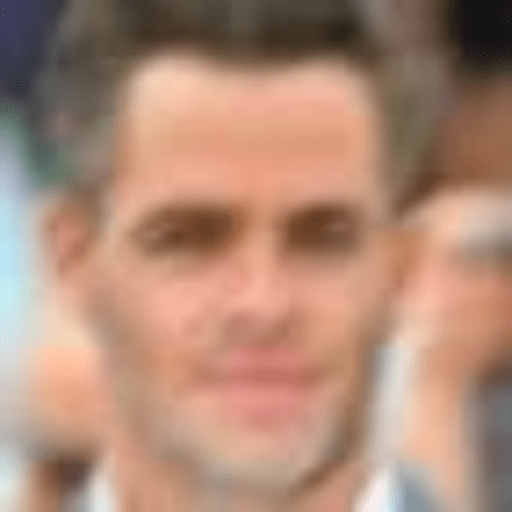}
    \includegraphics[width=0.16\linewidth]{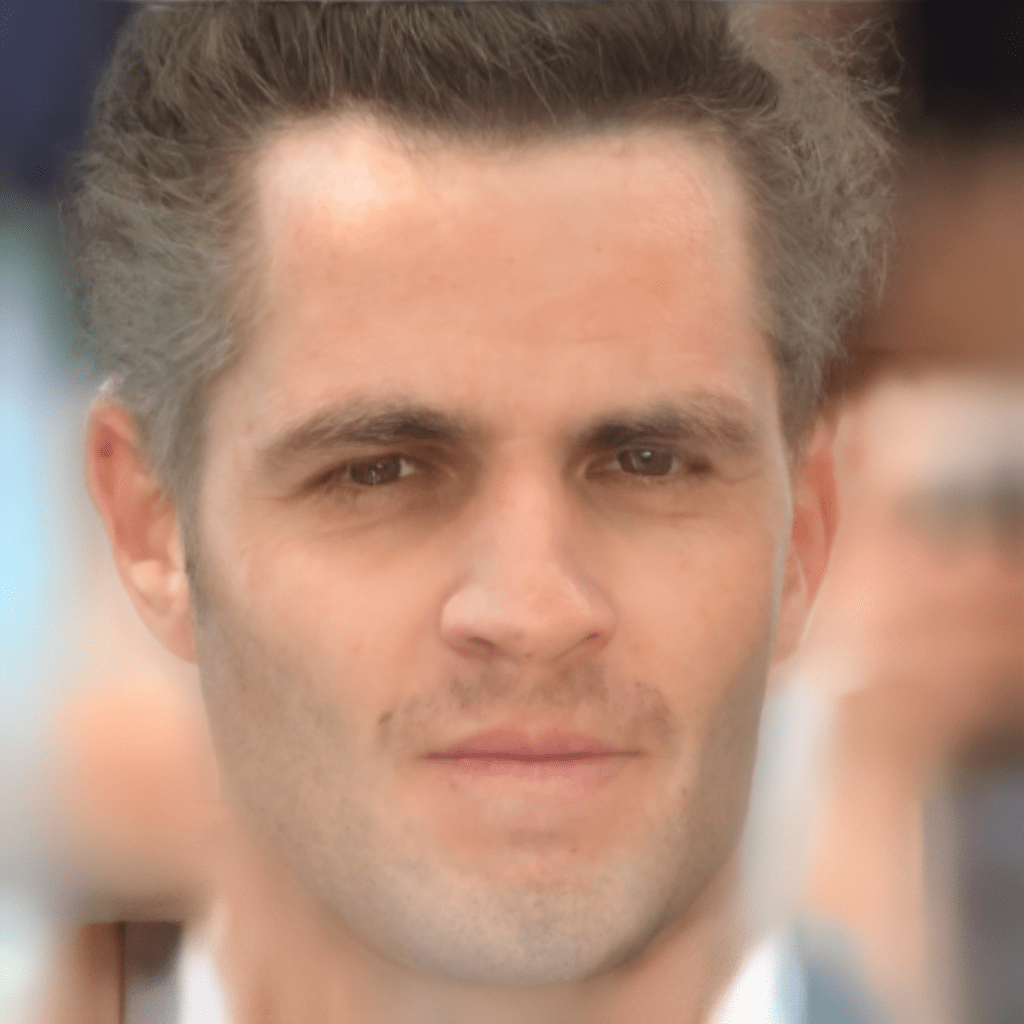}
   \includegraphics[width=0.16\linewidth]{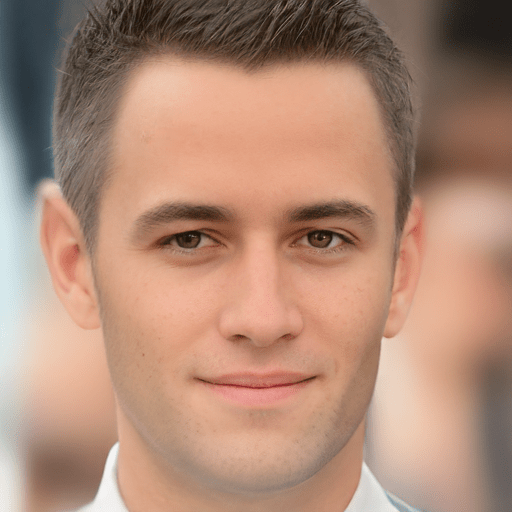}
   \includegraphics[width=0.16\linewidth]{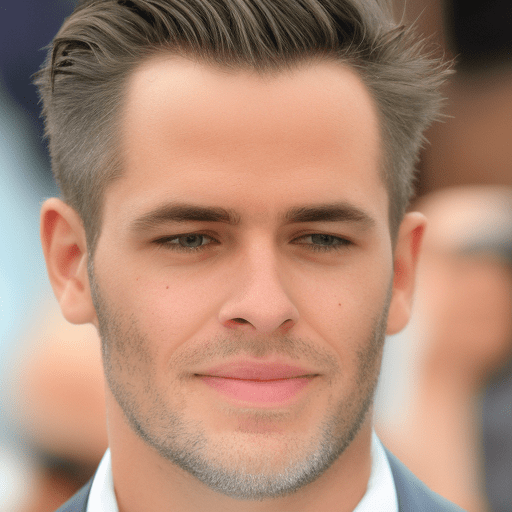}
   \includegraphics[width=0.16\linewidth]{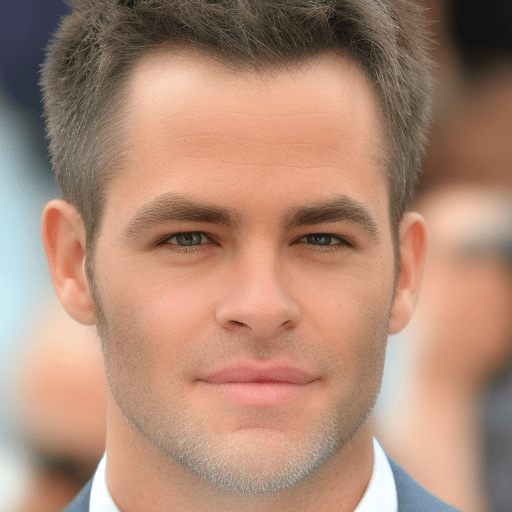}
   \includegraphics[width=0.16\linewidth]{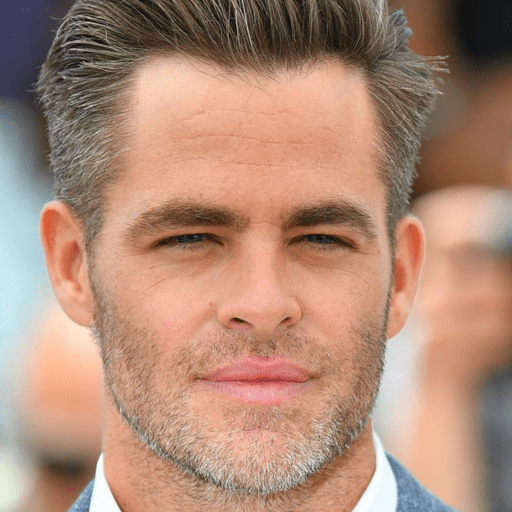}\\

   \begin{subfigure}[c]{0.16\textwidth}
        \centering    \caption{\footnotesize{\textsc{Degraded Image}}}
    \end{subfigure}
    \begin{subfigure}[c]{0.16\textwidth}
        \centering    \caption{\footnotesize{\textsc{DifFace~\cite{yue2022difface}}}}
    \end{subfigure}
    \begin{subfigure}[c]{0.16\textwidth}
        \centering    \caption{\footnotesize{\textsc{DR2~\cite{wang2023dr2}}}}
    \end{subfigure}
    \begin{subfigure}[c]{0.16\textwidth}
         \centering    \caption{\footnotesize{\textsc{DiffBIR~\cite{lin2023diffbir}}}}
    \end{subfigure}
    \begin{subfigure}[c]{0.16\textwidth}
         \centering    \caption{\footnotesize{\textsc{Ours}}}
    \end{subfigure}
    \begin{subfigure}[c]{0.16\textwidth}
         \centering    \caption{\footnotesize{\textsc{Ground Truth}}}
    \end{subfigure}

   \caption{\textbf{Comparison with existing diffusion-based blind image restoration methods.} We find that for the degree of degradation we deal with in our experiments, both DifFace~\cite{yue2022difface} as well as DR2~\cite{wang2023dr2} show unreliable performance, losing both identity and fidelity at different instances. DiffBIR~\cite{lin2023diffbir}, on the other hand, while not performing as well as our method in terms of identity and fidelity retention, is the best performer among the prior diffusion-based methods. We therefore choose DiffBIR as our base model and comparison benchmark for subsequent experiments.
   }
   \label{fig:diffusion_compar_supp}
\end{figure*}

\paragraph{User study.} We conduct a user study as a measure of perceptual quality and comparison with prior methods. We focus on two questions as part of the study: (a) can personalization help improve perceived image quality as well?, and (b) how does our method compare with prior methods in terms of identity-aware restoration. We perform our study across 30 participants. The survey consists of randomized, anonymized options that they can choose amongst. \cref{fig:study_quality,fig:study_identity} (a) show the exact guidelines provided as part of the study. For part (a), image quality, we compare against the base unconditional restoration method (DiffBIR~\cite{lin2023diffbir}), with the objective being to choose the better-quality image, irrespective of identity. For this case, we compare across 11 image pairs, containing both real and synthetic degradations. For part (b), identity-aware restoration, we compare with three methods: (i) DMDNet~\cite{li2022learning} (better-performing method both qualitatively and quantitatively when compared with ASFFNet~\cite{Li_2020_CVPR}); (ii) MyStyle~\cite{nitzan2022mystyle}, a personalized generative prior; (iii) DiffBIR~\cite{lin2023diffbir}, a unconditional diffusion-based method. For this case, we compare across 10 image sets, consists of synthetically degraded images.

We begin with discussing the first part of the user study: effect of personalization on identity-independent perceived image quality. \cref{fig:study_quality} shows the results of this part of the study. We observe that across the 30 study participants, a clear majority of the participants indicate the quality improvement that arises out of personalization. This is an unexpected result, and we explore this further by analyzing two specific instances, in \cref{fig:study_quality_instance}. The upper row shows a case where the users are broadly split between the candidate options. The unconditional method provides a restored image with detailed facial features such as wrinkles, while the proposed method provides less of such detail, with stronger identity cues. Both images look viable as natural images with good quality. However, the lower row in~\cref{fig:study_quality_instance} shows an instance where the study participants almost unanimously prefer our method. The reason for this is evident: the baselines method has considerable artifacts, especially near the eyes, while our method leads to a good quality, realistic face image. Through these observations, we can understand the effect of our personalization method on identity-independent image quality. In cases where the unconditional comparison method is able to perform, our personalized model remains stable and provides realistic looking faces. However, in cases where the comparison method fails, such as with high degradations, our method, throgh the strength of the identity prior, still results in realistic restored images. A combination of these two factors leads to superior perceived identity-independent image quality.

We next analyze the second part of the user study: the perceived strength of our identity-aware image restoration. \cref{fig:study_identity} shows the results of this part of the study. We see that, perhaps per expectation, participants rate our method as the predominant favorite in terms of identity-aware image restoration, while retaining faithfulness to the input degraded image. This can be seen across our various qualitative results and speaks to the strength and reliability of our personalized prior across identities and degradations. However, an interesting insight is the relative placement of the comparison methods. Specifically, study participants rate DiffBIR~\cite{lin2023diffbir} to be the second-best method, despite not retaining identity, as a result of its strong correlation to the input degraded image. On the other hand, MyStyle~\cite{nitzan2022mystyle}, while having a strong identity prior, is the third-best preferred method on average, as a result of it not being faithful to the input image. That is, perceptually, faithfulness to the input degraded image is given a higher priority by participants, despite the study guidelines placing both identity and faithfulness to input image at the same priority.

\section{Additional Qualitative Analysis}
\label{sec:qualitative_analysis_supp}

We now discuss additional qualitative results, comparing with a range of baselines. We first qualitatively compare our proposed method with existing diffusion-based blind restoration techniques. Specifically, we compare with DifFace~\cite{yue2022difface}, DR2~\cite{wang2023dr2} and DiffBIR~\cite{lin2023diffbir}. \cref{fig:diffusion_compar_supp} highlights these observations. We note that in our operating regime for degradation, both DR2 and DifFace result in inaccurate and unreliable restoration (in both these cases, the models trade off fidelity with restoration and need to be used accordingly depending on the degree of degradation). In constrast, DiffBIR, while worse in both detail and identity when compared with our method, proves to be a much more reliable blind restoration technique. We therefore use DiffBIR as our base model for our method, and use it as the representative diffusion-based blind restoration technique in our analysis.

\begin{figure*}[t]
  \centering
   \includegraphics[width=\linewidth]{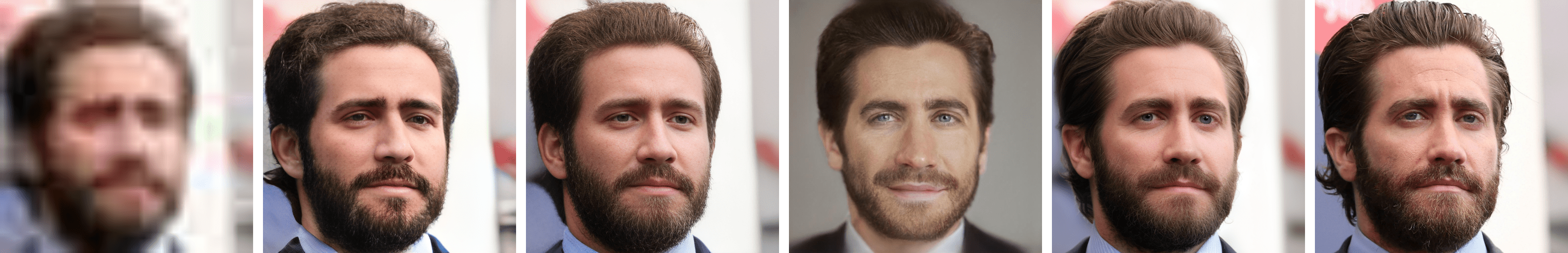}\\
   \includegraphics[width=\linewidth]{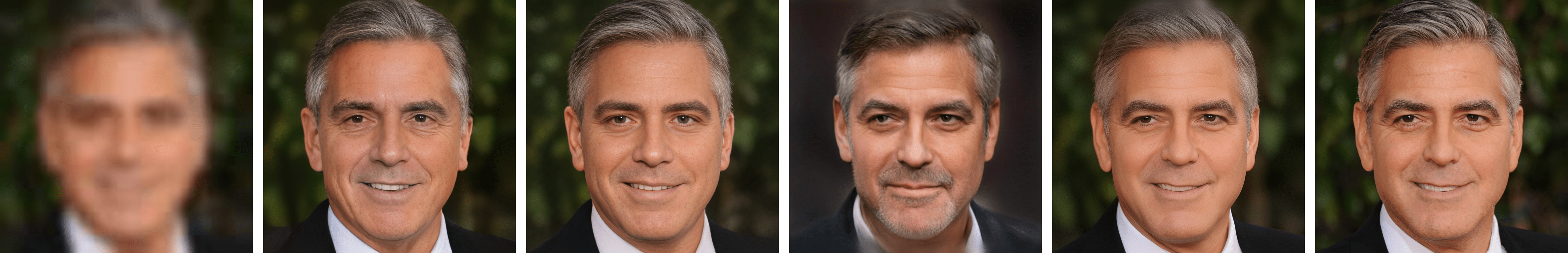}\\
   \includegraphics[width=\linewidth]{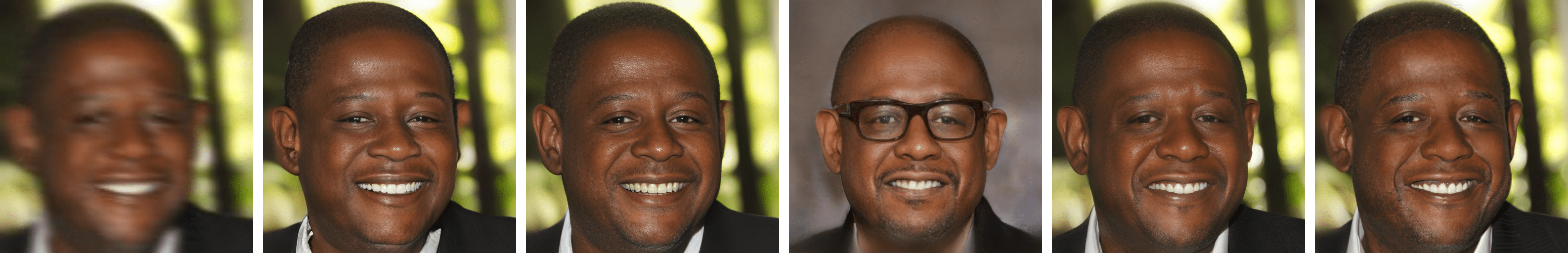}\\
   \includegraphics[width=\linewidth]{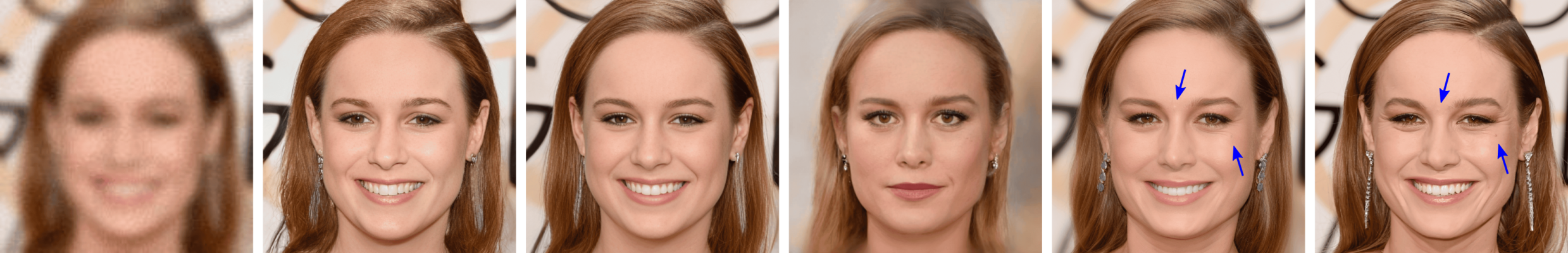}\\
   \includegraphics[width=\linewidth]{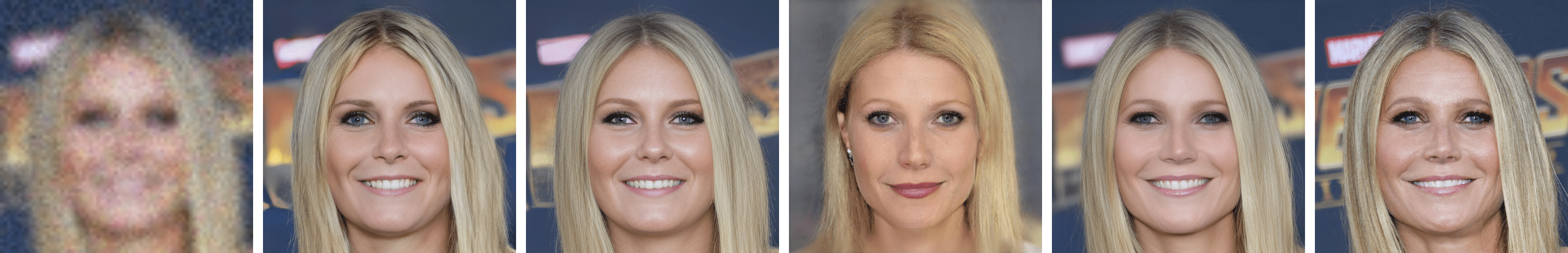}\\
   \includegraphics[width=\linewidth]{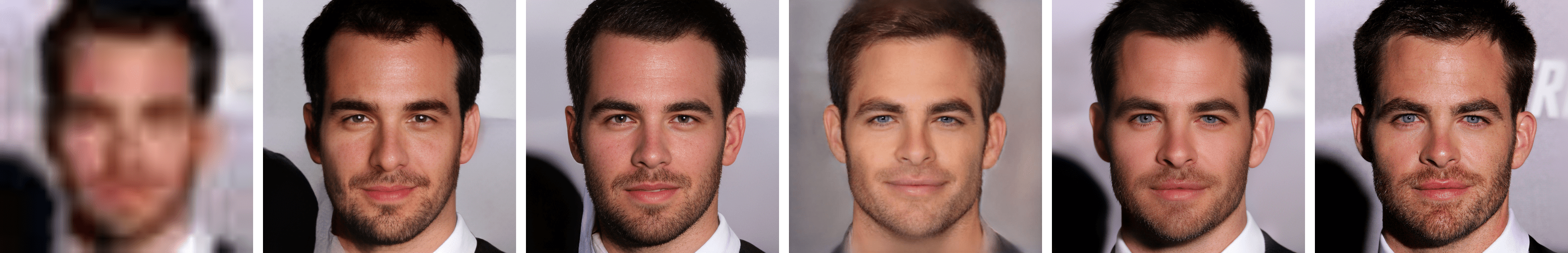}\\

   \begin{subfigure}[c]{0.16\textwidth}
        \centering    \caption{\footnotesize{\textsc{Degraded Image}}}
    \end{subfigure}
    \begin{subfigure}[c]{0.16\textwidth}
        \centering    \caption{\footnotesize{\textsc{GFPGAN~\cite{wang2021gfpgan}}}}
    \end{subfigure}
    \begin{subfigure}[c]{0.16\textwidth}
        \centering    \caption{\footnotesize{\textsc{CodeFormer~\cite{zhou2022towards}}}}
    \end{subfigure}
    \begin{subfigure}[c]{0.16\textwidth}
         \centering    \caption{\footnotesize{\textsc{MyStyle~\cite{nitzan2022mystyle}}}}
    \end{subfigure}
    \begin{subfigure}[c]{0.16\textwidth}
         \centering    \caption{\footnotesize{\textsc{Ours}}}
    \end{subfigure}
    \begin{subfigure}[c]{0.16\textwidth}
         \centering    \caption{\footnotesize{\textsc{Ground Truth}}}
    \end{subfigure}

   \caption{\textbf{Additional baseline methods on synthetically degraded images (in addition to the results in the main paper).} Similar observations to the synthetically degraded results can be made. The blind image restoration methods suffer both on identity and quality, while MyStyle is able to retain identity at the cost of losing faithfulness to the input image. Our method is able to retain identity in the restoration, while being faithful to the input image. In terms of specifics for each row, note the eye and eyebrow shape in row 1, eye shape, bags under the eyes, and teeth in row 2, eye shape and ears (specifically, left ear) in row 3, mole on the left cheek and mark between the eyes in row 4, eyes and mouth expression in row 5, and eye color in row 6. Please zoom in to observe these details more easily.}
   \label{fig:synth_deg_supp}
\end{figure*}

We next describe qualitative performance in comparision with GFPGAN~\cite{wang2021gfpgan} and CodeFormer~\cite{zhou2022towards} as blind image restoration methods, and MyStyle~\cite{nitzan2022mystyle} as a method with a personalized generative prior, in additional detail to the paper. \cref{fig:synth_deg_supp} shows results on synthetic degradations and corresponding restorations across six different identities. As can be seen, the blind restoration methods show clear drifts in identity across all six examples, while retaining fidelity to the input image. On the other hand,~\citet{nitzan2022mystyle}, while being able to retain a strong identity prior, sees a significant deviation in the restored image, from the input degraded image (potentially as a result of personalization on a small number of images (10)). In contrast, our method is able to achieve the best of both worlds: while maintaining a high degreee of faithfulness to the input image, we also see consistent identity retention across all examples. This leads to our results being closest to the reference image, despite the input having severe degradation in several cases.

\begin{figure*}[t]
  \centering

   \includegraphics[width=0.16\linewidth]{parent/figures/f_real_deg1/F_r1.png}
   \includegraphics[width=0.16\linewidth]{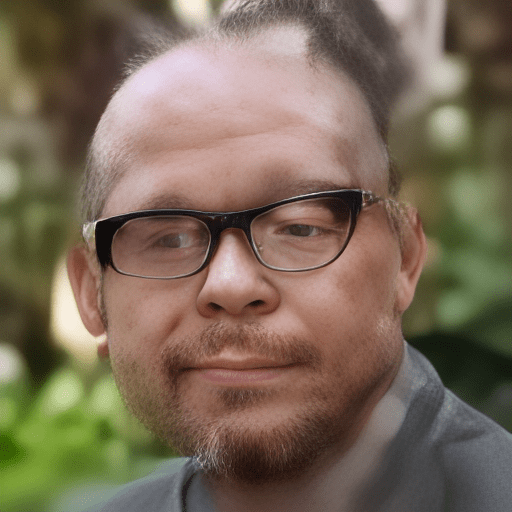}
   \includegraphics[width=0.16\linewidth]{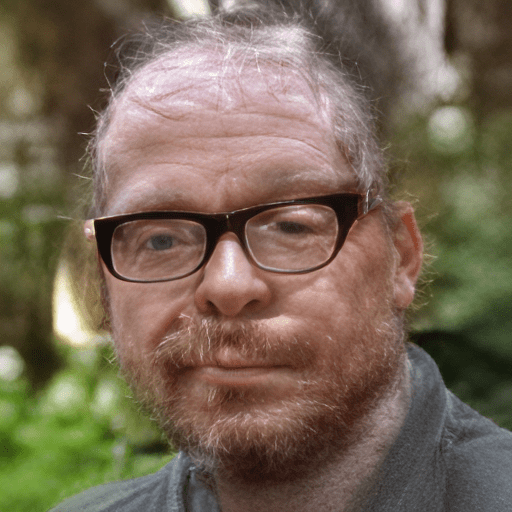}
   \includegraphics[width=0.16\linewidth]{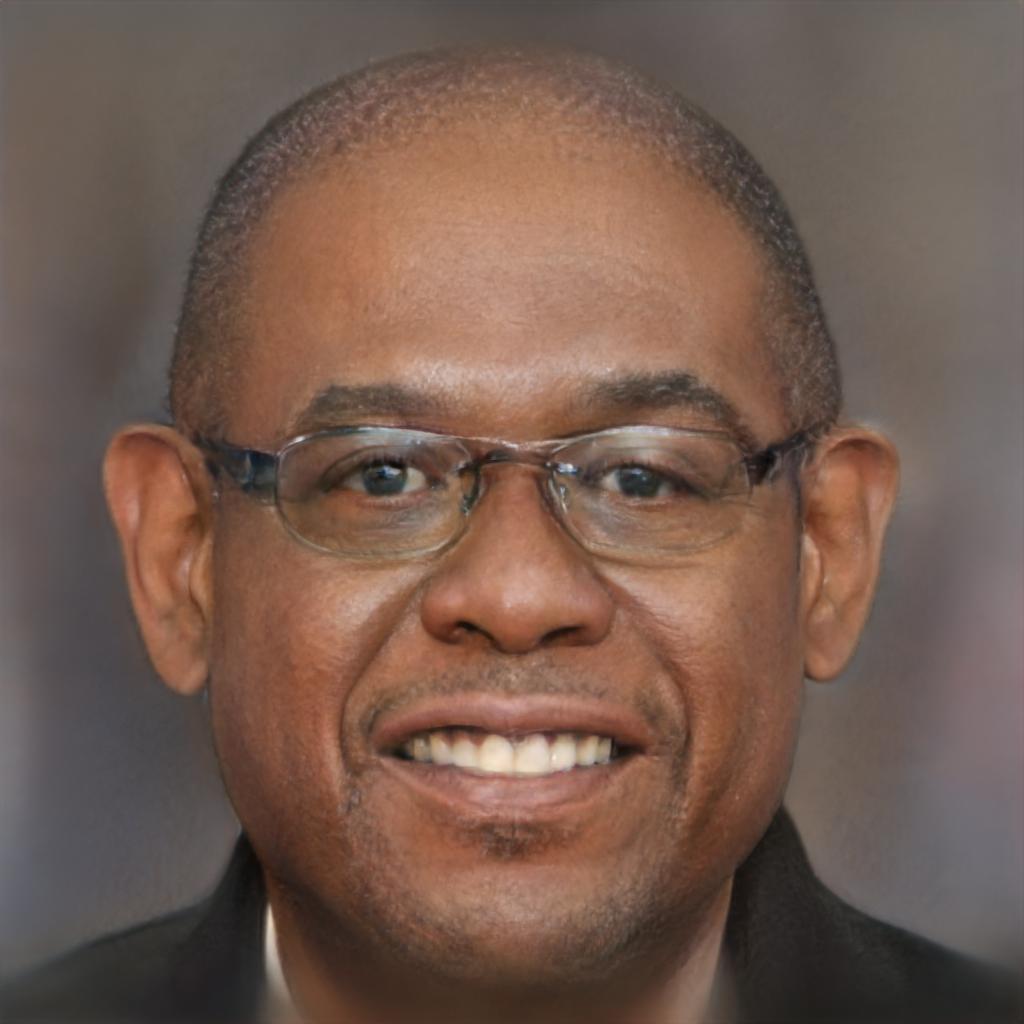}
   \includegraphics[width=0.16\linewidth]{parent/figures/f_real_deg1/F_r1_ours.png}
   \includegraphics[width=0.16\linewidth]{parent/figures/f_real_deg1/F_r1_refGT.png}\\

    \includegraphics[width=0.16\linewidth]{parent/figures/f_real_deg1/B_r1.png}
    \includegraphics[width=0.16\linewidth]{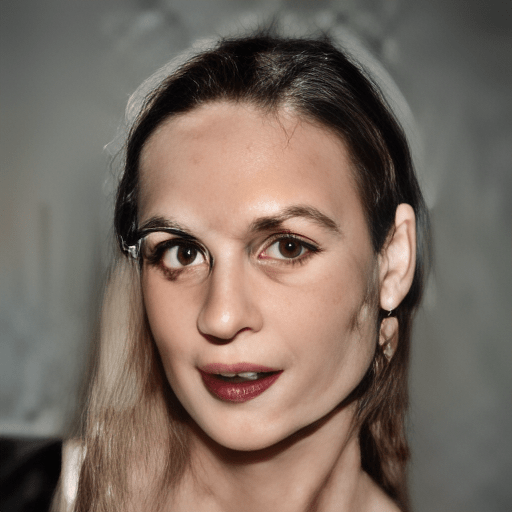}
   \includegraphics[width=0.16\linewidth]{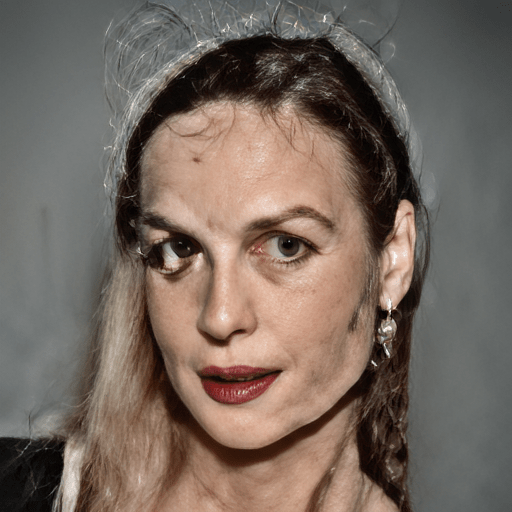}
   \includegraphics[width=0.16\linewidth]{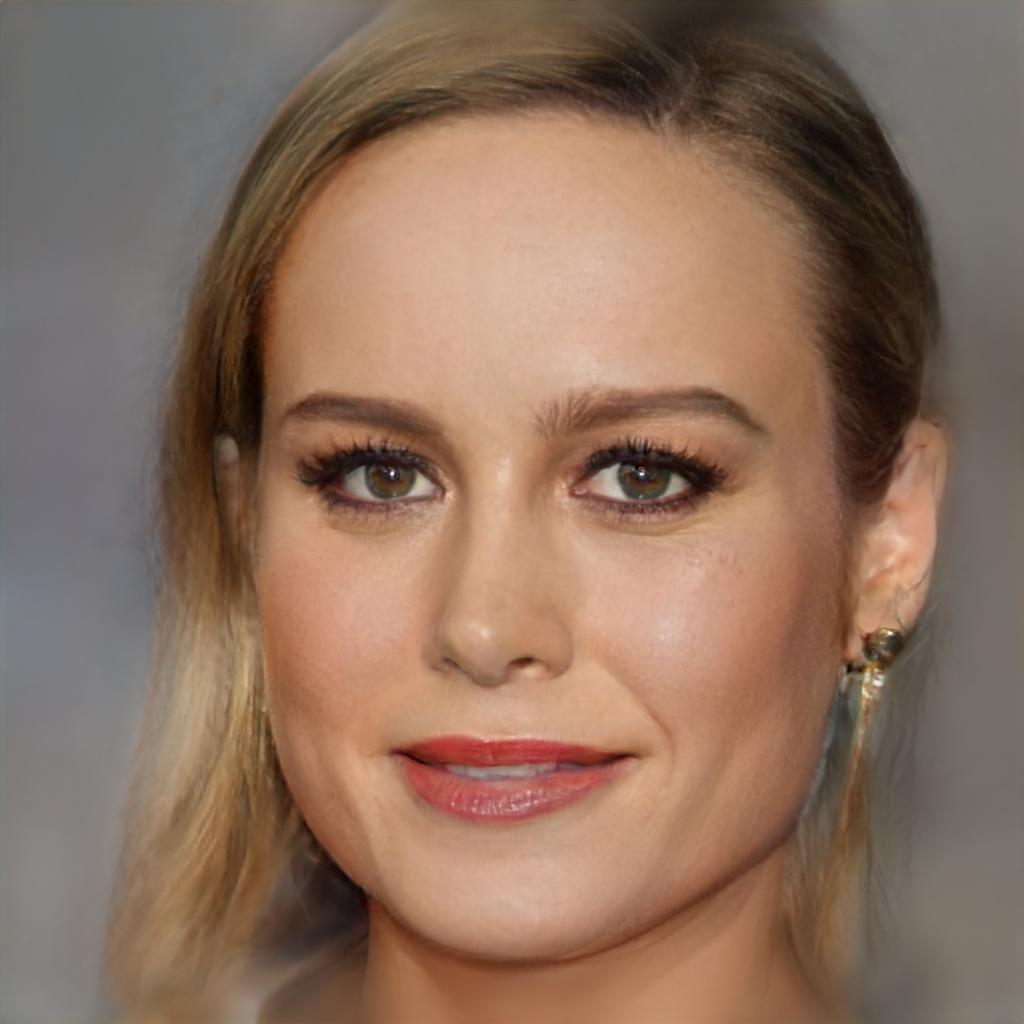}
   \includegraphics[width=0.16\linewidth]{parent/figures/f_real_deg1/B_r1_ours.png}
   \includegraphics[width=0.16\linewidth]{parent/figures/f_real_deg1/B_r1_ref.png}\\

   \includegraphics[width=0.16\linewidth]{parent/figures/f_real_deg1/G-4ip.png}
    \includegraphics[width=0.16\linewidth]{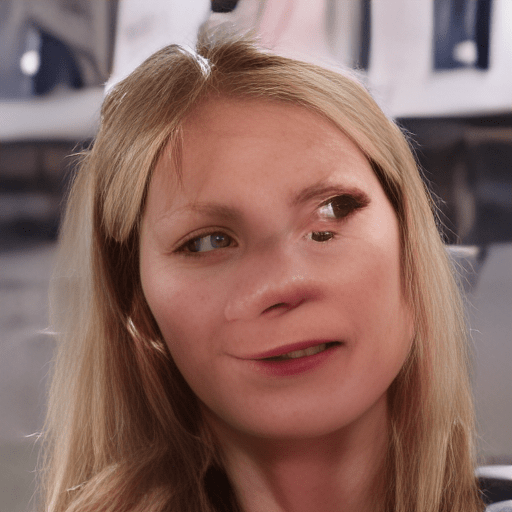}
   \includegraphics[width=0.16\linewidth]{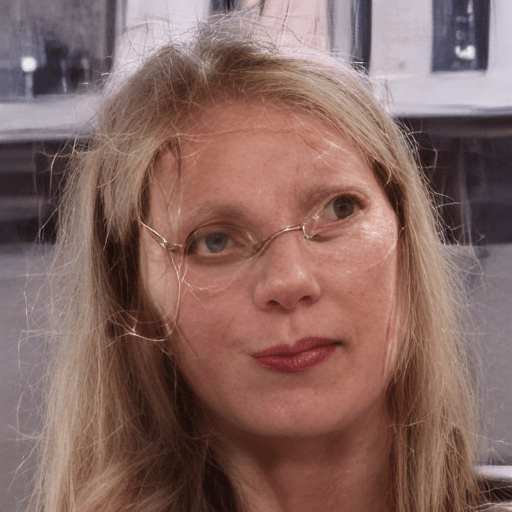}
   \includegraphics[width=0.16\linewidth]{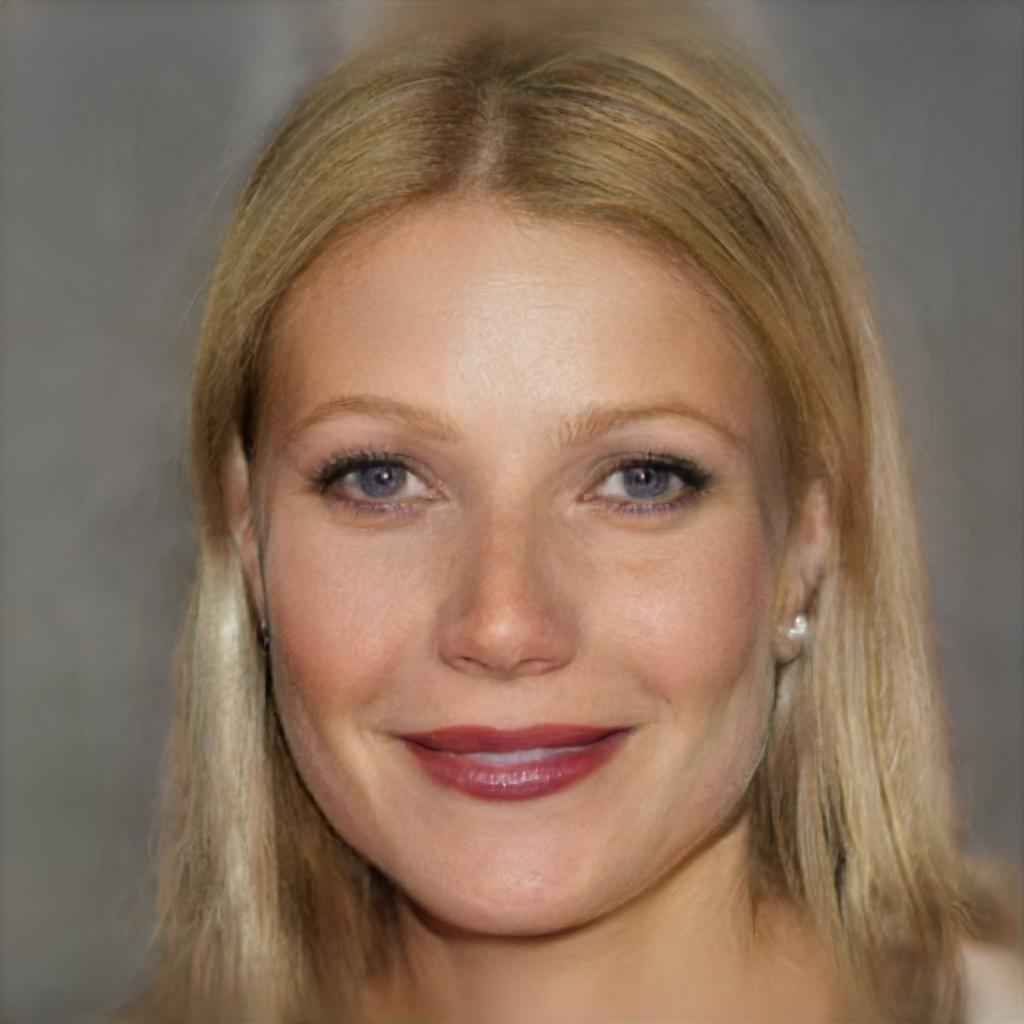}
   \includegraphics[width=0.16\linewidth]{parent/figures/f_real_deg1/G-4ours.png}
   \includegraphics[width=0.16\linewidth]{parent/figures/f_real_deg1/G_r1_ref.png}\\

   \begin{subfigure}[c]{0.16\textwidth}
        \centering    \caption{\footnotesize{\textsc{Degraded Image}}}
    \end{subfigure}
    \begin{subfigure}[c]{0.16\textwidth}
        \centering    \caption{\footnotesize{\textsc{GFPGAN~\cite{wang2021gfpgan}}}}
    \end{subfigure}
    \begin{subfigure}[c]{0.16\textwidth}
        \centering    \caption{\footnotesize{\textsc{CodeFormer~\cite{zhou2022towards}}}}
    \end{subfigure}
    \begin{subfigure}[c]{0.16\textwidth}
         \centering    \caption{\footnotesize{\textsc{MyStyle~\cite{nitzan2022mystyle}}}}
    \end{subfigure}
    \begin{subfigure}[c]{0.16\textwidth}
         \centering    \caption{\footnotesize{\textsc{Ours}}}
    \end{subfigure}
    \begin{subfigure}[c]{0.16\textwidth}
         \centering    \caption{\footnotesize{\textsc{Id. Reference}}}
    \end{subfigure}

   \caption{\textbf{Additional baseline methods on real degraded images (in addition to the results in the main paper).} It can be seen that even in images in the wild with real, unknown degradation kernels, our proposed method is superior to the baselines in terms of identity retention while maintaining high fidelity to the degraded input. Comparison methods either result in artifacts or in an image that is not faithful to the input, while the proposed method is able to stably inject relevant identity information while retaining fidelity to the input degraded image. Please zoom in to the Figure for a clearer view.
   }
   \label{fig:real_deg_supp}
\end{figure*}

We next look at \cref{fig:real_deg_supp}, for an analysis on real degraded images when compared with these additional baselines. Again, we note consistent observation. The blind restoration methods remain faithful to the input image, however they result in considerable identity drifts and artifacts in the restored images. MyStyle is able to continue retaining a strong personalized prior, however at the cost of losing all context relating to the input degraded image. Again, our method is able to retain identity, while being faithful to the input image.

\begin{figure*}[t]
  \centering
\begin{subfigure}[c]{0.32\textwidth}
\centering
    \begin{subfigure}[c]{0.32\textwidth}
        \includegraphics[width=\linewidth]{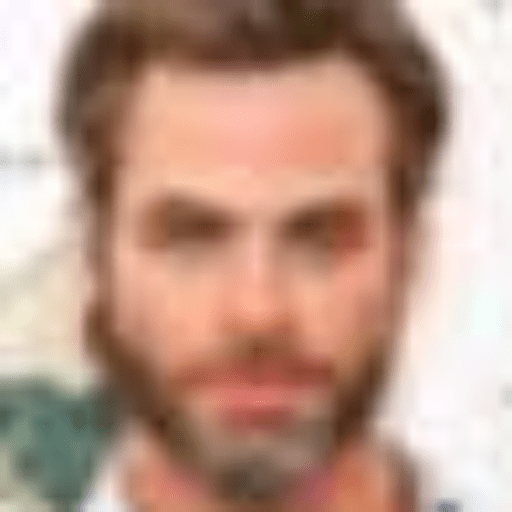}
        \centering    \caption*{\footnotesize{{Deg. Image}}}
    \end{subfigure}
    \begin{subfigure}[c]{0.32\textwidth}
        \includegraphics[width=\linewidth]{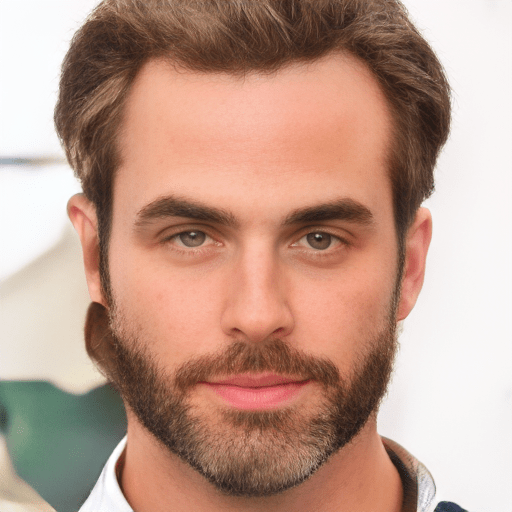}
        \centering    \caption*{\footnotesize{GFPGAN~\cite{wang2021gfpgan}}}
    \end{subfigure}
    \begin{subfigure}[c]{0.32\textwidth}
        \includegraphics[width=\linewidth]{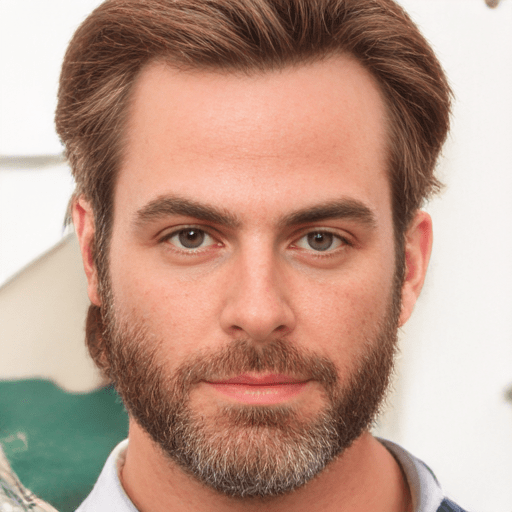}
        \centering    \caption*{\footnotesize{CdFrmer~\cite{zhou2022towards}}}
    \end{subfigure} \\


    \begin{subfigure}[c]{0.32\textwidth}
        \includegraphics[width=\linewidth]{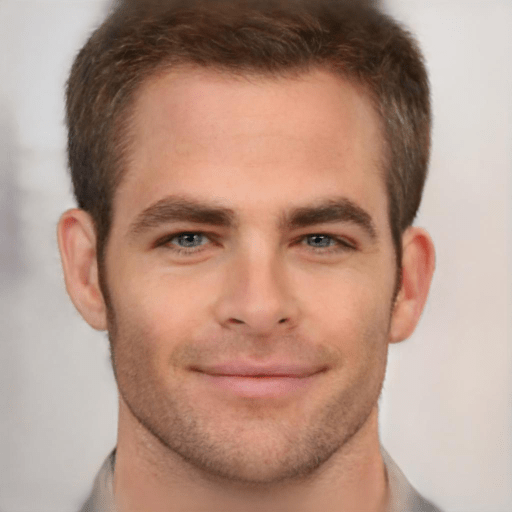}
        \centering    \caption*{\footnotesize{MyStyle~\cite{nitzan2022mystyle}}}
    \end{subfigure}
    \begin{subfigure}[c]{0.32\textwidth}
        \includegraphics[width=\linewidth]{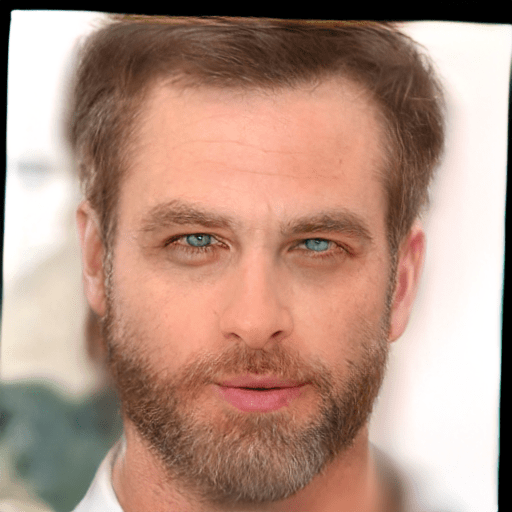}
        \centering    \caption*{\footnotesize{ASFFNet~\cite{Li_2020_CVPR}}}
    \end{subfigure}
    \begin{subfigure}[c]{0.32\textwidth}
        \includegraphics[width=\linewidth]{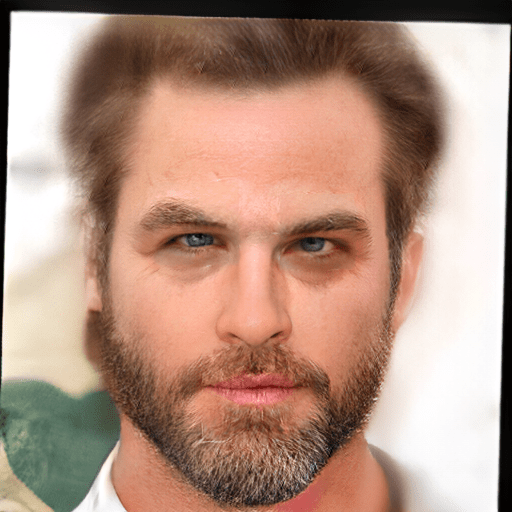}
        \centering    \caption*{\footnotesize{DMDNet~\cite{li2022learning}}}
    \end{subfigure}\\


    \begin{subfigure}[c]{0.32\textwidth}
        \includegraphics[width=\linewidth]{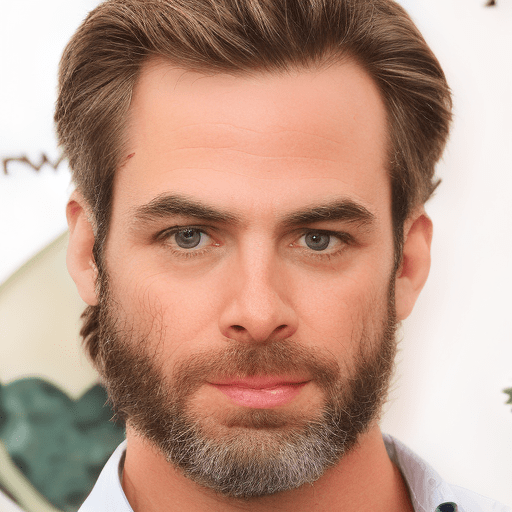}
        \centering    \caption*{\footnotesize{DiffBIR~\cite{lin2023diffbir}}}
    \end{subfigure}
    \begin{subfigure}[c]{0.32\textwidth}
        \includegraphics[width=\linewidth]{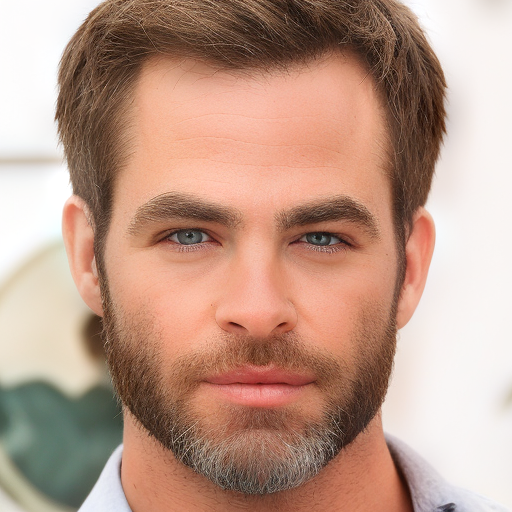}
        \centering    \caption*{\footnotesize{\textbf{Ours}}}
    \end{subfigure}
    \begin{subfigure}[c]{0.32\textwidth}
        \includegraphics[width=\linewidth]{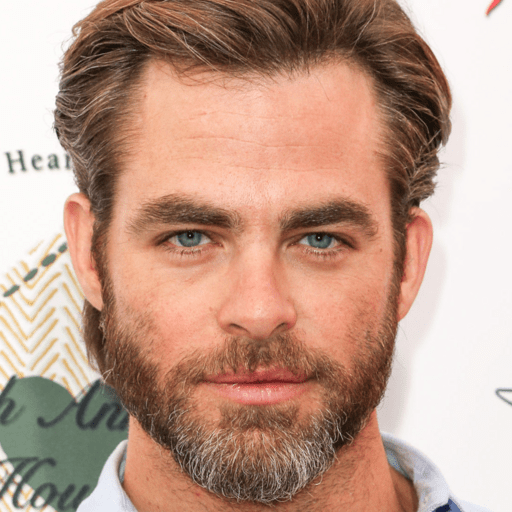}
        \centering    \caption*{\footnotesize{Id. Reference}}
    \end{subfigure} \\

        \caption{}
\end{subfigure}
\begin{subfigure}[c]{0.32\textwidth}
\centering
    \begin{subfigure}[c]{0.32\textwidth}
        \includegraphics[width=\linewidth]{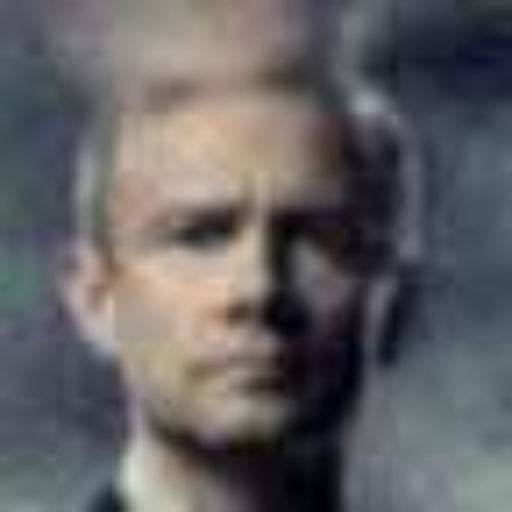}
        \centering    \caption*{\footnotesize{{Deg. Image}}}
    \end{subfigure}
    \begin{subfigure}[c]{0.32\textwidth}
        \includegraphics[width=\linewidth]{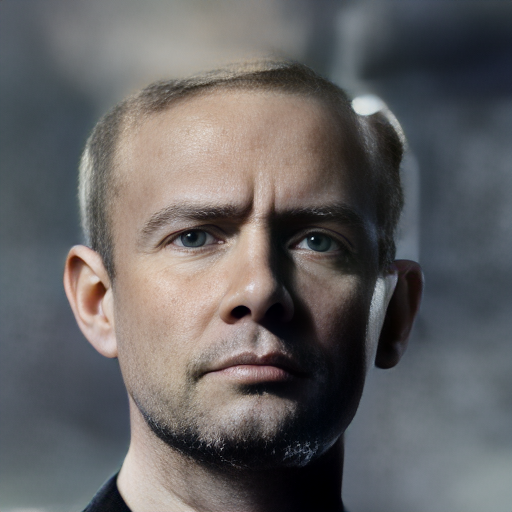}
        \centering    \caption*{\footnotesize{GFPGAN~\cite{wang2021gfpgan}}}
    \end{subfigure}
    \begin{subfigure}[c]{0.32\textwidth}
        \includegraphics[width=\linewidth]{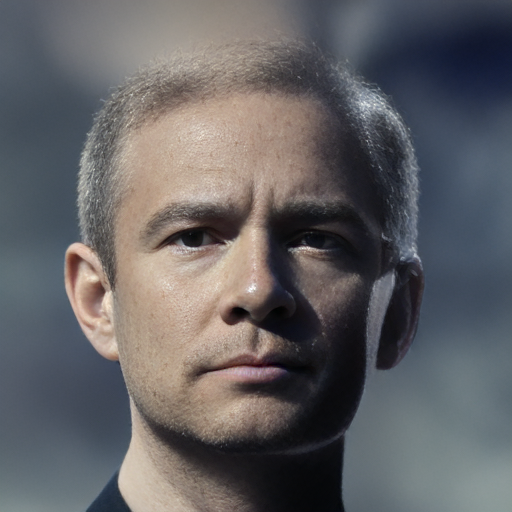}
        \centering    \caption*{\footnotesize{CdFrmer~\cite{zhou2022towards}}}
    \end{subfigure} \\


    \begin{subfigure}[c]{0.32\textwidth}
        \includegraphics[width=\linewidth]{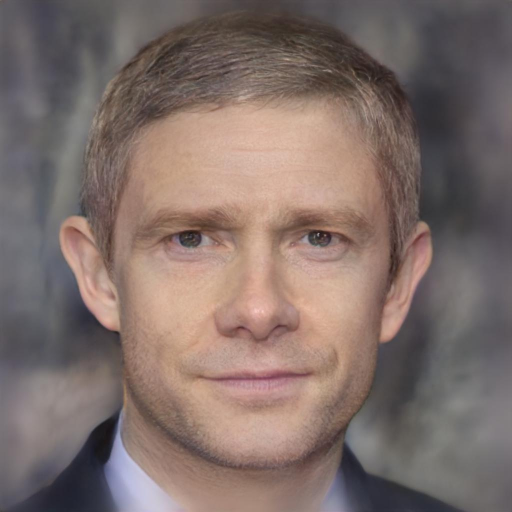}
        \centering    \caption*{\footnotesize{MyStyle~\cite{nitzan2022mystyle}}}
    \end{subfigure}
    \begin{subfigure}[c]{0.32\textwidth}
        \includegraphics[width=\linewidth]{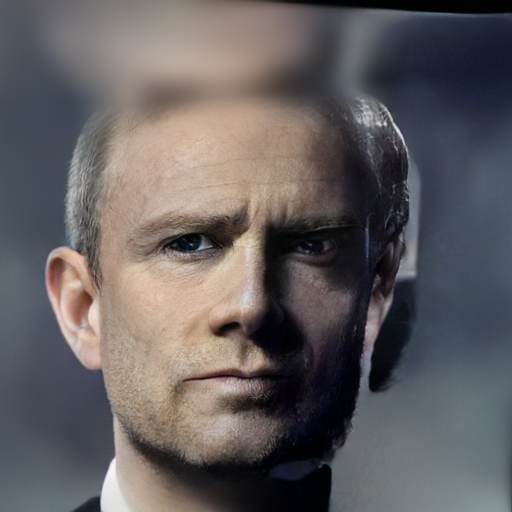}
        \centering    \caption*{\footnotesize{ASFFNet~\cite{Li_2020_CVPR}}}
    \end{subfigure}
    \begin{subfigure}[c]{0.32\textwidth}
        \includegraphics[width=\linewidth]{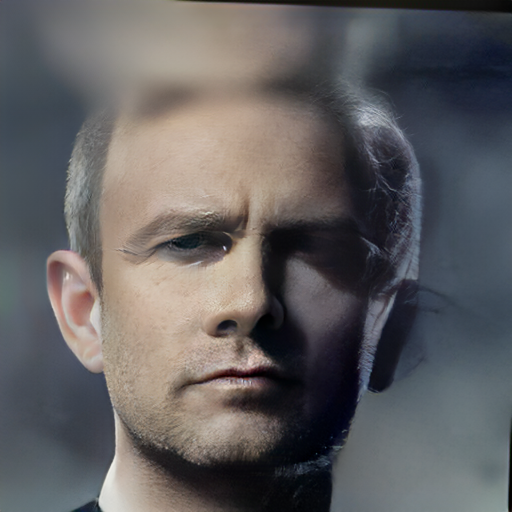}
        \centering    \caption*{\footnotesize{DMDNet~\cite{li2022learning}}}
    \end{subfigure}\\


    \begin{subfigure}[c]{0.32\textwidth}
        \includegraphics[width=\linewidth]{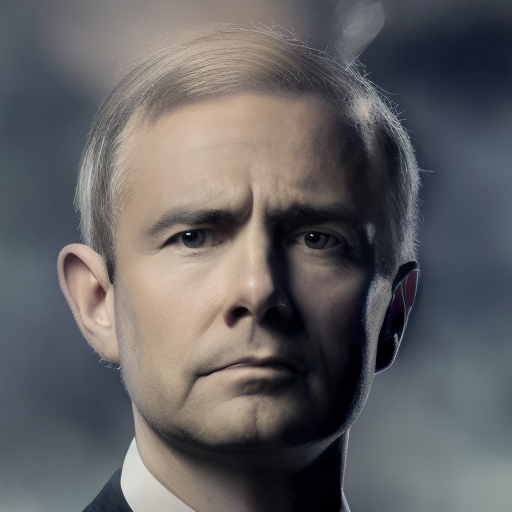}
        \centering    \caption*{\footnotesize{DiffBIR~\cite{lin2023diffbir}}}
    \end{subfigure}
    \begin{subfigure}[c]{0.32\textwidth}
        \includegraphics[width=\linewidth]{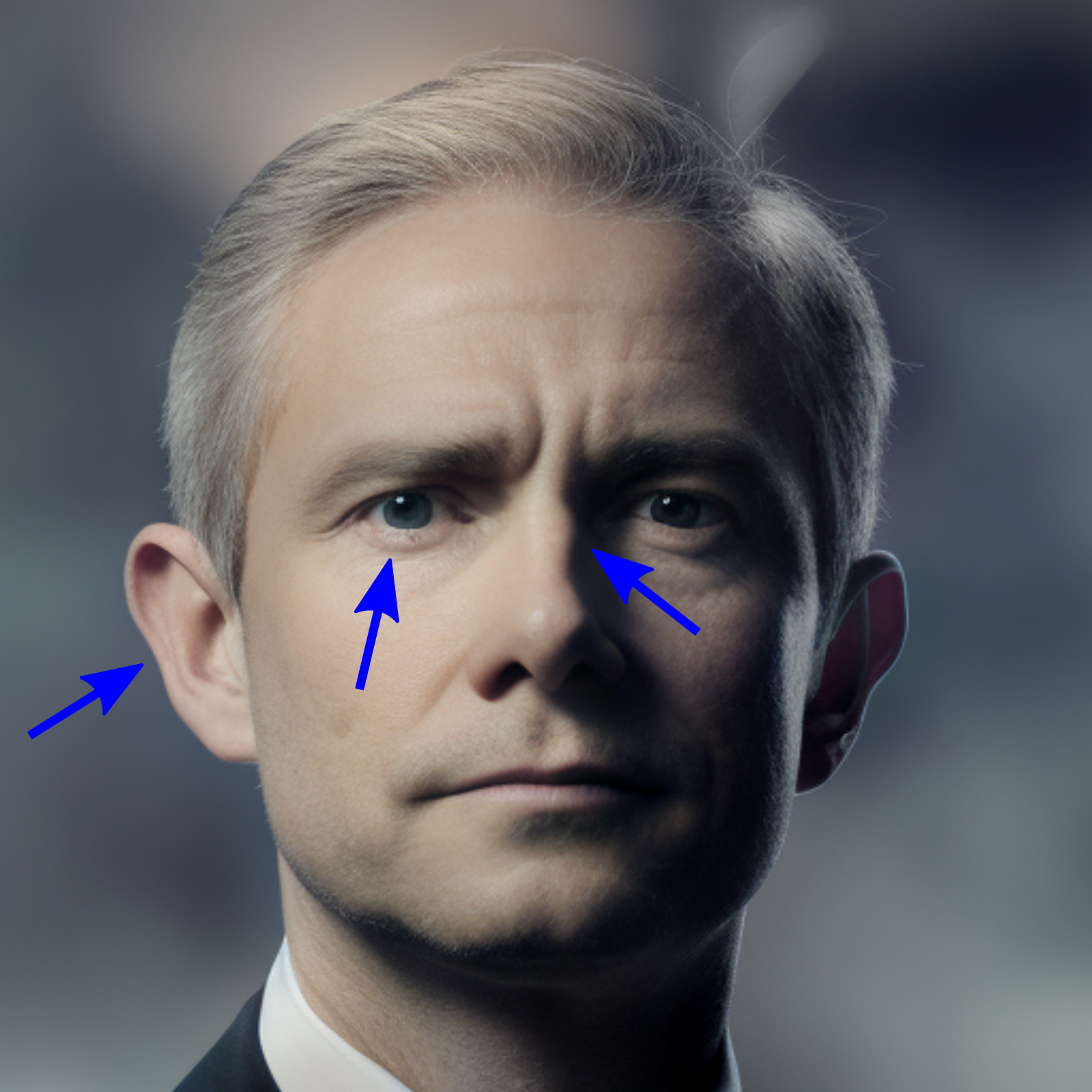}
        \centering    \caption*{\footnotesize{\textbf{Ours}}}
    \end{subfigure}
    \begin{subfigure}[c]{0.32\textwidth}
        \includegraphics[width=\linewidth]{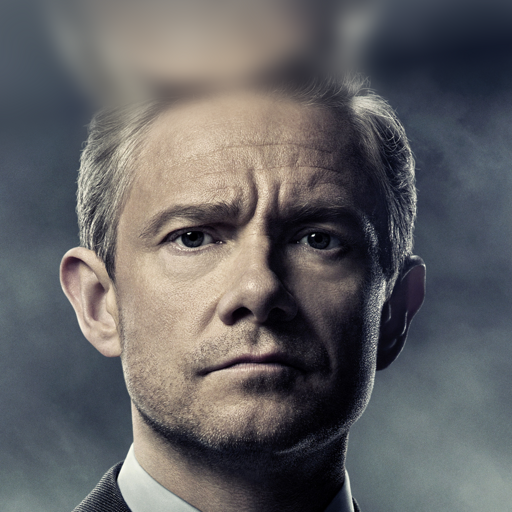}
        \centering    \caption*{\footnotesize{Id. Reference}}
    \end{subfigure} \\

        \caption{}
\end{subfigure}
\begin{subfigure}[c]{0.32\textwidth}
\centering
    \begin{subfigure}[c]{0.32\textwidth}
        \includegraphics[width=\linewidth]{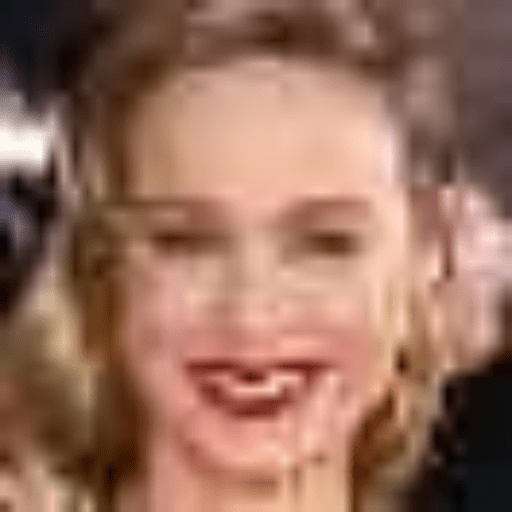}
        \centering    \caption*{\footnotesize{{Deg. Image}}}
    \end{subfigure}
    \begin{subfigure}[c]{0.32\textwidth}
        \includegraphics[width=\linewidth]{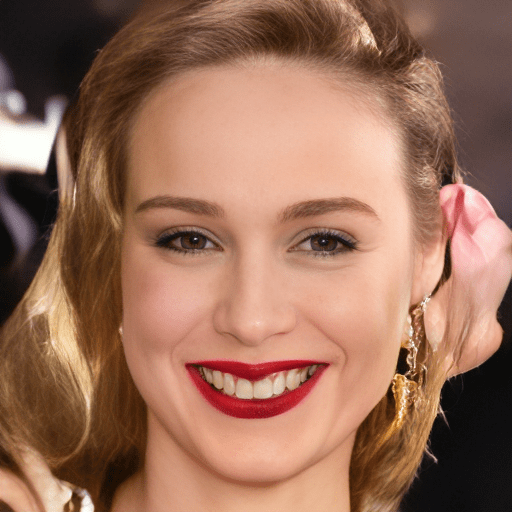}
        \centering    \caption*{\footnotesize{GFPGAN~\cite{wang2021gfpgan}}}
    \end{subfigure}
    \begin{subfigure}[c]{0.32\textwidth}
        \includegraphics[width=\linewidth]{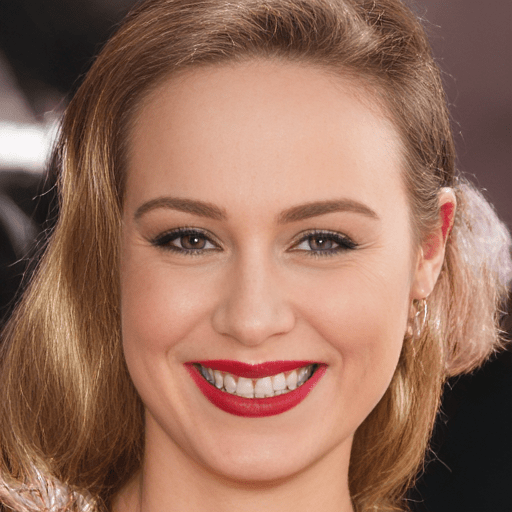}
        \centering    \caption*{\footnotesize{CdFrmer~\cite{zhou2022towards}}}
    \end{subfigure} \\


    \begin{subfigure}[c]{0.32\textwidth}
        \includegraphics[width=\linewidth]{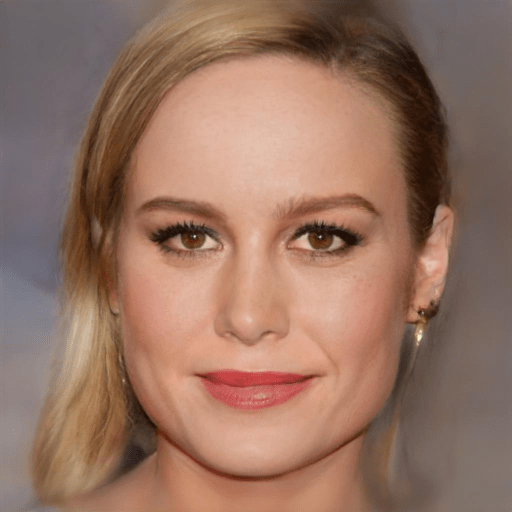}
        \centering    \caption*{\footnotesize{MyStyle~\cite{nitzan2022mystyle}}}
    \end{subfigure}
    \begin{subfigure}[c]{0.32\textwidth}
        \includegraphics[width=\linewidth]{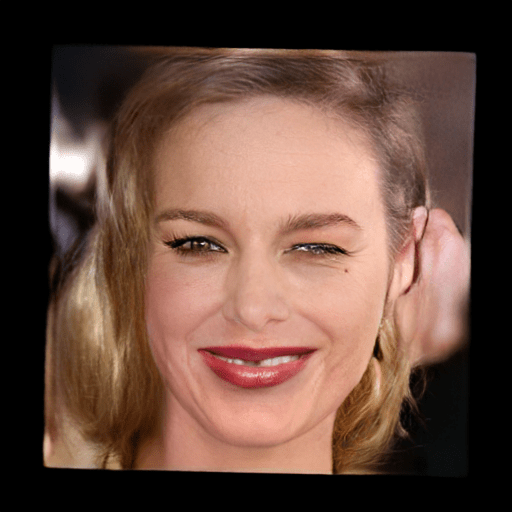}
        \centering    \caption*{\footnotesize{ASFFNet~\cite{Li_2020_CVPR}}}
    \end{subfigure}
    \begin{subfigure}[c]{0.32\textwidth}
        \includegraphics[width=\linewidth]{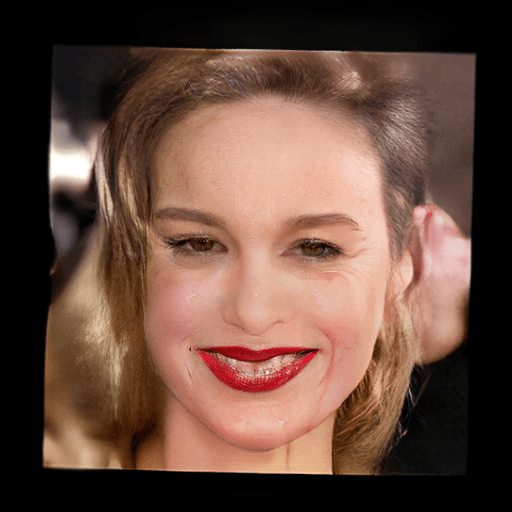}
        \centering    \caption*{\footnotesize{DMDNet~\cite{li2022learning}}}
    \end{subfigure}\\


    \begin{subfigure}[c]{0.32\textwidth}
        \includegraphics[width=\linewidth]{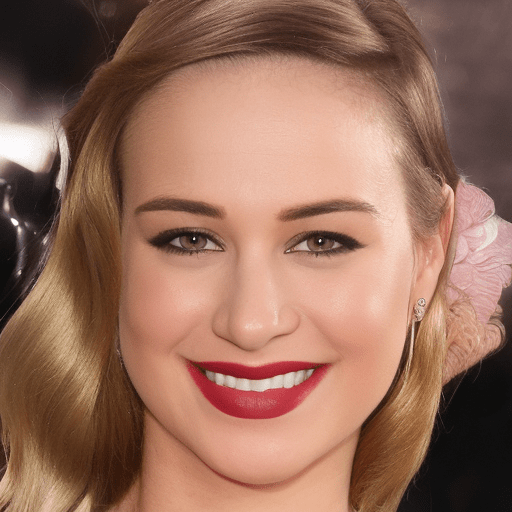}
        \centering    \caption*{\footnotesize{DiffBIR~\cite{lin2023diffbir}}}
    \end{subfigure}
    \begin{subfigure}[c]{0.32\textwidth}
        \includegraphics[width=\linewidth]{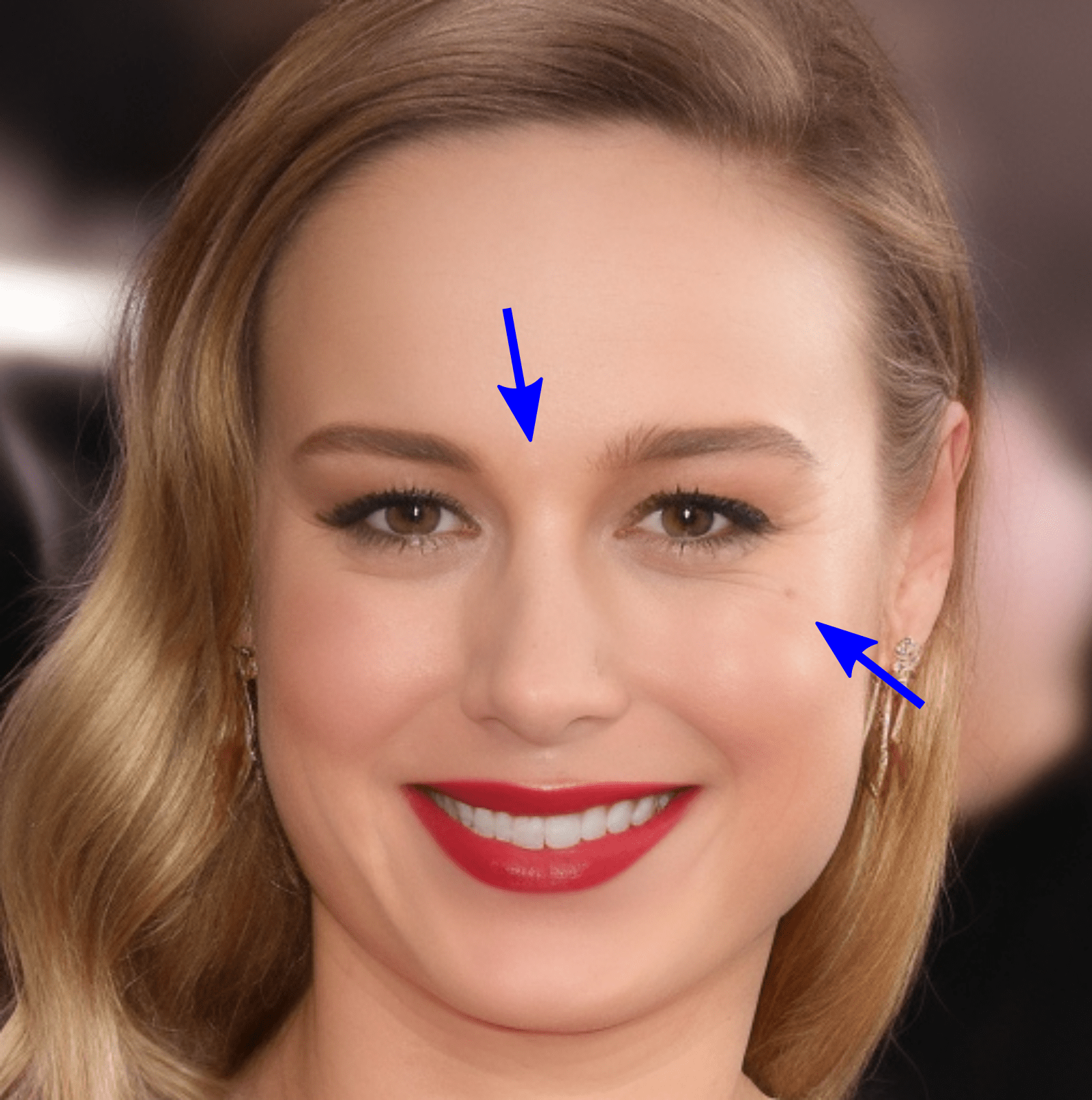}
        \centering    \caption*{\footnotesize{\textbf{Ours}}}
    \end{subfigure}
    \begin{subfigure}[c]{0.32\textwidth}
        \includegraphics[width=\linewidth]{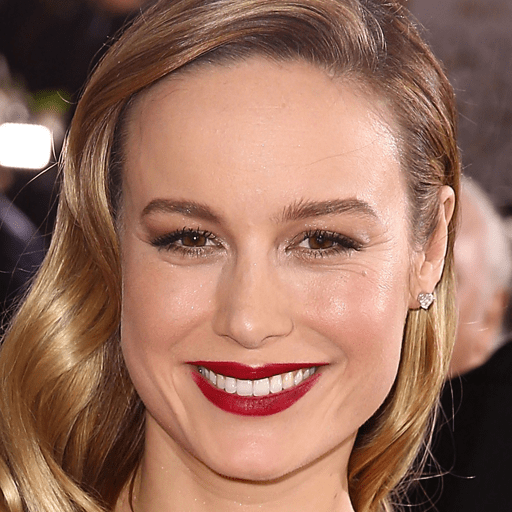}
        \centering    \caption*{\footnotesize{Id. Reference}}
    \end{subfigure} \\

        \caption{}
\end{subfigure}

   \caption{\textbf{Additional results on synthetic degraded images, comparing with all discussed major comparison methods.} For identity (a), zooming in shows that all baseline methods either lead to considerable artifacts in the restored image, or through identity drifts (eye color) and lack of fidelity with the input image. For identity (b) in cases where the restored image does not have significant artifacts, identity drifts can be noted in the form of nose shape (as highlighted by the shadow on the nose), eye color and ear shape. For identity (c), the mark between the eyes and the mole on the left cheek are identifying features that show the superiority of the proposed method.
   }
   \label{fig:synth_deg_supp_add_comb}
\end{figure*}

\begin{figure*}[t]
  \centering

    \begin{subfigure}[c]{0.16\textwidth}
        \includegraphics[width=\linewidth]{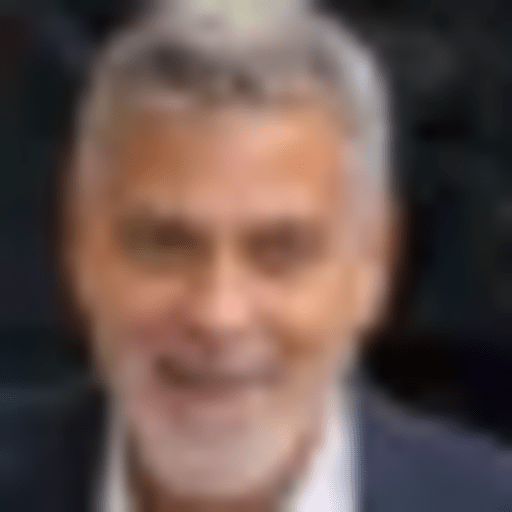}
        \centering    \caption*{\footnotesize{{Degraded Image}}}
    \end{subfigure}
    \begin{subfigure}[c]{0.16\textwidth}
        \includegraphics[width=\linewidth]{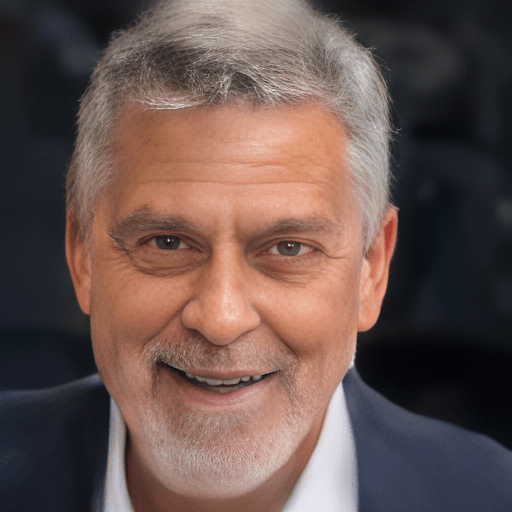}
        \centering    \caption*{\footnotesize{GFPGAN~\cite{wang2021gfpgan}}}
    \end{subfigure}
    \begin{subfigure}[c]{0.16\textwidth}
        \includegraphics[width=\linewidth]{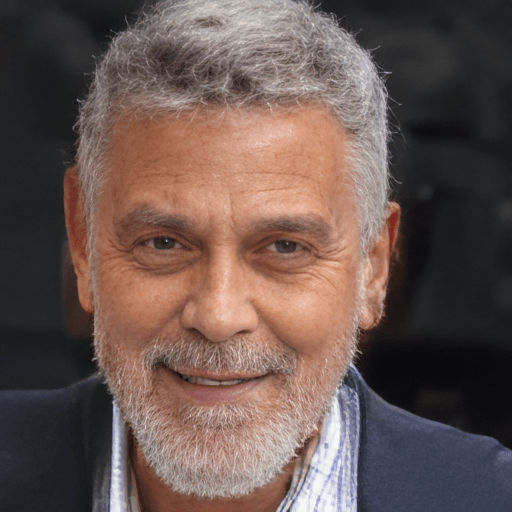}
        \centering    \caption*{\footnotesize{CodeFormer~\cite{zhou2022towards}}}
    \end{subfigure}
    \begin{subfigure}[c]{0.03\textwidth}
    \end{subfigure}
    \begin{subfigure}[c]{0.16\textwidth}
        \includegraphics[width=\linewidth]{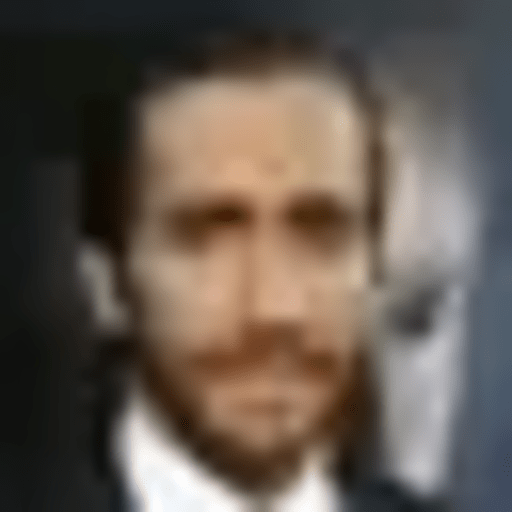}
        \centering    \caption*{\footnotesize{{Degraded Image}}}
    \end{subfigure}
    \begin{subfigure}[c]{0.16\textwidth}
        \includegraphics[width=\linewidth]{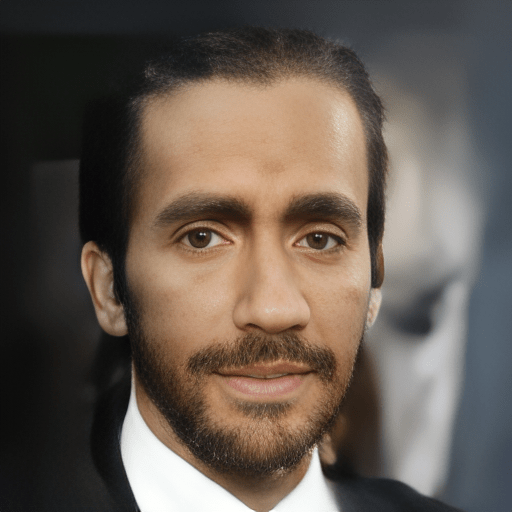}
        \centering    \caption*{\footnotesize{GFPGAN~\cite{wang2021gfpgan}}}
    \end{subfigure}
    \begin{subfigure}[c]{0.16\textwidth}
        \includegraphics[width=\linewidth]{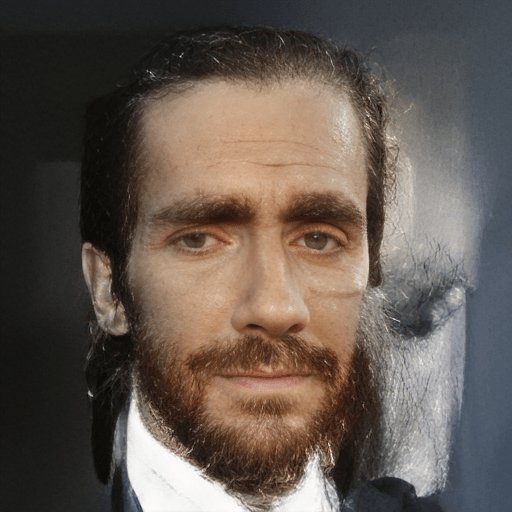}
        \centering    \caption*{\footnotesize{CodeFormer~\cite{zhou2022towards}}}
    \end{subfigure} \\


      \begin{subfigure}[c]{0.16\textwidth}
        \includegraphics[width=\linewidth]{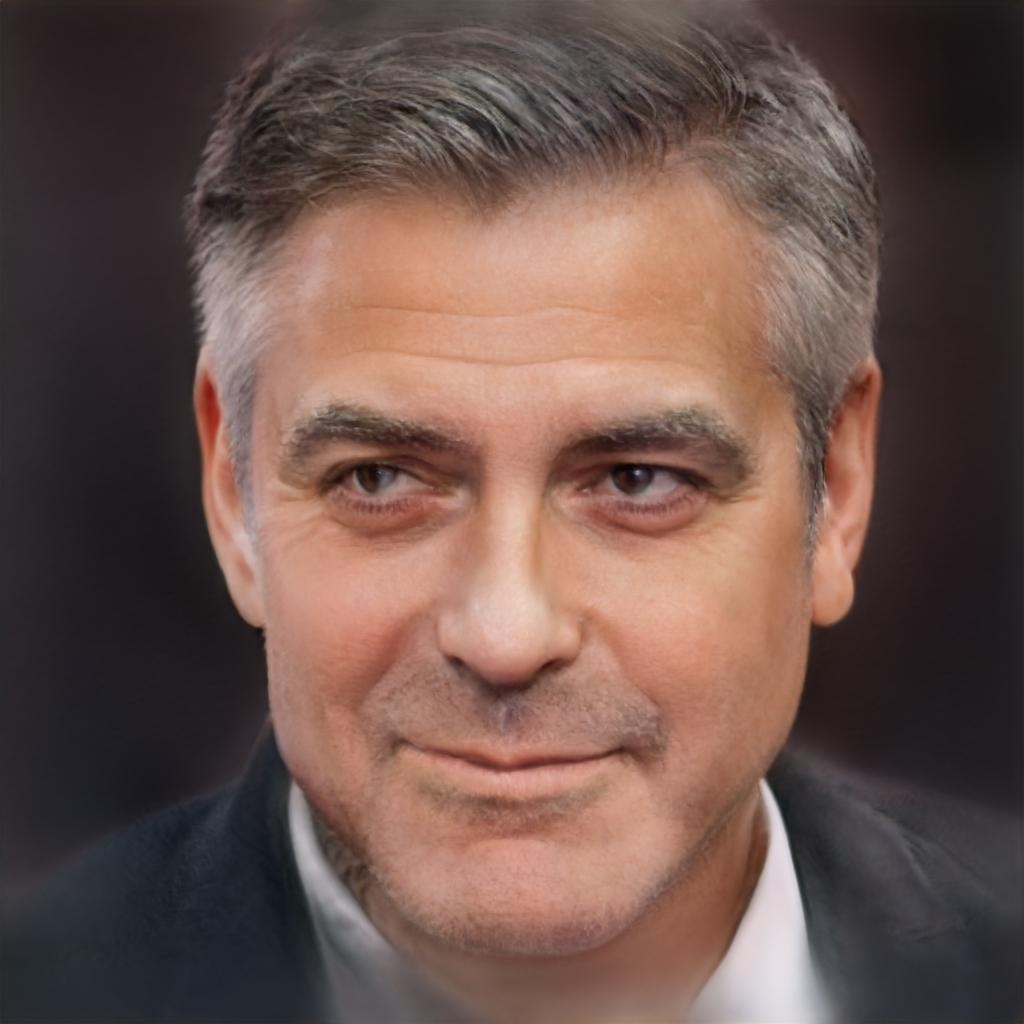}
        \centering    \caption*{\footnotesize{MyStyle~\cite{nitzan2022mystyle}}}
    \end{subfigure}
    \begin{subfigure}[c]{0.16\textwidth}
        \includegraphics[width=\linewidth]{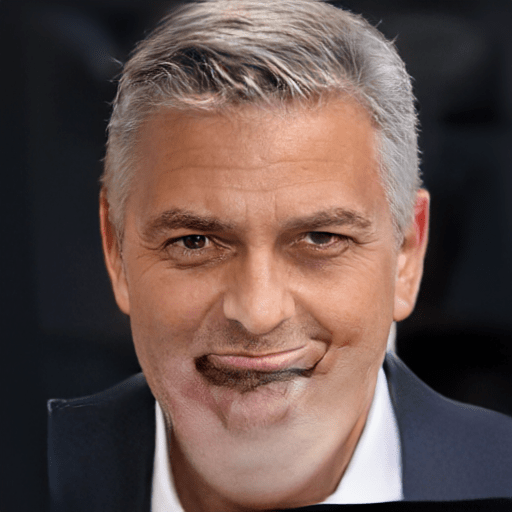}
        \centering    \caption*{\footnotesize{ASFFNet~\cite{Li_2020_CVPR}}}
    \end{subfigure}
    \begin{subfigure}[c]{0.16\textwidth}
        \includegraphics[width=\linewidth]{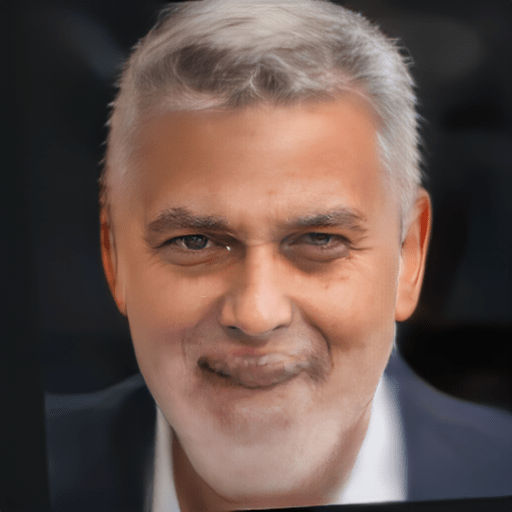}
        \centering    \caption*{\footnotesize{DMDNet~\cite{li2022learning}}}
    \end{subfigure}
    \begin{subfigure}[c]{0.03\textwidth}
    \end{subfigure}
    \begin{subfigure}[c]{0.16\textwidth}
        \includegraphics[width=\linewidth]{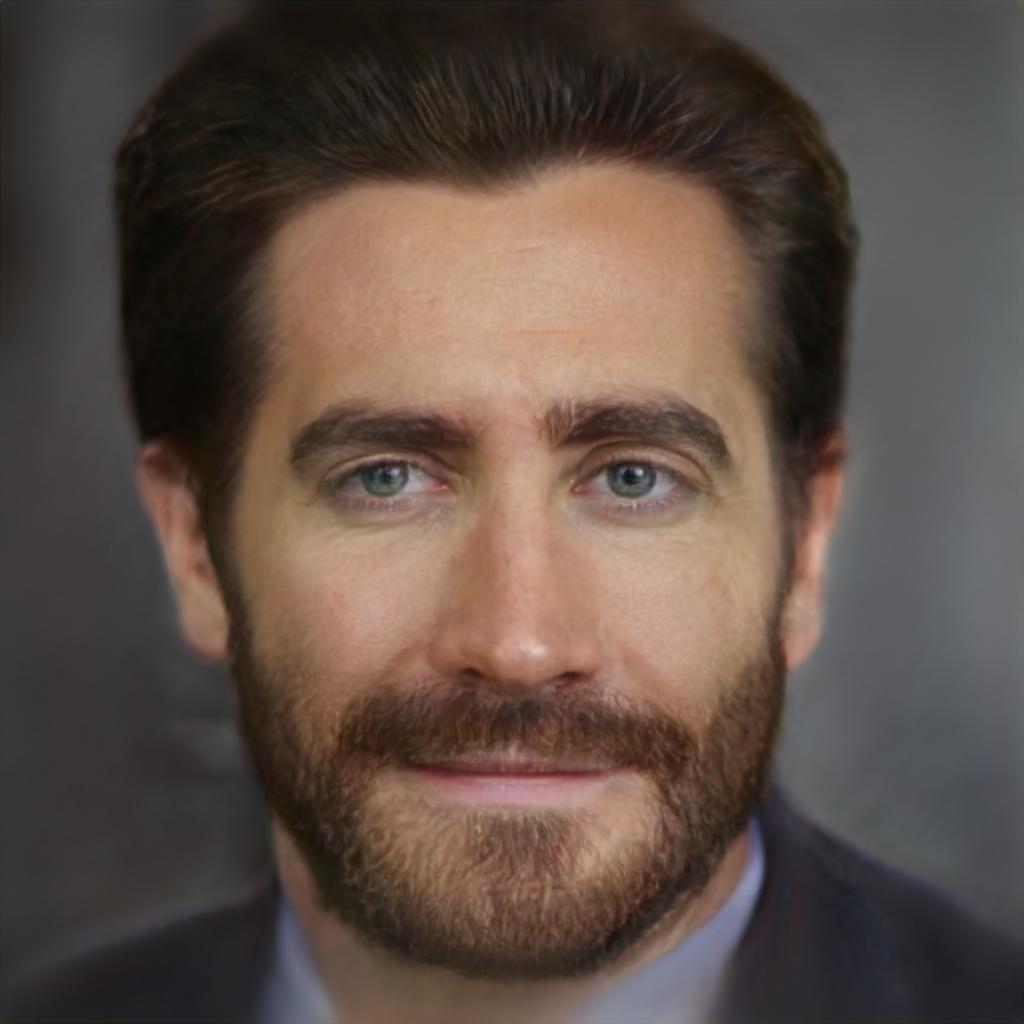}
        \centering    \caption*{\footnotesize{MyStyle~\cite{nitzan2022mystyle}}}
    \end{subfigure}
    \begin{subfigure}[c]{0.16\textwidth}
        \includegraphics[width=\linewidth]{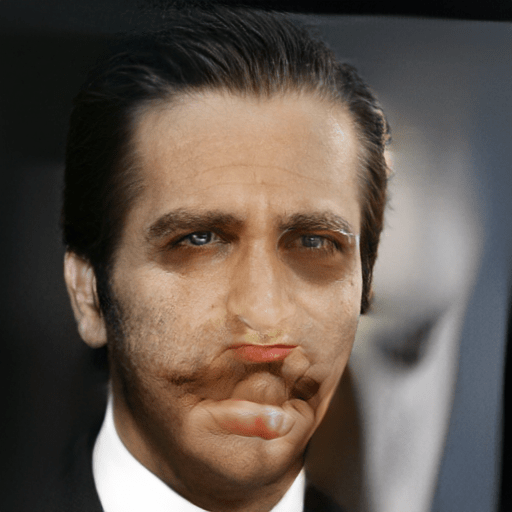}
        \centering    \caption*{\footnotesize{ASFFNet~\cite{Li_2020_CVPR}}}
    \end{subfigure}
    \begin{subfigure}[c]{0.16\textwidth}
        \includegraphics[width=\linewidth]{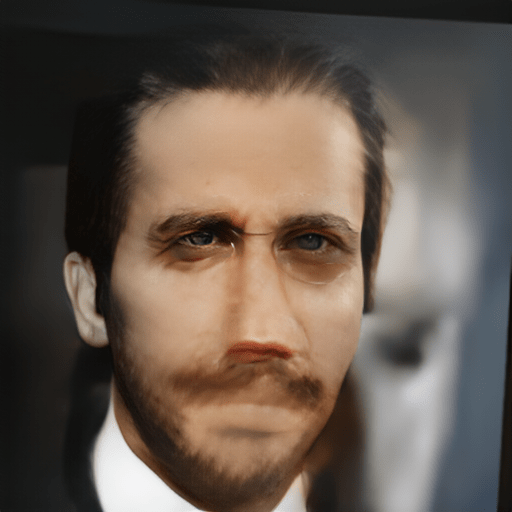}
        \centering    \caption*{\footnotesize{DMDNet~\cite{li2022learning}}}
    \end{subfigure}\\


    \begin{subfigure}[c]{0.16\textwidth}
        \includegraphics[width=\linewidth]{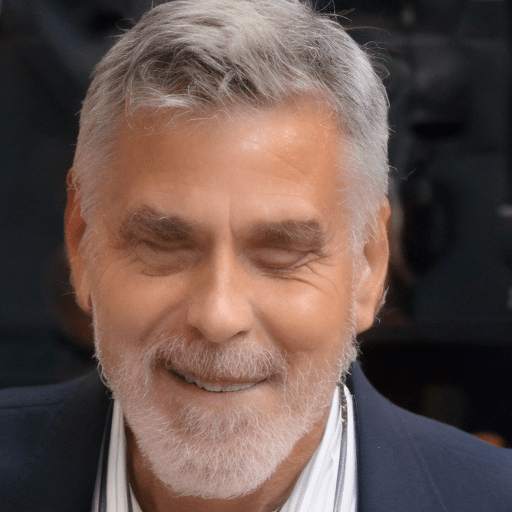}
        \centering    \caption*{\footnotesize{DiffBIR~\cite{lin2023diffbir}}}
    \end{subfigure}
    \begin{subfigure}[c]{0.16\textwidth}
        \includegraphics[width=\linewidth]{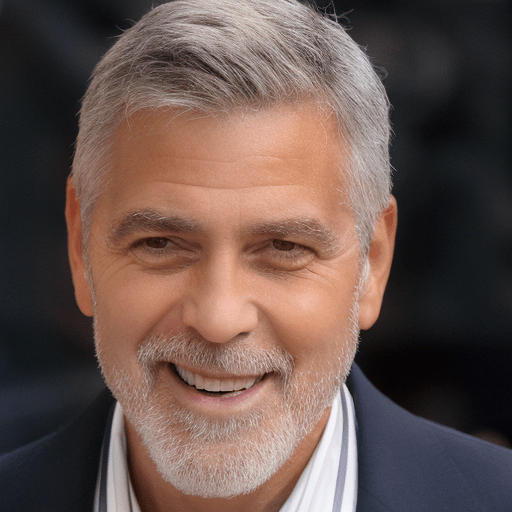}
        \centering    \caption*{\footnotesize{\textbf{Ours}}}
    \end{subfigure}
    \begin{subfigure}[c]{0.16\textwidth}
        \includegraphics[width=\linewidth]{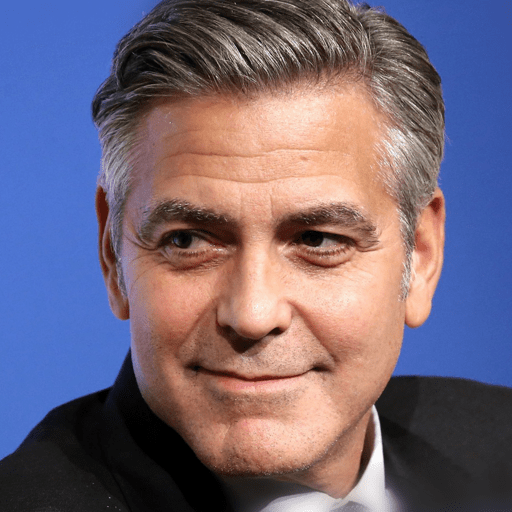}
        \centering    \caption*{\footnotesize{Id. Reference}}
    \end{subfigure}
    \begin{subfigure}[c]{0.03\textwidth}
    \end{subfigure}
    \begin{subfigure}[c]{0.16\textwidth}
        \includegraphics[width=\linewidth]{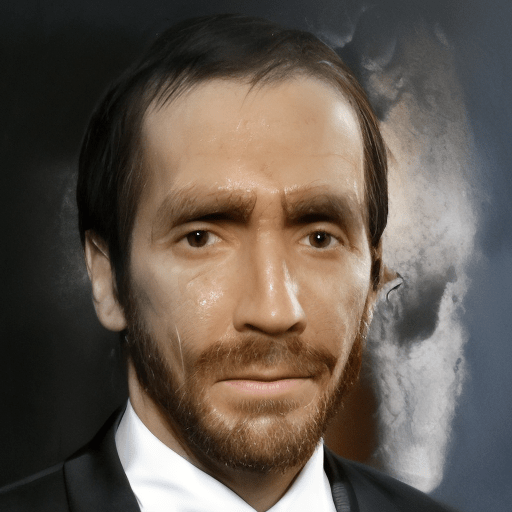}
        \centering    \caption*{\footnotesize{DiffBIR~\cite{lin2023diffbir}}}
    \end{subfigure}
    \begin{subfigure}[c]{0.16\textwidth}
        \includegraphics[width=\linewidth]{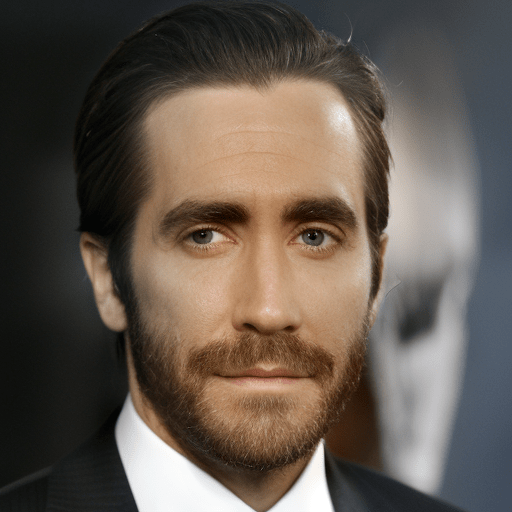}
        \centering    \caption*{\footnotesize{\textbf{Ours}}}
    \end{subfigure}
    \begin{subfigure}[c]{0.16\textwidth}
        \includegraphics[width=\linewidth]{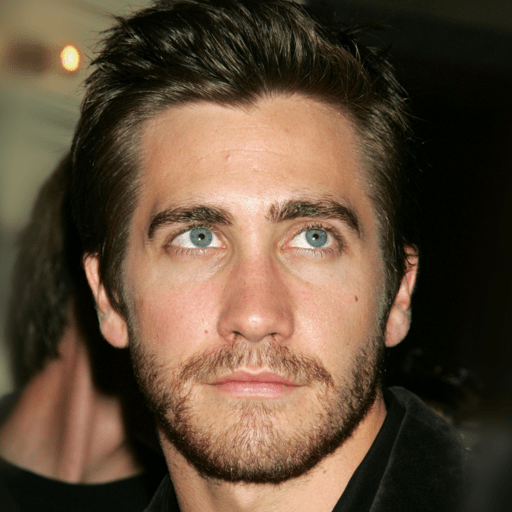}
        \centering    \caption*{\footnotesize{Id. Reference}}
    \end{subfigure} \\

   \caption{\textbf{Additional results on real degraded images, comparing with all discussed major comparison methods.} For both identities, zooming in shows that comparison methods (both prior reference-based methods and blind restoration methods) result in strong artifacts, identity drifts, and in the case of MyStyle~\cite{nitzan2022mystyle}, lack of fidelity with the input degraded image. The proposed method is the only one consistently able to incorporate identity in the restoration in a stable manner. Please zoom in to the image for a clearer visualization.
   }
   \label{fig:real_deg_supp_add_comb}
\end{figure*}

Further, we supply additional synthetic (\cref{fig:synth_deg_supp_add_comb}) as well as real (\cref{fig:real_deg_supp_add_comb}) degradation results across all baselines. We wish to highlight the robustness of our proposed method, across identites and degradations, over a larger number of image settings. These aspects can specifically be seen through identifying features in the participants, such as hair, teeth, ears, eye color and so on. The overall trends remain the same: prior methods are either able to retain strong identity without artifacts, or retaining faithfulness to the input image. It is through our personalzation regime that we achieve the best of both these worlds, getting identity-consistent restored images with high fidelity.

\section{Additional Observations}
\label{sec:additional_observations}
\paragraph{The Effect of Classifier Free Guidance.} The qualitative effect of classifier free guidance (CFG) was discussed in the main paper (Figure~8). Here, we perform a quantitative analysis, through \cref{tab:ablation}. We note that traditional fidelity metrics worsen as the CFG is increased. This is consistent with our other observations, where the unconditional method (DiffBIR) shows slightly better performance on these metrics. As we increase the CFG, we move farther from the unconditional model and therefore see these effects. In terms of identity, we see a small reduction with increased CFG. This is also consistent with our observations from the main paper. As we increase the CFG, the sharpness of the restored image increases and may lead it to look unrealistic. Overall, in terms of tradional metrics, a lower CFG values is optimal. However, visually, the CFG can serve as a useful control knob for restored image style and quality, as discussed in Figure~8, main paper.

\begin{figure}
    \centering
    \begin{subfigure}[c]{0.24\linewidth}
         \centering   
         \includegraphics[width=\linewidth]{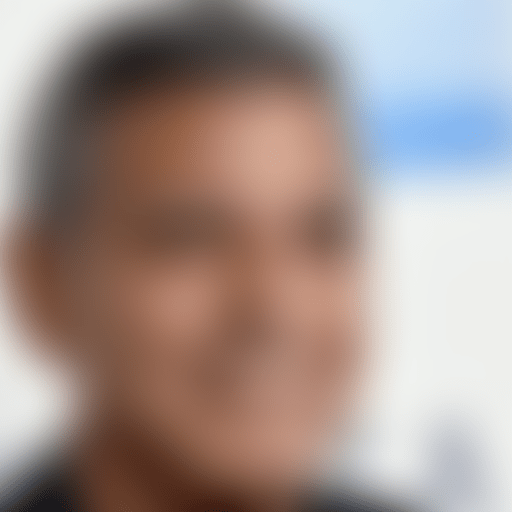}
         \caption{}
    \end{subfigure}  
    \begin{subfigure}[c]{0.24\linewidth}
         \centering   
         \includegraphics[width=\linewidth]{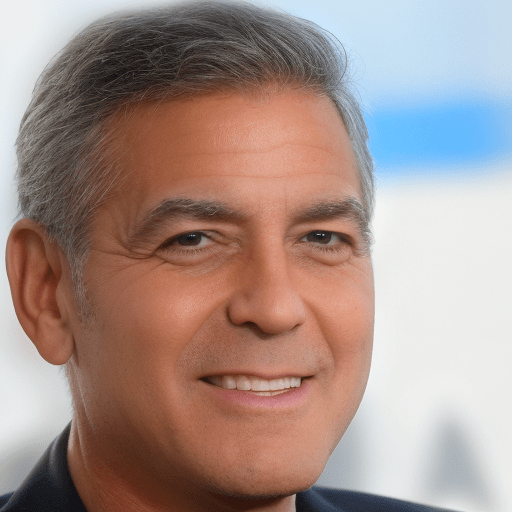}
         \caption{}
    \end{subfigure} 
    \begin{subfigure}[c]{0.24\linewidth}
         \centering   
         \includegraphics[width=\linewidth]{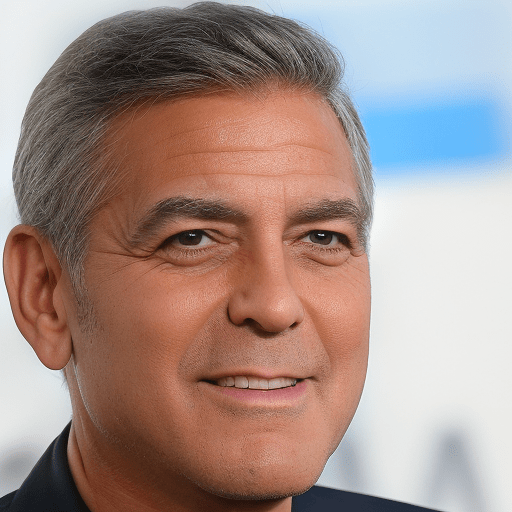}
         \caption{}
    \end{subfigure} 
    \begin{subfigure}[c]{0.24\linewidth}
         \centering   
         \includegraphics[width=\linewidth]{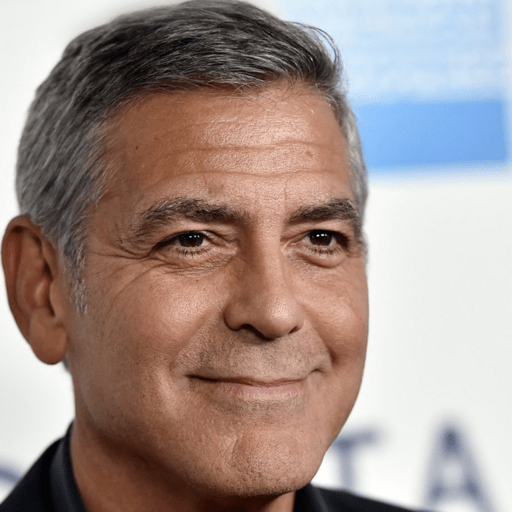}
         \caption{}
    \end{subfigure} 

    \caption{\textbf{Heavy degradation restoration.} Given a highly degraded image (a), a first pass (b) through our restoration pipeline provides a coarse estimate that pulls the image closer to the input domain of the model. The second pass (c) injects texture and identity information, leading to a better restored estimate with respect to the ground truth (d). }
    \label{fig:heavy_deg}

\end{figure}

\paragraph{Dealing with Heavy Degradations.} \cref{fig:heavy_deg} shows a potentially interesting use setting for the proposed method. Namely, in the case of very heavy degradations, multiple passes through the restoration model may be performed. As seen in the figure, (b) is able to obtain coarse details as well as shape and strcuture information. Through a second pass, (c) is able to improve of texutre and detail, in addtion to obtaining greater identity injection, leading to a better restored image than (b).

\section{Ethical Considerations}
\label{sec:ethical_considerations}
The use of generative models, while being extremely helpful for challenging tasks such as identity-aware image restoration, can also potentially have harmful effects. Specifically, generative models can be used for immoral tasks. In our case, applications that we discussed, such as face swapping as well as text-guided editing, can lead to generation of fake images, that may be used without the consent of the person whose image it is. We strongly condemn such and any other use cases of the proposed method.

\begin{table}[t]
    \caption{\textbf{Ablation: effect of classifier-free guidance scale on personalized image restoration}. We compare along PSNR and SSIM as fidelity metrics, while the ArcFace similarity serves as an identity metric.}
    \footnotesize
   
  \begin{center}
    \setlength{\tabcolsep}{0.01\columnwidth}
    
\resizebox{0.75\linewidth}{!}{
  \begin{tabular}{lccccc} 
    \toprule
    \textbf{CFG} & \textbf{PSNR (dB)} &
    \textbf{SSIM} &
    \textbf{ArcFace (Identity)}\\
    \midrule
    1.0 & 24.78 & 0.70 & 0.90\\
    2.0 & 24.54 & 0.69 & 0.89\\
    3.0 & 24.17 & 0.68 & 0.87\\
    4.0 & 23.73 & 0.67 & 0.85\\
    5.0 & 23.12 & 0.65 & 0.84\\
    \bottomrule
  \end{tabular} } 
 \end{center}
  \label{tab:ablation}
  \vspace{-4mm}
\end{table}

\end{document}